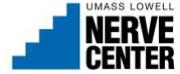

# UA-1 PH2 DECISIVE Testing Handbook

## Test Methods and Benchmarking Performance Results for sUAS in Dense Urban Environments

Adam Norton, Brendan Donoghue, and Peter Gavriel
New England Robotics Validation and Experimentation (NERVE) Center
University of Massachusetts Lowell
U.S. Army Combat Capabilities Development Command Soldier Center (DEVCOM-SC)
Contract # W911QY-20-2-0005

January 2025

**Abstract**: This report outlines all test methods and reviews all results derived from performance benchmarking of small unmanned aerial systems (sUAS) in dense urban environments conducted during Phase 2 of the Development and Execution of Comprehensive and Integrated Systematic Intelligent Vehicle Evaluations (DECISIVE) project by the University of Massachusetts Lowell (HEROES Project UA-1). Using 9 of the developed test methods, over 100 tests were conducted to benchmark the performance of 8 sUAS platforms: Cleo Robotics Dronut X1P (P = prototype), FLIR Black Hornet 3 PRS, Flyability Elios 2 GOV, Lumenier Nighthawk V3, Parrot ANAFI USA GOV, Skydio X2D, Teal Golden Eagle, and Vantage Robotics Vesper.

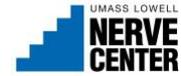

# Table of Contents

# Executive Summary

Over 100 tests were conducted using the 9 test methods in this handbook across 8 sUAS platforms: Cleo Robotics Dronut X1P (P = prototype), FLIR Black Hornet 3 PRS, Flyability Elios 2 GOV, Lumenier Nighthawk V3, Parrot ANAFI USA GOV, Skydio X2D, Teal Golden Eagle, and Vantage Robotics Vesper. It should be noted that the results contained in this report should be interpreted as benchmarks for each system at this particular moment in time and that their performance may differ in future evaluations due to system updates.

- = Not evaluated due to lack of system capability      - = Not evaluated due to lack of system availability or safety concerns

| Test Method | | Cleo Robotics Dronut X1P | FLIR Black Hornet 3 PRS | Flyability Elios 2 GOV | Lumenier Nighthawk V3 | Parrot ANAFI USA GOV | Skydio X2D | Teal Golden Eagle | Vantage Robotics Vesper |
|---|---|---|---|---|---|---|---|---|---|
| BLOS/NLOS Communications | | Safety concerns | Not available | 600 m range (50%) | 1200 m range (100%) | 1200 m range (100%) | 1200 m range (100%) | Not available | 700 m range (58%) |
| Automated Return to Home (RTH) | | Cannot RTH | Not available | Cannot RTH | Cannot RTH | 0 m error using GPS | 2 m error using GPS / 41 m error using VIO | Not available | 1 m error using GPS |
| Station Keeping in Wind | | Safety concerns | Not available | Not beyond 10 mph | Not available | Up to 25 mph | Up to 25 mph | Up to 20 mph | Up to 20 mph |
| Land and Takeoff In Wind | | Safety concerns | Not available | Not beyond 10 mph | Not available | Up to 25 mph | Up to 25 mph | Up to 20 mph | Up to 20 mph |
| Perched Field of Regard | Ground | 11% cover 20 mm acuity | Not available | 56% cover 12.8 mm acuity | 33% cover 20 mm acuity | 100% cover 3 mm acuity | 0% cover | 89% cover 7 mm acuity | 78% cover 5.9 mm acuity |
| | Wall | 22% cover 20 mm acuity | Not available | 56% cover 12.8 mm acuity | 22% cover 20 mm acuity | 67% cover 7.5 mm acuity | 0% cover | 56% cover 8.4 mm acuity | 44% cover 11 mm acuity |
| Exterior Building Clearing | Simple Building | Safety concerns | Not available | Safety concerns | 5 min 20 mm acuity | 9 min 2.1 mm acuity | 9 min 2.9 mm acuity | 12 min 8.5 mm acuity | Not available |
| | Complex Building | Safety concerns | Not available | Safety concerns | 11 min 19.3 mm acuity | 18 min 1.9 mm acuity | 20 min 2.8 mm acuity | 18 min 7.5 mm acuity | Not available |
| Fly Through Confined Outdoor Spaces | Alleyway, 1 m width | 33% 1 collision | 67% | 100% 1 collision | 100% 1 collision | 100% | 100% | Safety concerns | 67% 2 collisions |
| | Passageway, 2 m width | 0% 1 collision | 33% 1 collision | 100% 2 collisions | 100% | 100% | 100% | Safety concerns | 67% |
| | Corridor, 3 m width | Not available | 33% | 100% 1 collision | 100% | 100% | 100% | Safety concerns | 67% |
| Outdoor 2D Mapping Accuracy | Photos, Pix4Dreact | Cannot take photos | Not available | Not available | Not available | 2 min 1.22 m error | 1 min 1.17 m error | 2 min 1.1 m error | Not available |
| | Photos, Pix4Dmapper | Cannot take photos | Not available | Not available | Not available | 48 min 0.95 m error | 53 min 1.15 m error | 131 min 1.14 m error | Not available |
| | Photos, Farsight | Cannot take photos | Not available | Not available | Not available | 3 min 1.22 m error | 3 min 1.15 m error | 5 min 1.17 m error | Not available |
| | Video, Pix4Dmapper | Cannot record video | Not available | Not available | Not available | 25 min 1.2 m error | 16 min 1.02 m error | 16 min 4.1 m error | Not available |
| | Video, Farsight | Cannot record video | Not available | Not available | Not available | 45 min 1.3 m error | Map unable to be processed | 45 min 1.15 m error | Not available |
| Outdoor 3D Mapping Accuracy | Video, Pix4Dmapper | Cannot record video | Not available | Not available | Not available | 15 min 0.64 m error 100% features recognized | 16 min 0.35 m error 100% features recognized | 16 min 0.97 m error 90% features recognized | 48 min 0.14 m error 100% features recognized |
| | Video, Farsight | Cannot record video | Not available | Not available | Not available | 45 min 0.31 m error 100% features recognized | Map unable to be processed | 50 min 0.26 m error 100% features recognized | Map unable to be processed |

# Scope

The test methods specified in this handbook are scoped for evaluating sUAS intended for deployment in dense urban environments, with requirements for operating in these environments influenced by Army documents including ATP 3-06/MCTP 12-10B: Urban Operations[1], TC 90-1: Training for Urban Operations[2], and U.S. Army Subterranean and Dense Urban Environment MATDEV CoP Future Materiel Experiment (MATEx) Planning: Dense Urban Materiel Concepts and Capabilities RFI[3]. Specifications for nine test methods to evaluate sUAS capabilities are provided along with benchmarking data comparing the performance of several systems.

Each test method specification is written to function standalone, so some information is repeated across several test methods. All test method specifications follow a common format:

- Purpose: A brief description of the objectives of the test method.
- Summary of Test Method: A review of the pertinent components of the test method and the parameters that can be varied when defining a test.
- Apparatus and Artifacts: A description of any dimensional and material requirements of the environments where the tests are performed and any elements that need to be fabricated in order to run the test (e.g., wall panels, obstacles).
- Equipment: Tools and electronic devices used to support data collection (e.g., timers, sensors).
- Metrics: A definition of each metric evaluated in the test method.
- Procedure: Steps for the test administrator and operator to follow in order to conduct the test and analyze the metrics (if additional analysis description is needed beyond what is in the metrics section).

All test methods are designed to be run in real-world environments (e.g., MOUT sites) or using fabricated apparatuses (e.g., test bays built from wood, or contained inside of one or more shipping containers). All depictions of test method apparatuses contained throughout are examples.

Test results are provided after each test method specification, describing how the data was derived (e.g., where and when benchmarking took place, what the conditions of the test were, etc.) followed by charts and/or data tables to present all performance results.

Throughout the document, the term "sUAS" and "drone" are used interchangeably.

---

[1] Headquarters, Department of the Army; Headquarters, United States Marine Corps. ATP 3-06/MCTP 12-10B: Urban Operations. December 2017.
[2] Headquarters, Department of the Army. TC 90-1: Training for Urban Operations. May 2008.
[3] Army Subterranean and Dense Urban Environment Materiel Developer Community of Practice (SbT/DUE MATDEV CoP). U.S. Army Subterranean and Dense Urban Environment MATDEV CoP Future Materiel Experiment (MATEx) Planning: Dense Urban Materiel Concepts and Capabilities. February 2019, version 3.

# sUAS Platforms Evaluated

The sUAS platforms evaluated for this project were selected due to matching some of the desired performance capabilities for operating in subterranean and dense urban environments. These desired capabilities are derived from various Army reference documents, including the U.S. Army Subterranean and Dense Urban Environment MATDEV CoP Future Materiel Experiment (MATEx) Planning: Dense Urban Materiel Concepts and Capabilities RFI, as well as guidance from DEVCOM-SC. These capabilities include (in order of decreasing importance): GPS-denied operation, collision avoidance, ability to perch and stare, ability to operate in lowlight conditions, and small enough to comfortably fit through a typical door threshold.

A set of 8 platforms were evaluated, 4 of which are from the Blue sUAS list and the remaining 4 are NDAA compliant systems. While not all systems match all selection criteria, the 8 that were selected initially claimed to meet the minimum defined criteria for GPS-denied operation and being physically small enough to fit through a typical door threshold. The systems are listed and shown below.

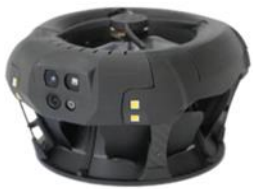
**Cleo Robotics Dronut X1P**
(P = prototype)

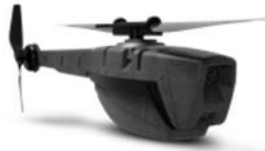
**FLIR Black Hornet PRS**

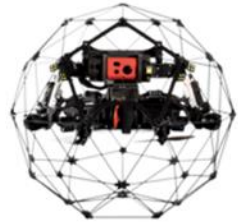
**Flyability Elios 2 GOV**

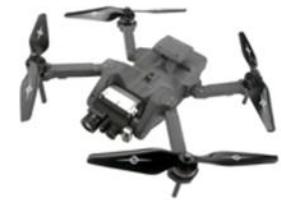
**Lumenier Nighthawk V3**

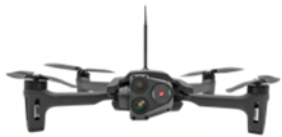
**Parrot ANAFI USA GOV**

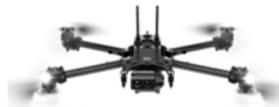
**Skydio X2D**

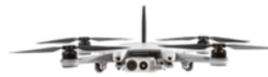
**Teal Golden Eagle**

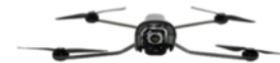
**Vantage Robotics Vesper**
(default = without prop guards;
CP = caged props with guards)

# BLOS/NLOS Outdoor Communications

## Test Method

**Purpose**

This test method evaluates the communications link between the OCU and the sUAS at increasing distances that exceed beyond line of sight (BLOS) and/or through obstructions for non-line-of-sight (NLOS), a common scenario for systems operating in dense urban environments. The ability to maintain connection and continue operation is vital for systems that operate without autonomous mission execution functionality such that the operator can continue operations despite BLOS/NLOS conditions.

**Summary of Test Method**

The sUAS attempts to fly to markers positioned away along a straight-line trajectory from the start position; if the system is able to maintain connection, the operator attempts to perform basic flight movements (hover, yaw, pitch, roll, ascend/descend, camera movement) and then inspect a visual acuity target. Successfully performing these actions demonstrates the quality of the communications connection to receive commands, maintain control, and transmit usable video back to the operator. The environment where this test is performed should allow for long-range flight along a straight path. The elevation of sUAS flight can be specified to be above any obstructions to not induce NLOS conditions or such that it is in line with any obstructions to induce NLOS. Regardless, it is expected that beyond a few hundred meters that BLOS conditions would be encountered due to limitations in visibility of the operator.

To outfit the test environment, set markers every 100 m along a straight line from a start position; see Figure 1 for example layouts. The markers should be designed to be large enough with significant color contrast to the ground such that they can be easily seen by the sUAS. Next to each distance marker, place an upward facing visual acuity target for the sUAS to inspect. To run the test, the sUAS is commanded to fly to each distance marker location, attempt to perform basic flight movements and target inspection, and then continue to the next marker location. The operator can manually fly the sUAS to each marker or GPS coordinates of the markers can be provided to the system for simpler navigation. This is performed until the system is no longer able to successfully complete these tasks, whereby the last distance in which all tasks were able to be successfully performed is then noted as the maximum BLOS distance. This provides an approximate measure of the point at which the sUAS would be expected to lose effective communications signal during a real-world deployment.

This test should be run prior to any other tests that involve operating the sUAS in a dense urban environment from a long-range distance. The primary performance metric, <u>maximum BLOS distance</u>, is used for the **Automated Return to Home (RTH)** test method.

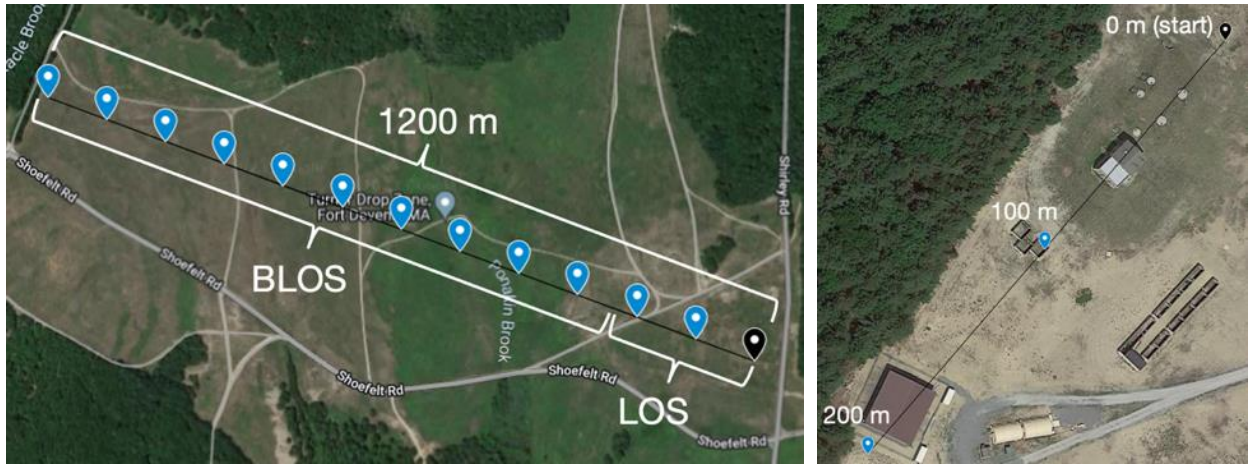

*Figure 1. Example layouts of the BLOS/NLOS Communications test method; the starting locations are shown in black and the distance markers every 100 m are shown in blue. <u>Left</u>: Unobstructed environment for BLOS testing over an open field up to 1200 m. <u>Right</u>: Obstructed urban environment for NLOS testing through buildings up to 200 m.*

## Apparatus and Artifacts

A real-world outdoor environment that allows for the maximum BLOS distance desired to be tested with obstructions (e.g., buildings, hills) or without (e.g., runway, open field). Numbered distance markers of sufficient size and color contrast and visual acuity targets are placed at each 100 m distance along the straight-line trajectory. See Figure 2 for examples of 60 cm [24 in] square distance markers consisting of a wooden panel covered in red duct tape with white numbers for increased visibility, along with the standard 20 cm [8 in] visual acuity targets mounted to bucket lids.

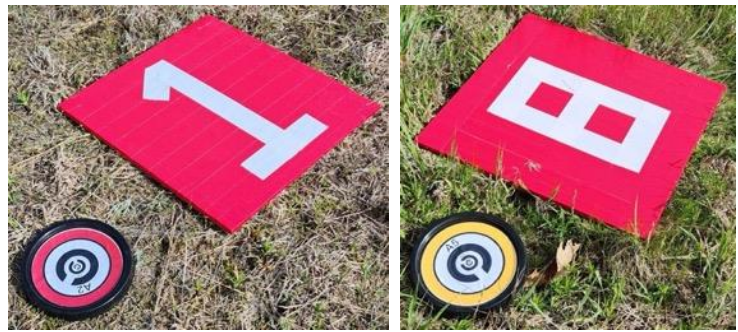

*Figure 2. Example numbered distance markers with visual acuity targets.*

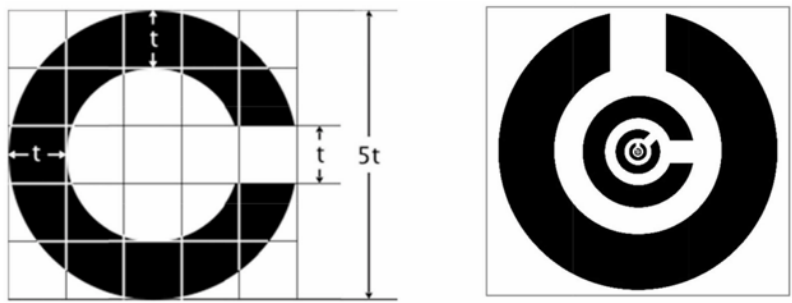

*Figure 3. <u>Left</u>: Dimensional configuration of a Landolt C using a standard unit (t). <u>Right</u>: Example nested Landolt C configuration from a visual acuity target. In this example, from the largest to the smallest C, the directions of the C openings are top, right, top-right, top, and top-left.*

**Equipment**

The numbered distance markers and visual acuity targets must be positioned every 100 m along a straight-line trajectory from the start position, requiring some form of measurement device to ensure these are placed accurately (i.e., within +/- 1 m). If this test is performed in an environment with pre-existing straight lines to measure off of (e.g., painted lines on a runway), then a rolling measure wheel can be used to make these marks. In a less structured outdoor environment (e.g., rolling outdoor hills, mock cityscape environment), GPS locations for each distance are calculated first and then a GPS unit is used to guide the placement of each marker and target.

**Metrics**

- Connection quality: Ability of the operator to send control commands through the OCU to the sUAS and stream video of the sUAS camera to the OCU. Qualitatively evaluated as either good (✓), bad (//), or none (X). Indicators of "bad" video link quality include screen tearing, pixelation, and other artifacts not present when video link quality is "good."
- Video link quality: Visual appearance of the video stream of the sUAS camera on the OCU. Qualitatively evaluated as either good (✓), bad (//), or none (X).
- OCU signal indication: Any signal indication provided on the OCU such as a bar chart or numerical readout for the control and/or video signal level.
- Flying performance: Ability of the operator to control the sUAS to hover, yaw, pitch forward and backward, roll left and right, ascend and descend, and move the camera, while the sUAS is at positioned at each distance marker. Qualitatively evaluated as either possible (✓) or not possible (X).
- Visual acuity: Level of detail that can be resolved in the available Landolt C artifacts during flight, reported per visual acuity target able to be seen. Each size of Landolt C opening, from the largest to the smallest C, is 20, 8, 3, 1.3, and 0.5 mm [0.8, 0.3, 0.125, 0.05, and 0.02 in], smaller is better (i.e., smaller details are able to be resolved). Reported in millimeters (mm).
- Maximum BLOS distance: The maximum position distance where the sUAS is able to maintain communications link for both flying performance (hover, yaw, pitch, roll, ascend/descend, and camera movement) and inspection performance (visual acuity of at least the largest C). Reported in meters (m) and as a percentage (%) of the maximum possible distance due to space limitations (1200 m).

**Procedure**

1. Position the numbered distance markers with visual acuity targets in order (i.e., 1 = 100 m, 2 = 200 m, etc.) at each 100 m distance segment along a straight-line trajectory from the start position.
2. Position the operator with OCU at the start position (i.e., 0 m).
3. Launch the sUAS and confirm initial connection quality, video link quality, OCU signal indication (if applicable), flying performance, and visual acuity at start position.
4. Command the sUAS to fly to the next distance marker.
5. Once the distance marker is reached, record connection quality, video link quality, and OCU signal indication (if applicable).
6. If connection quality and video link quality are viable, attempt flying performance: yawing in place, pitching forward and back, rolling left and right, ascending and descending, and moving the camera.
7. If flying performance is viable, attempt inspecting the visual acuity target.
8. Record the outcome of each task attempt.
9. Repeat steps 4 through 8 until both the connection quality and video link quality fail, until the sUAS does not allow itself to be flown further (e.g., it warns the operator and prevents them from continuing), or until the end of the available testing distance is reached.

## Test Results

Benchmarking was conducted at Fort Devens: Facility 10, Turner Drop Zone, which allows for 1200 m of long-range flight (see Figure 1). The range consists of undulating hills that, while not drastic (see Figure 4), does induce NLOS conditions when the sUAS is at lower elevations. Test results are derived from benchmarking conducted in May 2023. The Teal Golden Eagle was not operational at testing time, so no test results are provided.

A summary of the test results is provided in Figure 5 followed by detailed results. Achieving a higher maximum BLOS distance (i.e., longer range) and with higher acuity (i.e., smaller details able to be resolved) is desirable.

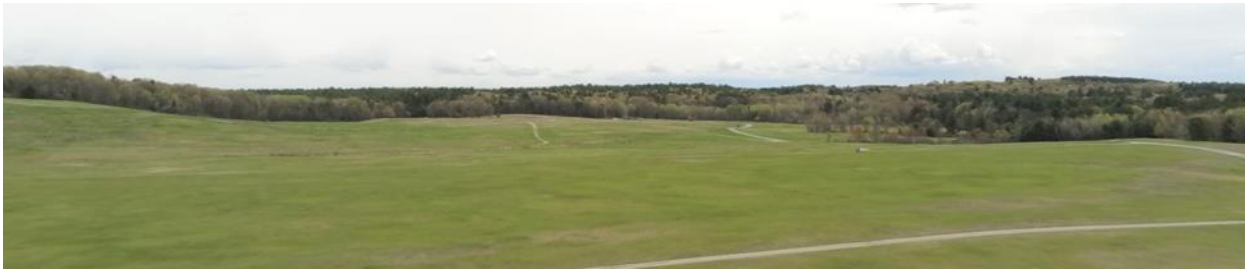

*Figure 4. A view of the undulating hills in Turner Drop Zone as seen by the Parrot ANAFI USA GOV.*

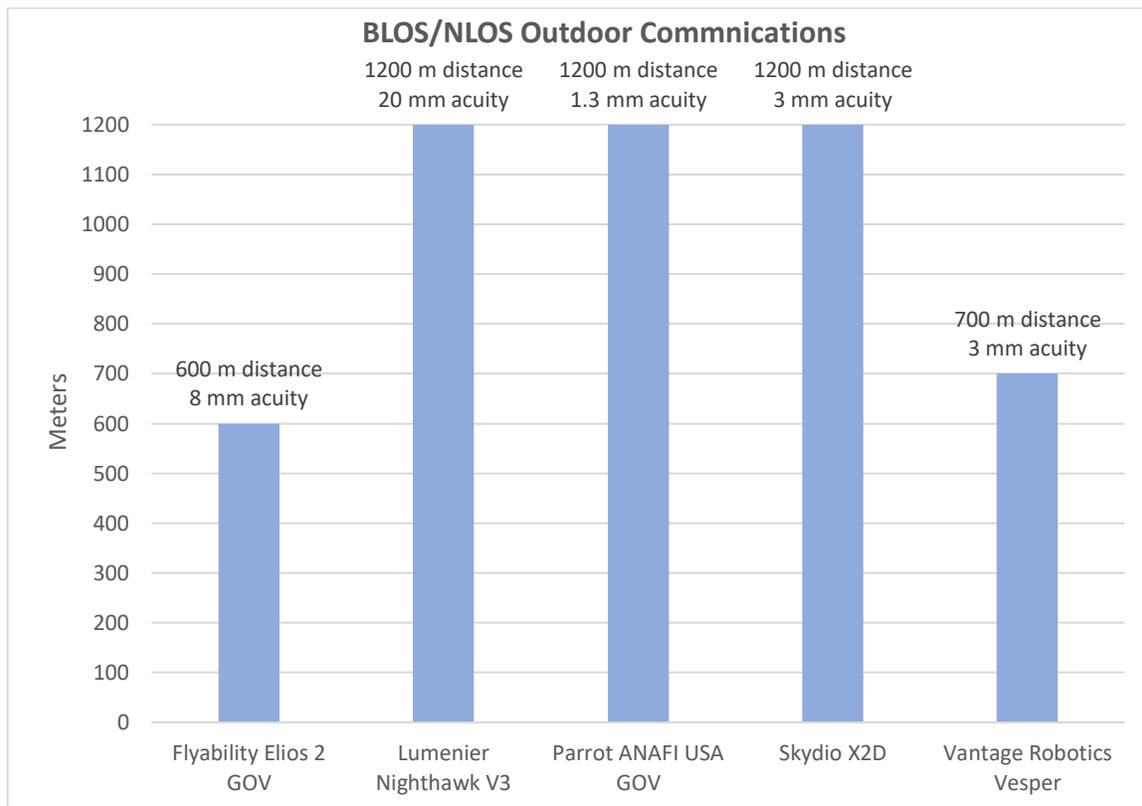

*Figure 5. Summarized BLOS/NLOS Outdoor Communications test results.*

| sUAS | Metrics | Distance (m) | | | | | | | | | | | | | Example Acuity Target (from sUAS camera) |
| --- | --- | --- | --- | --- | --- | --- | --- | --- | --- | --- | --- | --- | --- | --- | --- |
| | | LOS | | | BLOS | | | | | | | | | | |
| | | 0 | 100 | 200 | 300 | 400 | 500 | 600 | 700 | 800 | 900 | 1000 | 1100 | 1200 | |
| Flyability Elios 2 GOV | Connect | ✓ | ✓ | // | // | // | // | // | ✗ | | | | | | |
| | Video link | ✓ | ✓ | ✓ | ✓ | ✓ | ✓ | ✓ | ✗ | | | | | | |
| | OCU | 5/5 | 5/5 | 3/5 | 3/5 | 2/5 | 1/5 | 1/5 | ✗ | | | | | | |
| | Fly | ✓ | ✓ | ✓ | ✓ | ✓ | ✓ | ✓ | ✗ | | | | | | |
| | Acuity (mm) | 8 | 8 | 8 | 8 | 8 | 8 | 8 | ✗ | | | | | | |
| | Max BLOS | 600 m (50%) | | | | | | | | | | | | | |
| Lumenier Nighthawk V3 | Connect | ✓ | ✓ | ✓ | ✓ | ✓ | ✓ | ✓ | ✓ | ✓ | ✓ | ✓ | ✓ | ✓ | |
| | Video link | ✓ | ✓ | ✓ | ✓ | ✓ | ✓ | ✓ | ✓ | ✓ | ✓ | ✓ | ✓ | ✓ | |
| | OCU | n/a | | | | | | | | | | | | | |
| | Fly | ✓ | ✓ | ✓ | ✓ | ✓ | ✓ | ✓ | ✓ | ✓ | ✓ | ✓ | ✓ | ✓ | |
| | Acuity (mm) | 20 | 20 | 20 | 20 | 20 | 20 | 20 | 20 | 20 | 20 | 20 | 20 | 20 | |
| | Max BLOS | 1200 m (100%) | | | | | | | | | | | | | |
| Parrot ANAFI USA GOV | Connect | ✓ | ✓ | ✓ | ✓ | ✓ | ✓ | ✓ | ✓ | ✓ | ✓ | ✓ | ✓ | ✓ | |
| | Video link | ✓ | ✓ | ✓ | ✓ | ✓ | ✓ | ✓ | ✓ | ✓ | ✓ | ✓ | ✓ | ✓ | |
| | OCU | ✓ | ✓ | ✓ | ✓ | ✓ | ✓ | ✓ | ✓ | ✓ | ✓ | ✓ | ✓ | ✓ | |
| | Fly | ✓ | ✓ | ✓ | ✓ | ✓ | ✓ | ✓ | ✓ | ✓ | ✓ | ✓ | ✓ | ✓ | |
| | Acuity (mm) | 1.3 | 1.3 | 1.3 | 1.3 | 1.3 | 1.3 | 1.3 | 1.3 | 1.3 | 1.3 | 1.3 | 1.3 | 1.3 | |
| | Max BLOS | 1200 m (100%) | | | | | | | | | | | | | |
| Skydio X2D | Connect | ✓ | ✓ | ✓ | ✓ | ✓ | ✓ | ✓ | ✓ | // | // | ✓ | ✓ | ✓ | |
| | Video link | ✓ | ✓ | ✓ | ✓ | ✓ | ✓ | ✓ | ✓ | ✓ | ✓ | ✓ | ✓ | ✓ | |
| | OCU | 4/4 | 4/4 | 4/4 | 4/4 | 4/4 | 4/4 | 4/4 | 4/4 | 3/4 | 3/4 | 4/4 | 4/4 | 4/4 | |
| | Fly | ✓ | ✓ | ✓ | ✓ | ✓ | ✓ | ✓ | ✓ | ✓ | ✓ | ✓ | ✓ | ✓ | |
| | Acuity (mm) | 3 | 3 | 3 | 3 | 3 | 3 | 3 | 3 | 3 | 3 | 3 | 3 | 3 | |
| | Max BLOS | 1200 m (100%) | | | | | | | | | | | | | |
| Vantage Robotics Vesper | Connect | ✓ | ✓ | ✓ | // | // | // | // | // | ✗ | | | | | |
| | Video link | ✓ | ✓ | ✓ | ✓ | ✓ | ✓ | ✓ | // | ✗ | | | | | |
| | OCU | 5/5 | 5/5 | 5/5 | 3/5 | 3/5 | 2/5 | 2/5 | 2/5 | ✗ | | | | | |
| | Fly | ✓ | ✓ | ✓ | ✓ | ✓ | ✓ | ✓ | ✓ | ✗ | | | | | |
| | Acuity (mm) | 3 | 3 | 3 | 3 | 3 | 3 | 3 | 3 | ✗ | | | | | |
| | Max BLOS | 700 m (58%) | | | | | | | | | | | | | |

*Table 1. BLOS/NLOS Outdoor Communications test results.*

# Automated Return to Home (RTH)

## Test Method

**Purpose**

This test method evaluates the sUAS' capability to automatically return to home (RTH) from a specified distance. This capability is often engaged when a mission is complete (i.e., the operator commands the system to return to home) or as a preventative measure due to issues with power or communications (e.g., the OCU loses power and comms link is lost, so the system automatically returns to home as a safety measure).

**Summary of Test Method**

The sUAS is flown to a desired distance and then commanded to perform automated return to home (RTH). Typically, systems perform RTH using GPS to navigate from their current position back to the start position, but some systems are also able to do so in GPS-denied environments by using visual inertial odometry (VIO) and/or dead reckoning. Once the sUAS has reached what it believes is the "home" position, it is commanded to land. The environment where this test is performed should allow for long-range flight along a straight path. The elevation of sUAS flight can be specified to be above any obstructions to not interfere with it while performing RTH (the *default* configuration) or such that it is in line with any obstructions that it would have to avoid while performing RTH if simultaneous evaluation of obstacle avoidance is desired (referred to as the *obstructed* configuration).

To outfit the test environment, set markers every 100 m along a straight line from a start position; see Figure 6 for example layouts. The markers should be designed to be large enough with significant color contrast to the ground such that they can be easily seen by the sUAS. To run the test, the sUAS is commanded to perform automated RTH from a specified distance and then land once it has reached what it believes is the start position. The operator can manually fly the sUAS to a desired distance marker or GPS coordinates of the markers can be provided to the system for simpler navigation. For simplified testing, the sUAS can be commanded to fly to the maximum BLOS distance (derived from the **BLOS/NLOS Outdoor Communications** test method) and attempt RTH. Additional tests can be conducted at shorter distances if desired. Once landed, the distance of the sUAS from its actual starting position is measured and used to evaluate its accuracy in performing RTH.

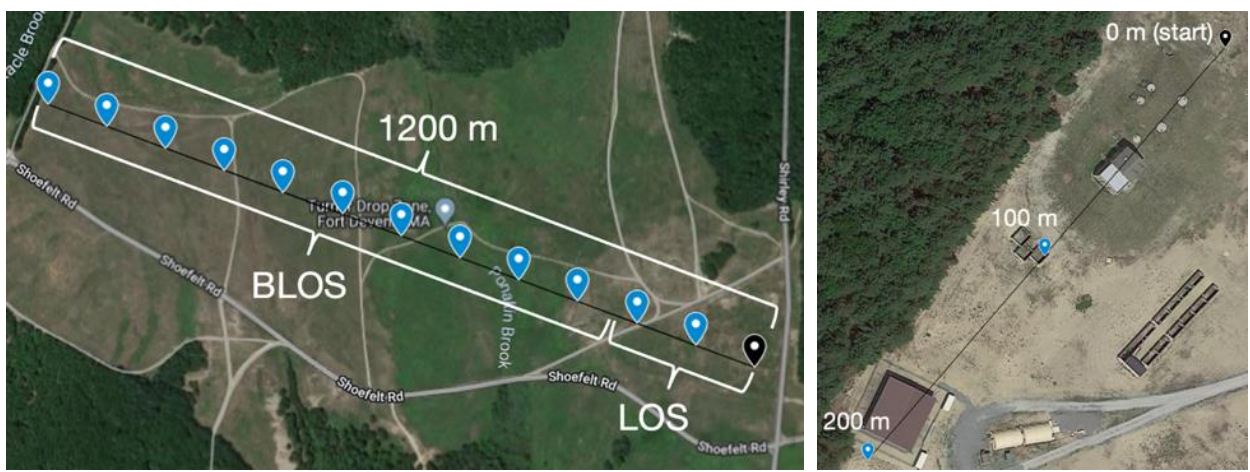

*Figure 6. Example layouts of the Automated Return to Home (RTH) test method; the starting locations are shown in black and the distance markers every 100 m are shown in blue. Left: Default configuration in an unobstructed environment over an open field up to 1200 m. Right: Obstructed configuration through buildings up to 200 m.*

## Apparatus and Artifacts

A real-world outdoor environment that allows for the maximum BLOS distance desired to be tested with obstructions (e.g., buildings, hills; *obstructed* configuration) or without (e.g., runway, open field; *default* configuration). Numbered distance markers of sufficient size and color contrast are placed at each 100 m distance along the straight-line trajectory with a start position launch/landing pad position at the start (i.e., 0 m). See Figure 7 for examples of 60 cm [24 in] square distance markers consisting of a wooden panel covered in red duct tape with white numbers for increased visibility and an example start position.

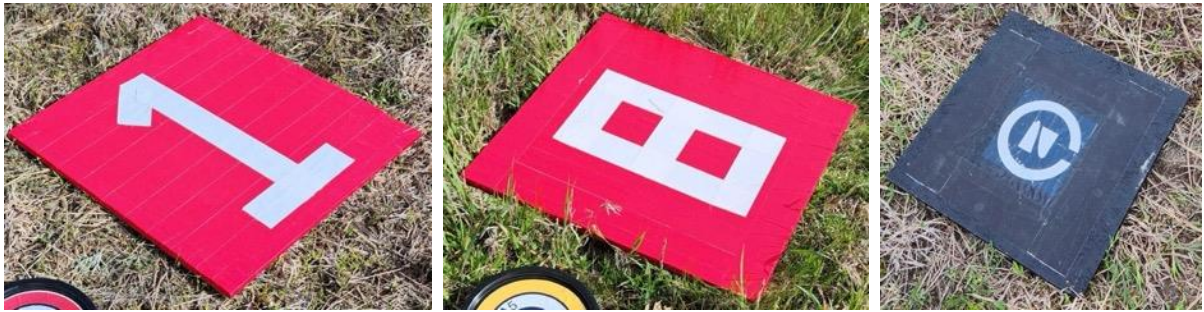

*Figure 7. Left: Example numbered distance markers. Right: Start position launch/land pad.*

## Equipment

The numbered distance markers must be positioned every 100 m along a straight-line trajectory from the start position, requiring some form of measurement device to ensure these are placed accurately (i.e., within +/- 1 m). If this test is performed in an environment with pre-existing straight lines to measure off of (e.g., painted lines on a runway), then a rolling measure wheel can be used to make these marks. In a less structured outdoor environment (e.g., rolling outdoor hills, mock cityscape environment), GPS locations for each distance are calculated first and then a GPS unit is used to guide the placement of each marker and target.

A timer is used to measure time taken to return to home. A measure tape or similar is used to measure the distance between the location where the sUAS lands and the start position.

## Metrics

- Maximum BLOS distance: The maximum position distance where the sUAS is able to maintain communications link for both flying performance (hover, yaw, pitch, roll, ascend/descend, and camera movement) and inspection performance (visual acuity of at least the largest C), as derived from the **BLOS/NLOS Outdoor Communications** test method. Reported in meters (m).
- Return time: The amount of time between when the sUAS is commanded to perform RTH and when it lands at what it believes is the start position. Reported in minutes (min).
- Return distance error: The distance between the start position and the location where the sUAS lands when it believes it is at the start position. Reported in meters (m).

**Procedure**

1. Conduct the **BLOS/NLOS Outdoor Communications** test and derive the <u>maximum BLOS distance</u> metric.
2. Position the sUAS at the start position.
3. Launch the sUAS.
4. Command the sUAS to fly to the distance marker at the <u>maximum BLOS distance</u>.
5. Once the distance marker is reached, command the sUAS to perform automated RTH and start the timer.
6. Once the sUAS reaches what it believes is the start position, command it to land and stop the timer.
7. Measure the distance between where the sUAS lands and the start position.

## Test Results

Benchmarking was conducted at Fort Devens: Facility 10, Turner Drop Zone, which allows for 1200 m of long-range flight (see Figure 6). The range consists of undulating hills that, while not drastic (see Figure 8), does break line-of-sight when the sUAS is at lower elevations. Test results are derived from benchmarking conducted in May 2023. Only the Parrot ANAFI USA GOV, Skydio X2D, Teal Golden Eagle, and Vantage Robotics Vesper platforms are able to perform automated RTH, but the Teal Golden Eagle was not operational at testing time. All systems were able to perform RTH using GPS while only the Skydio X2D was able to do so using vision.

Achieving a lower return distance error (i.e., higher accuracy) is desirable.

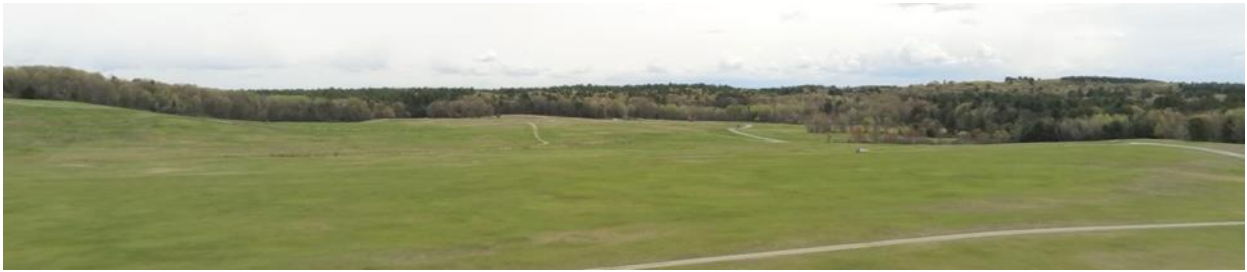

*Figure 8. A view of the undulating hills in Turner Drop Zone as seen by the Parrot ANAFI USA GOV.*

| sUAS | Metrics | Performance | | | |
|---|---|---|---|---|---|
| Parrot ANAFI USA GOV | Max BLOS | 1200 m | | | |
| | RTH Method | GPS | | | |
| | Return time | 2 min | | | |
| | Return distance error | 0 m | | | |
| Skydio X2D | Max BLOS | 1200 m | | 1200 m | |
| | RTH Method | GPS | | Vision | |
| | Return time | 3 min | | 3 min | |
| | Return distance error | 2 m | | 41 m | |
| Vantage Robotics Vesper | Max BLOS | 700 m | | | |
| | RTH Method | GPS | | | |
| | Return time | 1 min | | | |
| | Return distance error | 1 m | | | |

*Table 2. Automated Return to Home (RTH) test results.*

# Station Keeping In Wind

## Test Method

**Purpose**

This test method is used to evaluate the impact of wind on a sUAS' ability to hover in place (i.e., station keep) using induced wind conditions of varying speed and direction. This is an essential capability when operating in outdoor environments subject to weather effects.

**Summary of Test Method**

The sUAS attempts to station keep while within wind flow of varying speed and the orientation at which the wind is pushing against the system. To evaluate the ability to keep station, visual acuity targets in buckets are placed around the area where the system is supposed to hover in place, positioned such that if the sUAS maintains its position, then the view of the target shall be maintained. As a controllable wind source, a wind generator is used for consistent wind and the sUAS is positioned within the wind flow. At each desired wind speed, the sUAS is oriented either 0°, 90°, 180°, or 270° to the wind flow and attempts to inspect the visual acuity targets. Two configurations of visual acuity targets can be used: (1) one upward-facing target below the system's hover position, or (2) four targets positioned radially around the system's hover position. See Figure 9.

The sUAS hovers such that its elevation height is approximately centered in the wind flow. Once positioned as specified and aligned with the target – meaning the entirety of the black or white ring on the inside of the outer color ring is completely visible (see Figure 10) – the target must remain aligned within the camera view for at least 5 seconds while the operator is either hands off (i.e., the system is able to keep station on its own) or hands on (i.e., the operator must continue commanding the system to correct its position in order for it to keep station), with hands off performance valued higher than hands on performance. The results of this test indicate the types of wind conditions that the sUAS should be expected to be able to survive in with or without operator input to maintain its position and remain in the air.

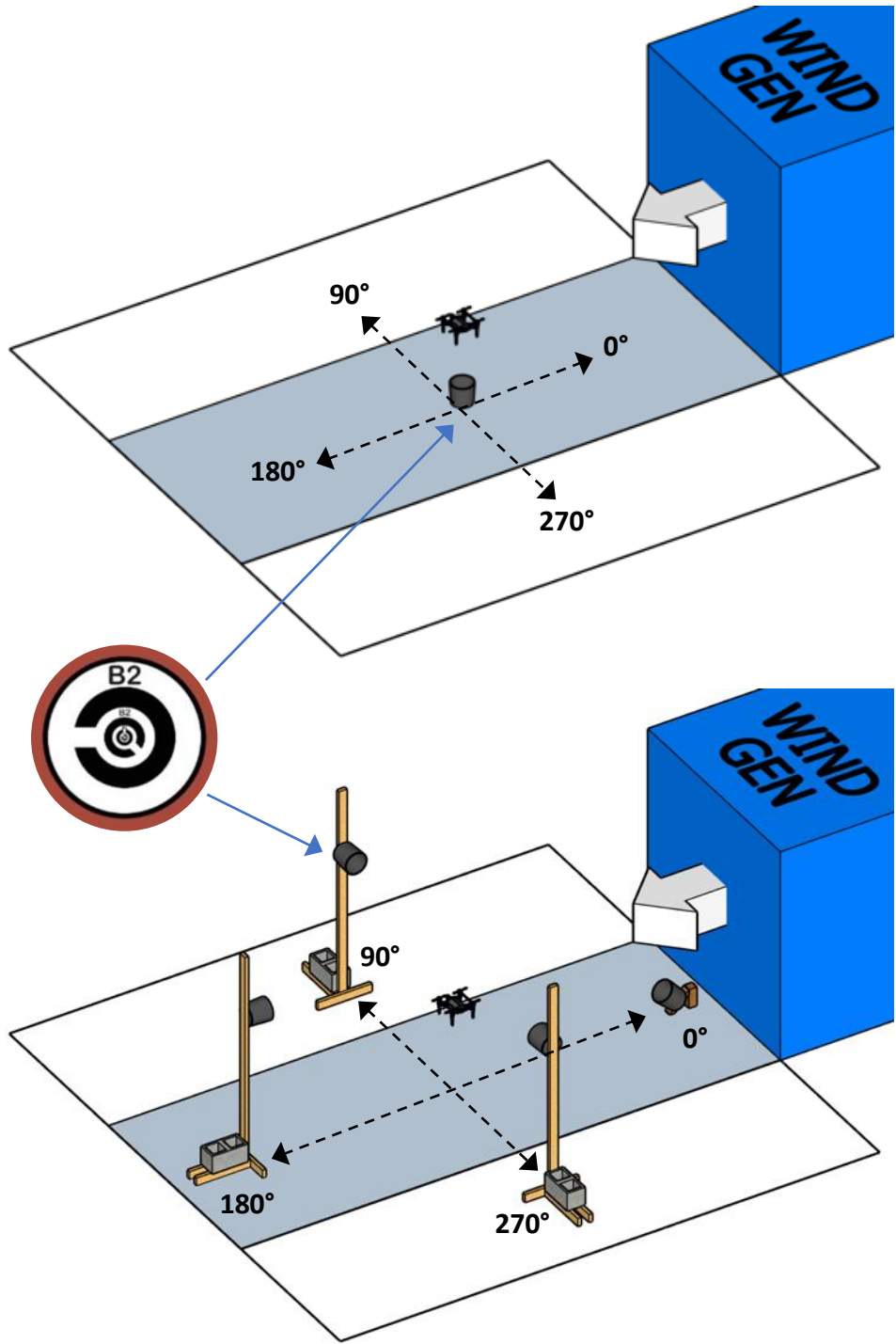

*Figure 9. Two configurations for the Station Keeping In Wind test method. <u>Top</u>: Using a one target below the sUAS. <u>Bottom</u>: Using four targets positioned around the sUAS.*

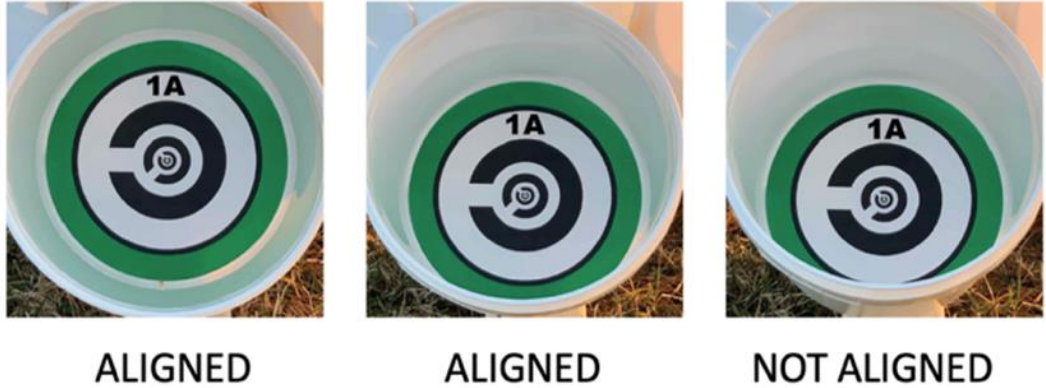

*Figure 10. Correct alignment of the visual target is defined as when the operator can see the entire black or white ring on the inside of the colored ring, as shown in the left and middle image.*

**Apparatus and Artifacts**

Visual acuity targets consist of five nested Landolt C optotypes of varying sizes and orientations; the opening of each C corresponds to a different level of detail that, if the orientation of the C opening is able to be correctly identified, correspond to different levels of detail able to be visually resolved. The dimensions of a Landolt C are described using a standard unit (t); see ASTM E2566[4] for more information. For the visual acuity targets used in this test method, from the largest to the smallest C, the size of t = 20, 8, 3, 1.3, and 0.5 mm [0.8, 0.3, 0.125, 0.05, and 0.02 in]. See Figure 11.

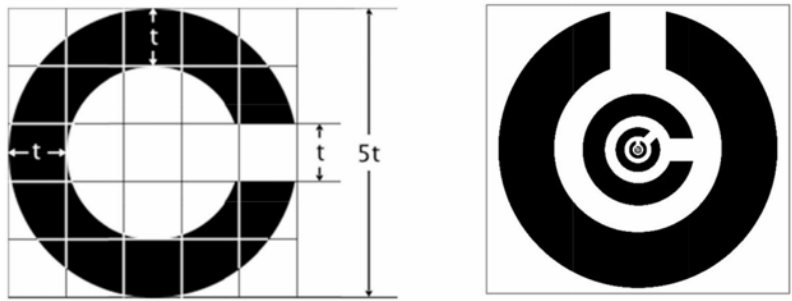

*Figure 11. Left: Dimensional configuration of a Landolt C using a standard unit (t). Right: Example nested Landolt C configuration from a visual acuity target. In this example, from the largest to the smallest C, the directions of the C openings are top, right, top-right, top, and top-left.*

There are two possible configurations of visual acuity targets that can be used: (1) one upward-facing visual acuity target (viewable from above) is placed on the ground directly underneath the station keeping position (so it can be inspected by the system's downward-facing camera, or a gimbaled camera pointed downwards, that the operator can see on the OCU), or (2) four visual acuity targets are positioned radially around the sUAS at 0°, with three of them mounted on posts as forward-facing targets (viewable from the front) elevated to the sUAS hover height and positioned at 90°, 180°, and 270°, and one angled target (viewable from 45° up from the ground) positioned at 0° (i.e., below the wind generator source). In either configuration, the targets are mounted inside of 1-gallon buckets that measure approximately 20 $cm^3$ [8 $in^3$]. See Figure 9 for the two layouts and Figure 12 for each target type.

---

[4] ASTM International. ASTM E2566/E2566M – 24: Standard Test Method for Evaluating Response Robot Sensing: Visual Acuity. ASTM International Book of Standards Volume 15.08, DOI: 10.1520/E2566_E2566M-24.

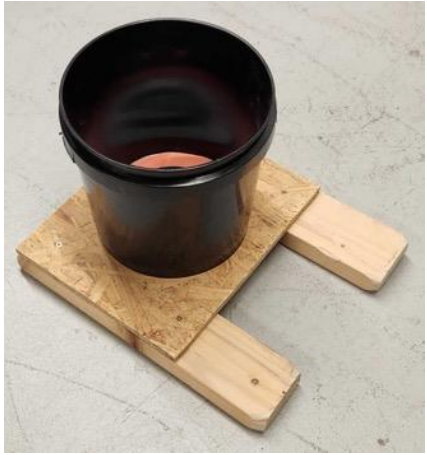

Upward-facing target (viewable from above)

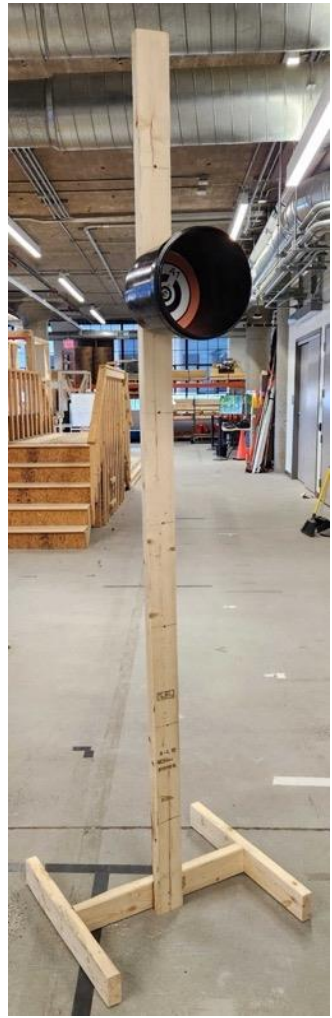

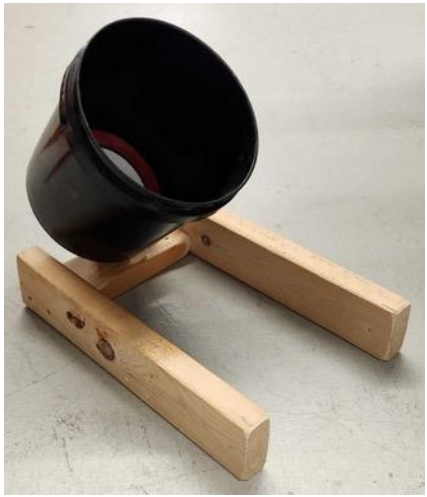

Angled target (viewable from 45°)   Forward-facing target (viewable from the front)

*Figure 12. Targets used for the Station Keeping In Wind test method.*

## Equipment

A wind generator is used to induce wind flow for testing. Ideally, a system that generates laminar flow with controllable wind speed is used. A wind meter should be used to confirm the wind speed being generated before evaluating the sUAS' ability to keep station. Safety nets should be positioned around the testing area to catch the system should it become unstable during testing.

A timer is used to count down from 5 seconds each time the sUAS is in position and attempting to keep station.

## Metrics

- <u>Wind speed</u>: The speed of the wind flow used during a test trial. Reported in miles per hour (mph) in increments of 5 mph.
- <u>Orientation</u>: The sUAS orientation to the wind flow during a test trial, either 0°, 90°, 180°, or 270°.
- <u>Control method</u>: Whether or not the operator was actively teleoperating the sUAS while it was attempting to keep station by nudging it back to center position in order to maintain view of the visual acuity target. Reported as Hands ON or Hands OFF, where Hands OFF is valued higher than Hands ON.
- <u>Success</u>: The number of trials that were successful, where 1 trial = 1 orientation and wind speed combination. Hands OFF = 25% per successful orientation (i.e., 1 out of 4 orientations possible), Hands ON = 12.5% per successful orientation (i.e., half of 25% to favor Hands OFF performance), and X = 0%, for a possible total of 100% per orientation and wind speed combination. Reported as a percentage (%).

## Procedure

1. Select the desired visual acuity target configuration and layout the test space accordingly.
2. Instruct the operator that, once the administrator confirms the sUAS camera is properly aligned with the target, the operator will indicate that the 5 second countdown for station keeping is ready to begin either with their hands on or off of the controller. For example, the administrator may verbally confirm alignment followed by the operator indicating they are ready for the countdown by raising a finger (i.e., to indicate they are hands off).
3. Command the sUAS to takeoff and hover in place in front of the wind generator where the wind will be once it is actively flowing.
4. While under ambient wind conditions (i.e., no induced wind from the wind generator), instruct the operator to align the sUAS camera with the target at 0°.
5. After the established alignment confirmation and countdown indication markers have been communicated, use the timer to countdown 5 seconds.
6. If the operator is attempting to keep station while hands off, and the system becomes unaligned during the 5 second countdown, the operator can re-align to try hands off station keeping again or they can try again hands on (either way, an additional trial for that condition will be recorded).
7. After that condition is completed, instruct the operator to rotate to 90° and align the sUAS camera with the next target at that orientation.
8. Repeat steps 5-7 until all four orientations have been completed.
9. Turn on the wind generator to the desired wind speed.
10. Instruct the operator to align the sUAS camera with the target at 0°.
11. Repeat steps 5-7 until all four orientations have been completed, then increase the wind speed by 5 mph.
12. Repeat steps 10-11 until all desired wind speeds have been tested.

## Test Results

Benchmarking was conducted at the KRI ECUAS Lab using the wind profile generator (WPG) in the outdoor netted area (see Figure 13, Figure 14, and Figure 15) using the single upward facing target configuration. Test results are derived from benchmarking conducted in November 2023.

A summary of the test results is provided in Figure 16 followed by detailed results.

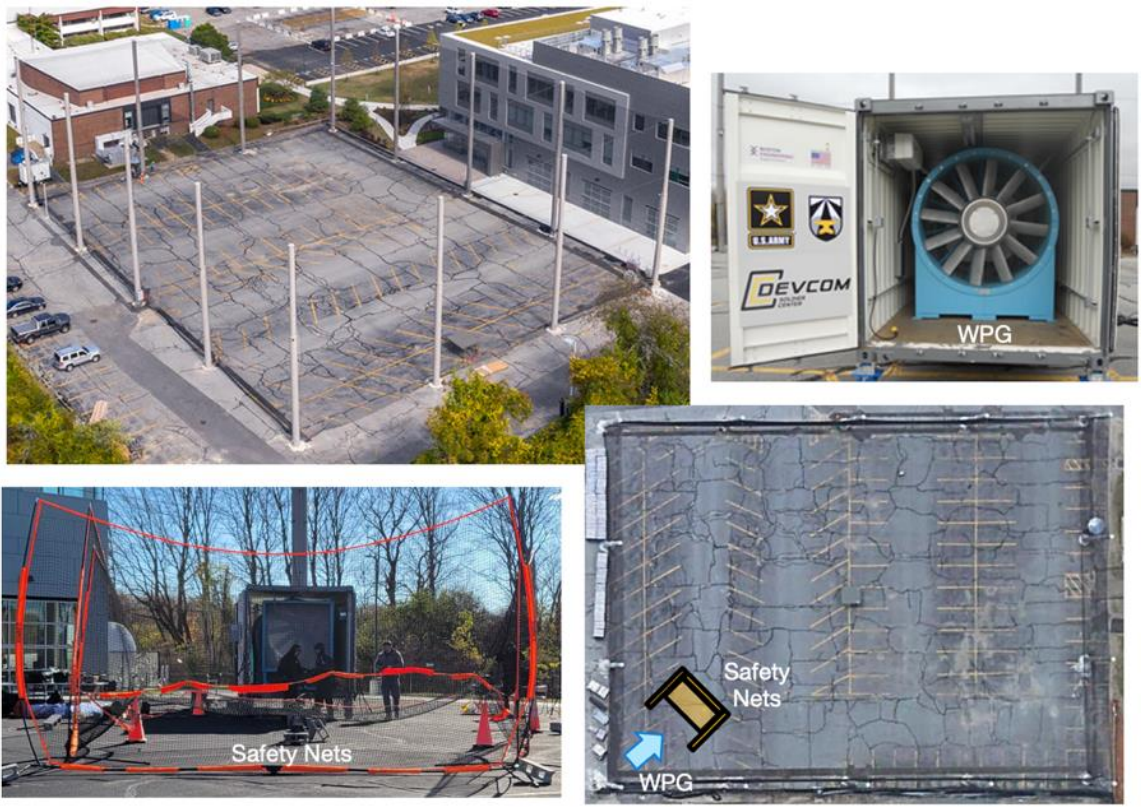

Figure 13. The outdoor netted area and wind profile generator (WPG) at the KRI ECUAS Lab with safety nets used for testing.

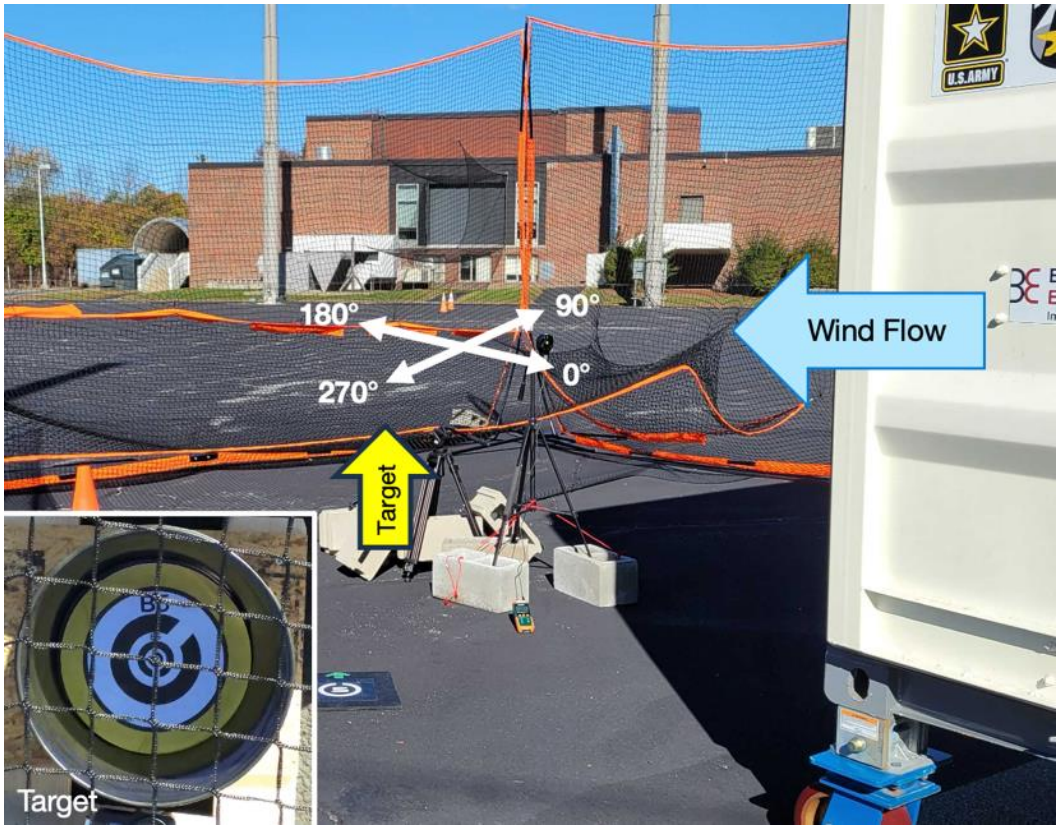

*Figure 14. Experimental set-up for the Station Keeping In Wind test method at the KRI ECUAS Lab.*

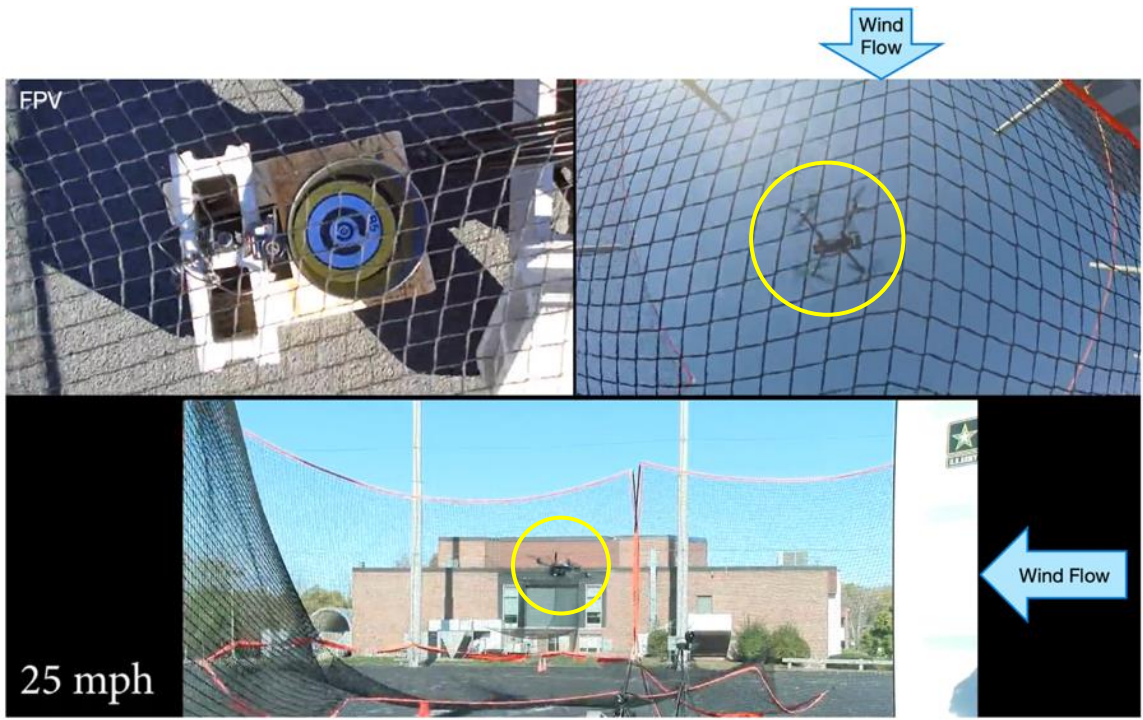

*Figure 15. Screenshot from test video of Skydio X2D performing the Station Keeping In Wind test in 25 mph wind speed at 90° orientation.*

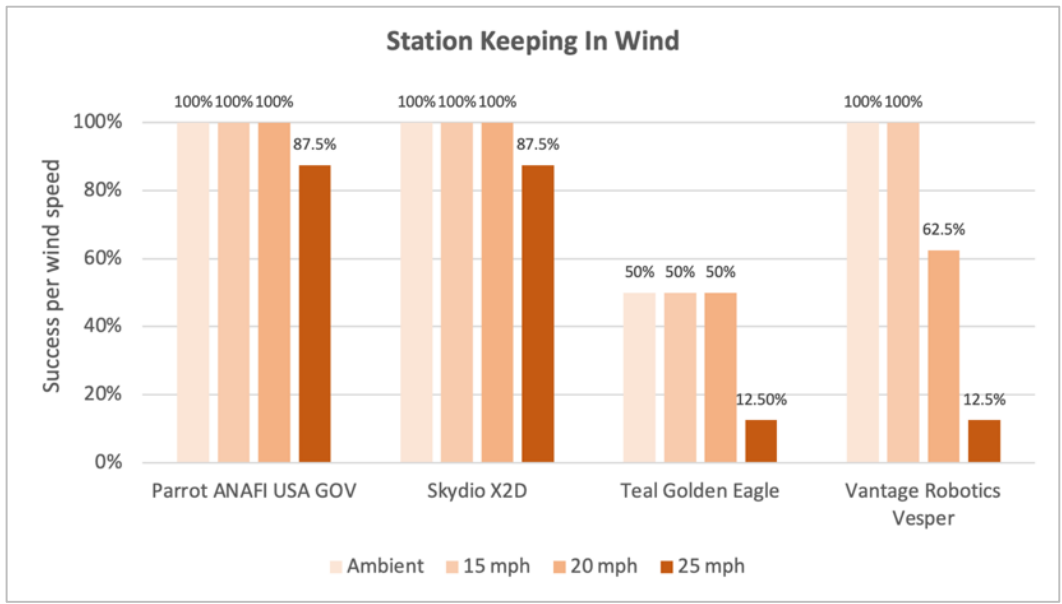
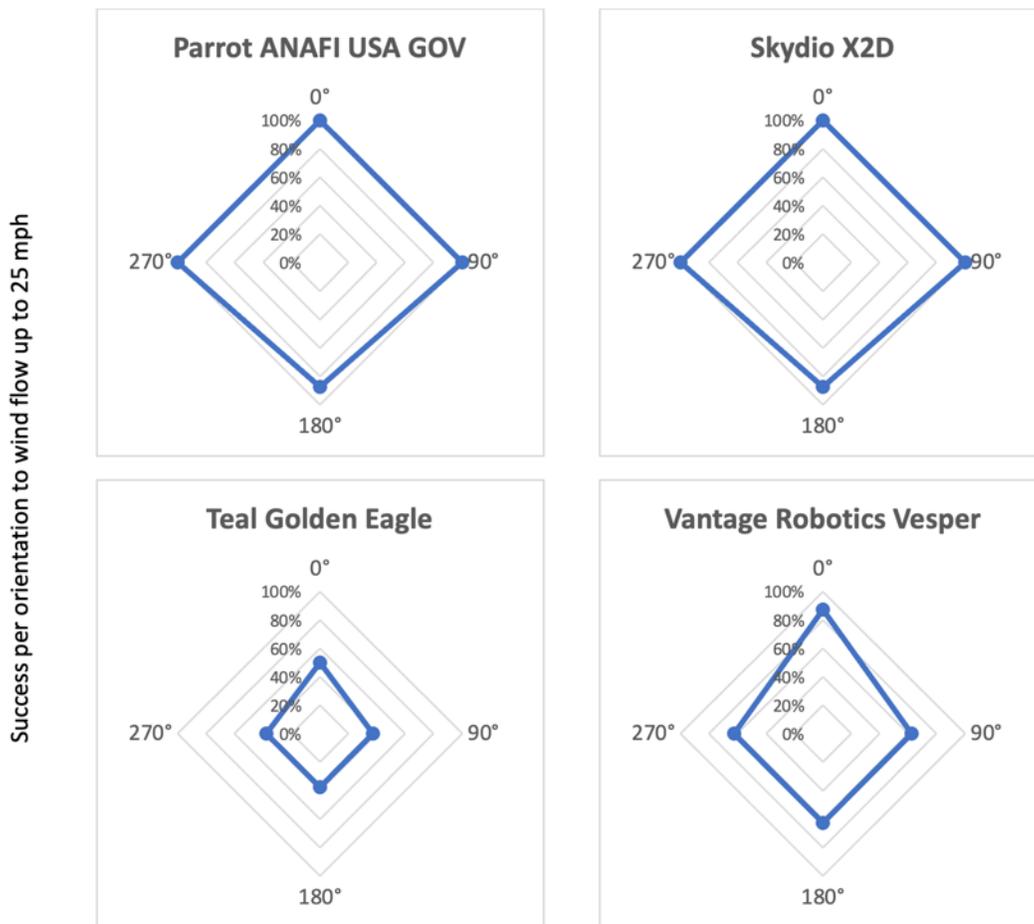

*Figure 16. Station Keeping In Wind test results per wind speed (top) and per sUAS orientation to wind flow (bottom).*

| sUAS | Orientation / Metrics | Wind Speed | | | | | Success per orientation* |
|---|---|---|---|---|---|---|---|
| | | Ambient | 15 mph | 20 mph | 25 mph | 30 mph | |
| Parrot ANAFI USA GOV | 0° | Hands OFF | Hands OFF | Hands OFF | Hands OFF | X | 100% |
| | 90° | Hands OFF | Hands OFF | Hands OFF | Hands OFF | | 100% |
| | 180° | Hands OFF | Hands OFF | Hands OFF | Hands ON | | 87.5% |
| | 270° | Hands OFF | Hands OFF | Hands OFF | Hands OFF | | 100% |
| | Success per wind speed | 100% | 100% | 100% | 87.5% | 0% | --- |
| Skydio X2D | 0° | Hands OFF | Hands OFF | Hands OFF | Hands OFF | X | 100% |
| | 90° | Hands OFF | Hands OFF | Hands OFF | Hands OFF | | 100% |
| | 180° | Hands OFF | Hands OFF | Hands OFF | Hands ON | | 87.5% |
| | 270° | Hands OFF | Hands OFF | Hands OFF | Hands OFF | | 100% |
| | Success per wind speed | 100% | 100% | 100% | 87.5% | 0% | --- |
| Teal Golden Eagle | 0° | Hands ON | Hands ON | Hands ON | Hands ON | | 50% |
| | 90° | Hands ON | Hands ON | Hands ON | X | | 37.5% |
| | 180° | Hands ON | Hands ON | Hands ON | | | 37.5% |
| | 270° | Hands ON | Hands ON | Hands ON | | | 37.5% |
| | Success per wind speed | 50% | 50% | 50% | 12.5% | 0% | --- |
| Vantage Robotics Vesper | 0° | Hands OFF | Hands OFF | Hands OFF | Hands ON | | 87.5% |
| | 90° | Hands OFF | Hands OFF | Hands ON | X | | 62.5% |
| | 180° | Hands OFF | Hands OFF | Hands ON | | | 62.5% |
| | 270° | Hands OFF | Hands OFF | Hands ON | | | 62.5% |
| | Success per wind speed | 100% | 100% | 62.5% | 12.5% | 0% | --- |

*Table 3. Station Keeping In Wind test results. *Note: success per orientation is only calculated using results up to 25 mph because no systems were able to perform successfully beyond that wind speed.*

# Land and Takeoff In Wind

## Test Method

**Purpose**

This test method is used to evaluate the impact of wind on a sUAS' ability to land and takeoff using induced wind conditions of varying speed and direction.

**Summary of Test Method**

The sUAS attempts to takeoff and land while within wind flow of varying speed and the orientation at which the wind is pushing against the system. Performance is evaluated as either success or failure, indicated by whether or not the system was able to takeoff and land without crashing. After takeoff segment, the sUAS must be able to remain hovering in the air within the wind flow for 5 seconds without crashing, at which point the operator can then move onto the landing segment. As a controllable wind source, a wind generator is used for consistent wind and the sUAS is positioned below the wind flow such that when it launches it will enter the wind flow. When landing, it starts within the wind flow before engaging in its landing sequence. At each desired wind speed, the sUAS is oriented either 0°, 90°, 180°, or 270° to the wind flow. See Figure 17.

This test builds off of the **Land and Takeoff** test method (ASTM WK85838[5]), which specifies conditions for the ground plane material (high, medium, and low grip; e.g., expanded steel, oriented strand board, and artificial grass/turf, respectively), angle (pitched 0 – 90° in 5° increments), and orientation or pitch (pitch peak, pitch valley, or roll), as well as horizontal and vertical confinement. This test method induces wind impacts into these conditions. The results of this test indicate the types of wind conditions that the sUAS should be expected to be able to survive in with or without operator input to maintain its position and remain in the air.

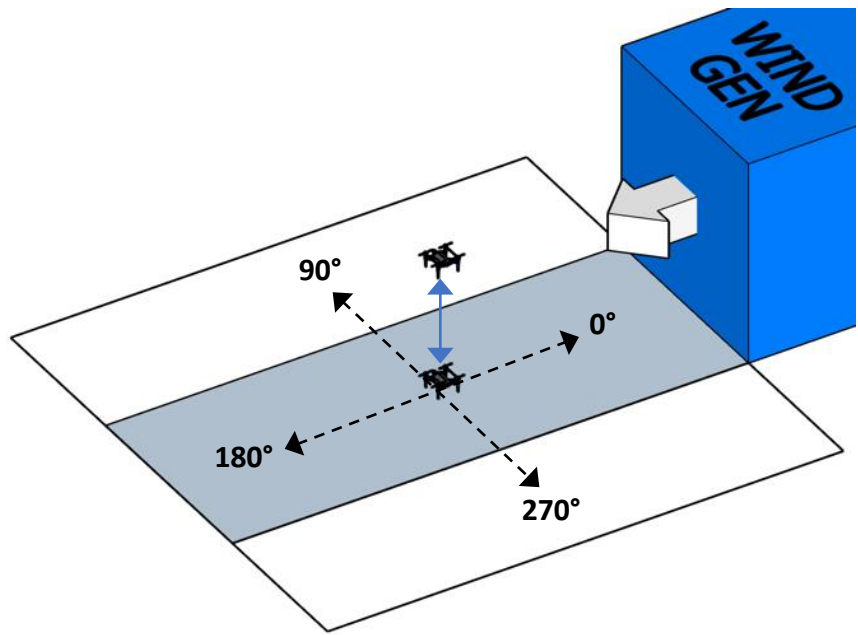

*Figure 17. The Land and Takeoff In Wind test method.*

---

[5] ASTM International. ASTM WK85838: New Test Method for Evaluating Aerial Response Robot Capabilities: Takeoff and Land. Registered work item currently under development.

## Apparatus and Artifacts

A takeoff and landing platform is positioned below the wind flow such that when the sUAS launches from it or lands on it, the system will pass through the wind flow.

## Equipment

A wind generator is used to induce wind flow for testing. Ideally, a system that generates laminar flow with controllable wind speed is used. A wind meter should be used to confirm the wind speed being generated before evaluating the sUAS' ability to keep station. Safety nets should be positioned around the testing area to catch the system should it become unstable during testing.

A timer is used to count down from 5 seconds after the sUAS is attempting to remain in the air after takeoff.

## Metrics

- Wind speed: The speed of the wind flow used during a test trial. Reported in miles per hour (mph) in increments of 5 mph.
- Orientation: The sUAS orientation to the wind flow during a test trial, either 0°, 90°, 180°, or 270°.
- Success: The number of trials that were successful, where 1 trial = 1 orientation and wind speed combination, and success means both takeoff and landing were able to be performed without crashing. Each successfully performed orientation = 25% and X = 0%, for a possible total of 100% per orientation and wind speed combination. Reported as a percentage (%).

## Procedure

1. Position the sUAS on the platform such that it is underneath and oriented 0° to where the wind flow will be once it is actively flowing.
2. Turn on the wind generator to the desired wind speed.
3. Instruct the operator to attempt launching the sUAS.
4. Once the sUAS has finished taking off, use the timer to countdown 5 seconds.
5. If takeoff is successful, instruct the operator to attempt landing the sUAS back down through the wind flow. The system is not required to land on the platform, but it must pass through the wind flow on descension. If takeoff is not successful, mark the trial as failed and instruct the operator to disarm the sUAS.
6. If landing is successful, instruct the operator to disarm the sUAS. If landing is unsuccessful, mark the trial as failed and instruct the operator to disarm the sUAS.
7. Reposition the sUAS on the platform and rotate it 90° to the wind flow.
8. Repeat steps 3-7 until all four orientations have been completed, then increase the wind speed by 5 mph.
9. Repeat step 8 until all desired wind speeds have been tested.

## Test Results

Benchmarking was conducted at the KRI ECUAS Lab using the wind profile generator (WPG) in the outdoor netted area (see Figure 18 and Figure 19). Test results are derived from benchmarking conducted in November 2023.

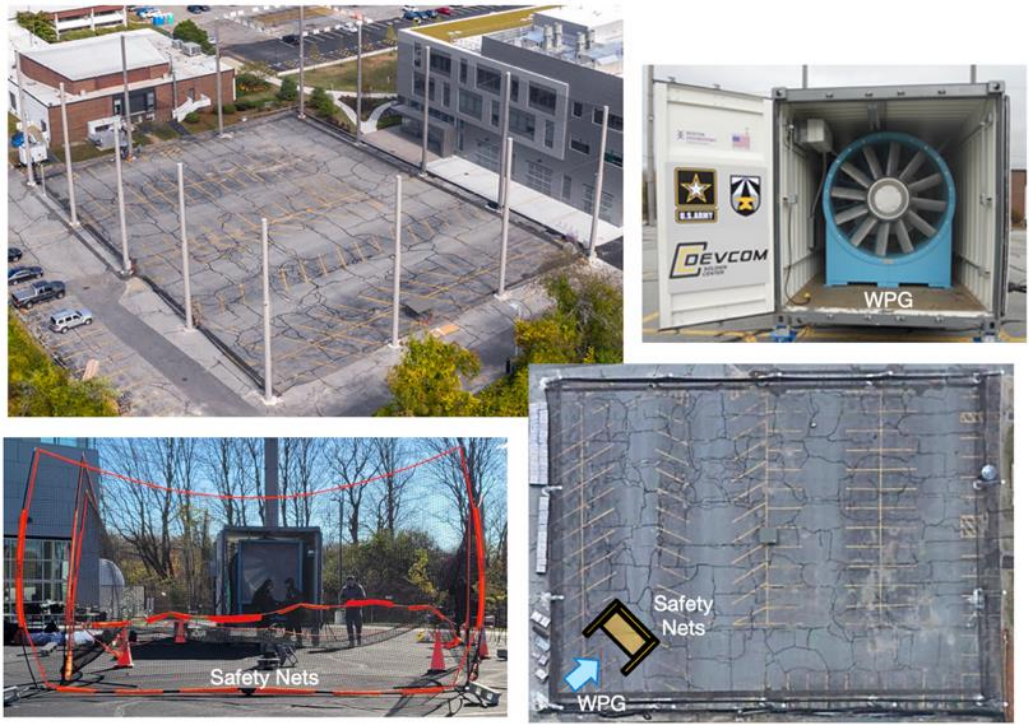

*Figure 18. The outdoor netted area and wind profile generator (WPG) at the KRI ECUAS Lab with safety nets used for testing.*

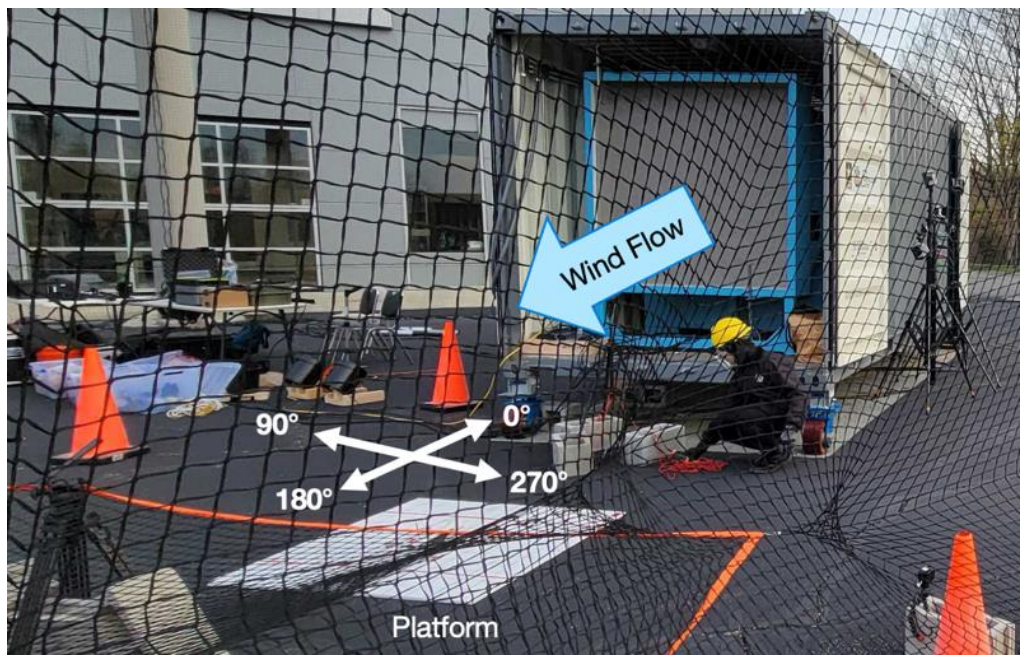

*Figure 19. Experimental set-up for the Land and Takeoff In Wind test method at the KRI ECUAS Lab.*

Test settings as per ASTM WK85838:

- Ground plane material: medium grip (pavement)
- Ground plane angle: 0°
- No confinement

| sUAS | Orientation / Metrics | Wind Speed | | | Success per orientation* |
|---|---|---|---|---|---|
| | | 15 mph | 20 mph | 25 mph | |
| Parrot ANAFI USA GOV | 0° | Takeoff and Land | Takeoff and Land | Takeoff and Land | 100% |
| | 90° | Takeoff and Land | Takeoff and Land | Takeoff and Land | 100% |
| | 180° | Takeoff and Land | Takeoff and Land | Takeoff and Land | 100% |
| | 270° | Takeoff and Land | Takeoff and Land | Takeoff and Land | 100% |
| | Success per wind speed | 100% | 100% | 100% | - |
| Skydio X2D | 0° | Takeoff and Land | Takeoff and Land | Takeoff and Land | 100% |
| | 90° | Takeoff and Land | Takeoff and Land | Takeoff and Land | 100% |
| | 180° | Takeoff and Land | Takeoff and Land | Takeoff and Land | 100% |
| | 270° | Takeoff and Land | Takeoff and Land | Takeoff and Land | 100% |
| | Success per wind speed | 100% | 100% | 100% | - |
| Teal Golden Eagle | 0° | Takeoff and Land | Takeoff and Land | X | 67% |
| | 90° | Takeoff and Land | Takeoff and Land | | 67% |
| | 180° | Takeoff and Land | Takeoff and Land | | 67% |
| | 270° | Takeoff and Land | Takeoff and Land | | 67% |
| | Success per wind speed | 100% | 100% | 0% | - |
| Vantage Robotics Vesper | 0° | Takeoff and Land | Takeoff and Land | X | 67% |
| | 90° | Takeoff and Land | Takeoff and Land | | 67% |
| | 180° | Takeoff and Land | Takeoff and Land | | 67% |
| | 270° | Takeoff and Land | Takeoff and Land | | 67% |
| | Success per wind speed | 100% | 100% | 0% | - |

*Table 4. Land and Takeoff In Wind test results. *Note: success per orientation is only calculated using 33% per each successfully sUAS orientation and wind speed pairing up to 25 mph because no systems were able to perform successfully beyond that wind speed.*

# Perched Field of Regard

## Test Method

**Purpose**

This test method evaluates the field of regard (FOR) of a sUAS, which is its field of view (FOV) as enabled by camera gimbal movement, while perched at elevation (e.g., looking down an alley from a rooftop), a common mission operation referred to as "perch and stare."

**Summary of Test Method**

The field of regard (FOR) for a sUAS refers to the field of view (FOV) of its camera(s) plus the additional horizontal or vertical range enabled by its gimbal. For example, if a camera with a vertical FOV of 70° is attached to a gimbal that is able to pitch up 45° and down 90° degrees, then the vertical FOR of the sUAS equals 80° above horizon (vertical FOV 70°/2 + gimbal up range 45°) and 125° below horizon (vertical FOV 70°/2 + gimbal down range of 90°). The FOR metric is useful on its own, but it lacks operational context for a given mission. In this test method, a system is perched at a given height overlooking a defined volume of space and its ability to inspect targets within its FOR in that space is evaluated. The height of the sUAS perch and the layout of targets (which corresponds to the boundaries of the space the system overlooks) is variable based on desired mission parameters. Targets are positioned on the ground and walls of the space at distances that correspond with every 10° of gimbal movement projected from the perch position onto the ground. See Figure 20 for an example test set-up.

The distance of the sUAS to the edge of the platform on which it is perched will impact its ability to inspect targets in the space below it, so for testing purposes the system is positioned as close to the edge as its ground contacts will allow (i.e., the ideal ledge perch position). It is assumed that the sUAS being tested is likely not able to accurately and reliably land and perch in this position (it at all), so the results of this test represent a best-case scenario for perched field of regard. The visual acuity able to be achieved while perched and the coverage of ground and wall targets are the primary performance metrics.

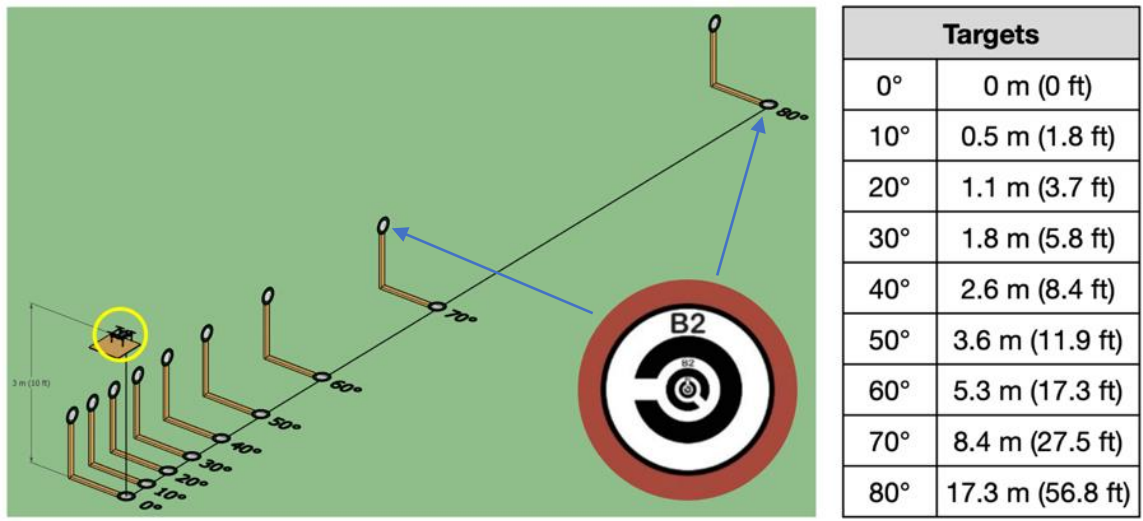

| Targets | |
|---|---|
| 0° | 0 m (0 ft) |
| 10° | 0.5 m (1.8 ft) |
| 20° | 1.1 m (3.7 ft) |
| 30° | 1.8 m (5.8 ft) |
| 40° | 2.6 m (8.4 ft) |
| 50° | 3.6 m (11.9 ft) |
| 60° | 5.3 m (17.3 ft) |
| 70° | 8.4 m (27.5 ft) |
| 80° | 17.3 m (56.8 ft) |

*Figure 20. Example experimental set-up for the Perched Field of Regard test with the sUAS elevated 3 m [10 ft] high with visual acuity targets (example shown in center) positioned every 10° at the distances shown on the right.*

## Apparatus and Artifacts

Visual acuity targets consist of five nested Landolt C optotypes of varying sizes and orientations; the opening of each C corresponds to a different level of detail that, if the orientation of the C opening is able to be correctly identified, correspond to different levels of detail able to be visually resolved. The dimensions of a Landolt C are described using a standard unit (t); see ASTM E2566[6] for more information. For the visual acuity targets used in this test method, from the largest to the smallest C, the size of t = 20, 8, 3, 1.3, and 0.5 mm [0.8, 0.3, 0.125, 0.05, and 0.02 in]. See Figure 21.

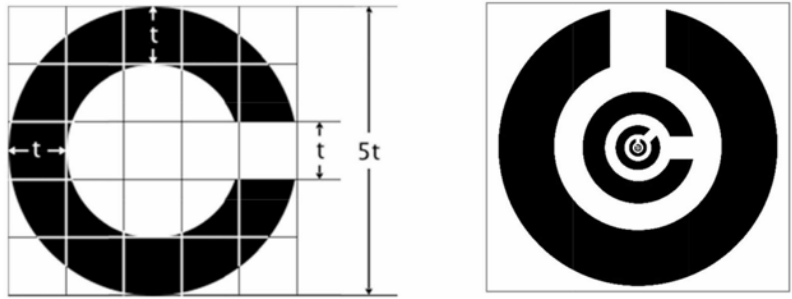

*Figure 21. <u>Left</u>: Dimensional configuration of a Landolt C using a standard unit (t). <u>Right</u>: Example nested Landolt C configuration from a visual acuity target. In this example, from the largest to the smallest C, the directions of the C openings are top, right, top-right, top, and top-left.*

For evaluating perched field of regard in a lab setting, the visual acuity targets are mounted onto two wooden posts: one that sits on the ground and the other that is 90° upright. The length of each post can vary based on the scenario being evaluated. For a basic test that simulates perching above and looking down a corridor, both posts can measure 1.5 m [5 ft], equating to a corridor that is 3 m [10 ft] wide and 3 m [10 ft] tall where targets are mounted down the center of the corridor ground and halfway up the corridor walls; see Figure 22. If evaluating in a real-world setting, the targets can be mounted in the environment down the middle of a space and on the walls; see Figure 23 for an example of embedding this test method a corridor of similar dimensions (3 x 3 m [10 x 10 ft]) in a standard squad and platoon task and technique trainer (Station 2) as specified in TC 90-1[7].

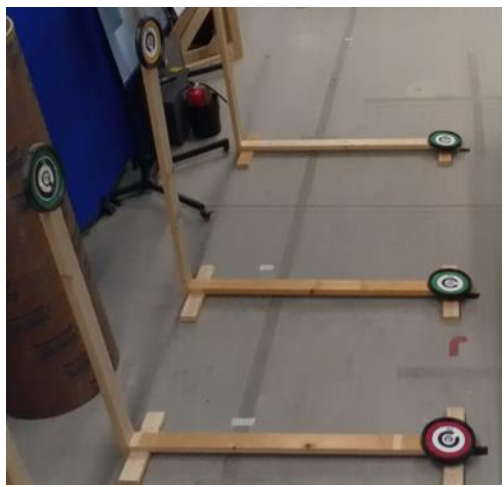

*Figure 22. Visual acuity targets mounted on wooden posts for the Perched Field of Regard test method.*

---

[6] ASTM International. ASTM E2566/E2566M – 24: Standard Test Method for Evaluating Response Robot Sensing: Visual Acuity. ASTM International Book of Standards Volume 15.08, DOI: 10.1520/E2566_E2566M-24.
[7] Headquarters, Department of the Army. TC 90-1: Training for Urban Operations. May 2008.

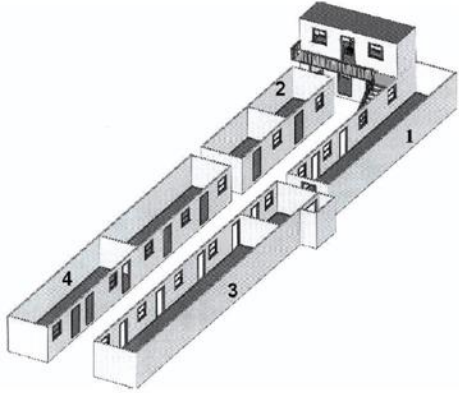

Standard squad and platoon task and technique trainer, Station 2 [TC 90-1[8]]

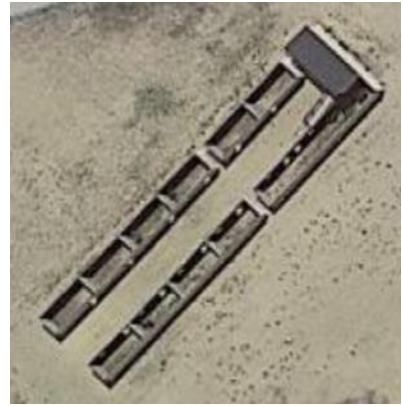

Station 2 at Fort Devens: Facility 12 Urban Assault Course

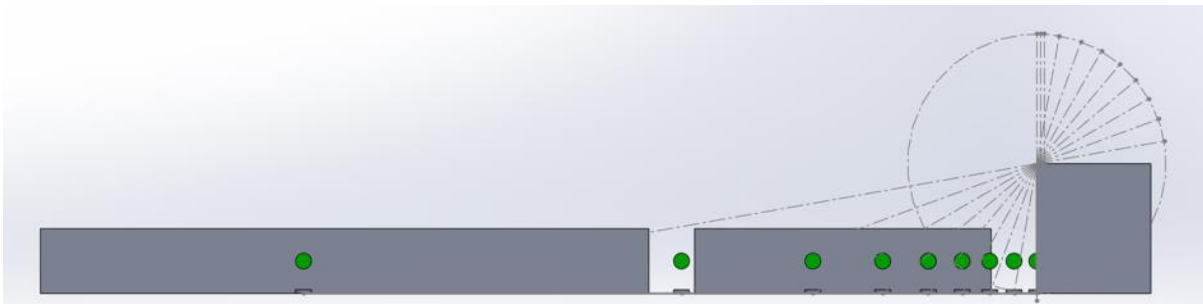

Side layout of Perched Field of Regard test in Station 2

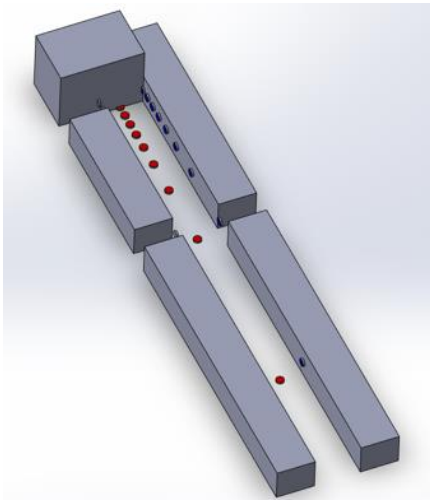
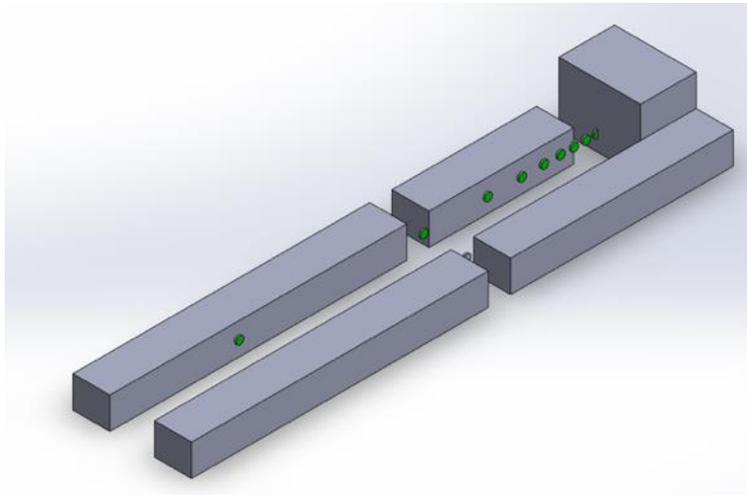

Angled views of Perched Field of Regard test in Station 2

*Figure 23. Example of embedding the Perched Field of Regard test within a standard squad and platoon task and technique trainer, Station 2.*

---

[8] Headquarters, Department of the Army. TC 90-1: Training for Urban Operations. May 2008.

The elevated platform where the sUAS will be perched can be set to any height that is operationally relevant for a given mission. For lab testing, it is recommended that perch height be set to 3 m [10 ft]. See Figure 24 for the target layout positions when using this perch height, which correspond to every 10° of gimbal camera movement downrange along centerline from the perch position.

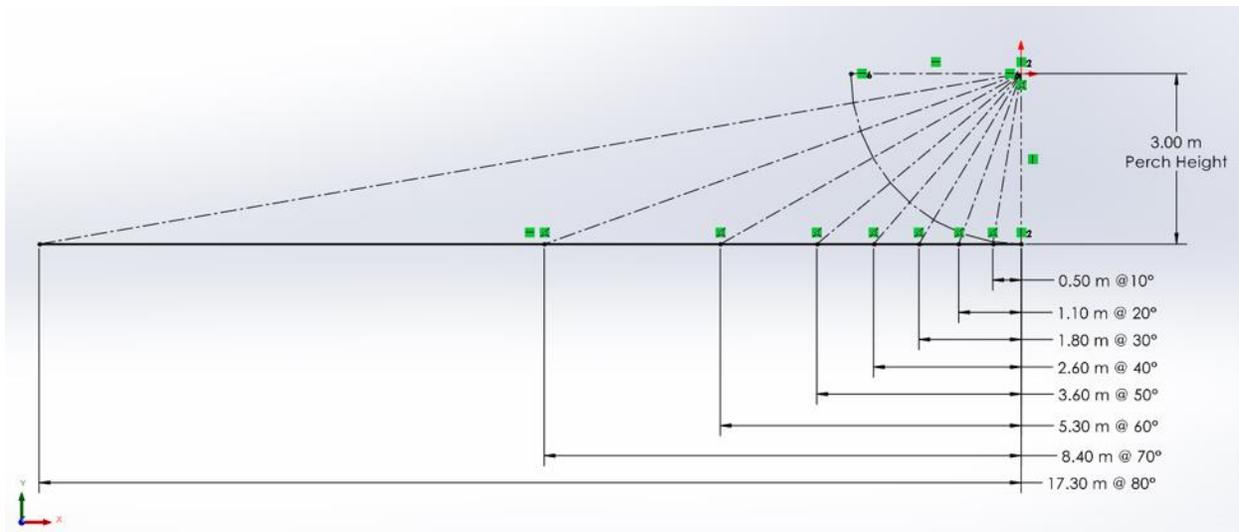

*Figure 24. Positions of the sUAS (perch height of 3 m [10 ft]) and each of the targets placed every 10° from 0 – 80°.*

## Equipment

No additional equipment is required to conduct this test.

## Metrics

- Visual acuity: Level of detail that can be resolved in the available Landolt C artifacts during the test, reported per visual acuity target able to be seen. These measures are split into targets on the ground and targets on the wall. Reported in millimeters (mm), smaller is better, per target and as an average.
- Coverage: The range of targets able to be inspected while perched, split into ground target coverage and wall target coverage. Reported in degrees (°) for the angle of the target placement in relation to the sUAS and in meters (m) for the distance of the target placement in relation to the sUAS.

## Procedure

1. Select desired perch height and corridor width then set-up the test apparatus accordingly, with visual acuity targets positioned every 10° along the corridor ground and walls.
2. Position the sUAS on the perch platform with its ground contacts as close to the edge.
3. Instruct the operator to power on and connect to the sUAS.
4. Instruct the operator to manipulate the camera gimbal, zoom, white balance, etc., to inspect as many of the targets as possible, reporting the directions of the Landolt C openings to indicate what level of detail is able to be resolved in each target, until all possible targets have been inspected.

## Test Results

Benchmarking was conducted at the UMass Lowell NERVE Center with the sUAS at an elevation of 3 m [10 ft] with targets placed at 0, 0.5, 1.1, 1.8, 2.6, 3.6, 5.3, 8.4, and 17.3 m [0, 1.8, 3.7, 5.8, 8.4, 11.9, 17.3, 27.5, and 56.8 ft], which is every 10° from 0 – 80°, as shown in Figure 24. Test results are derived from benchmarking conducted in December 2023. See Figure 25 and Figure 26 for photos from the test session.

A summary of the test results is provided in Figure 27 followed by detailed results. Achieving a wider range of distance (i.e., more targets able to be inspected while perched) at higher acuity (i.e., smaller details able to be resolved) is desirable.

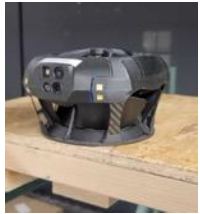
Cleo Robotics Dronut X1P

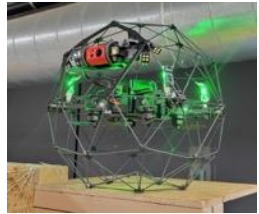
Flyability Elios 2 GOV

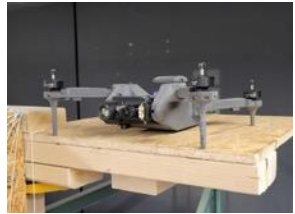
Lumenier Nighthawk V3

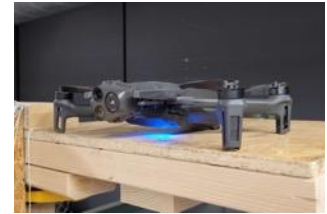
Parrot ANAFI USA GOV

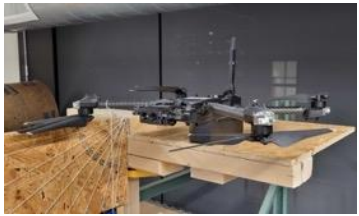
Skydio X2D

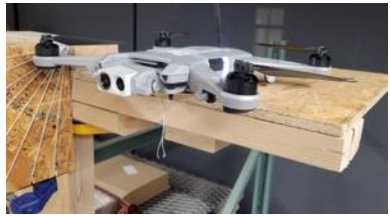
Teal Golden Eagle

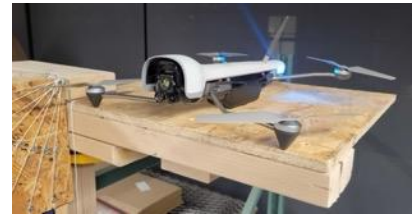
Vantage Robotics Vesper

*Figure 25. Each sUAS perched on the elevated platform as far to the edge as their ground contacts would allow.*

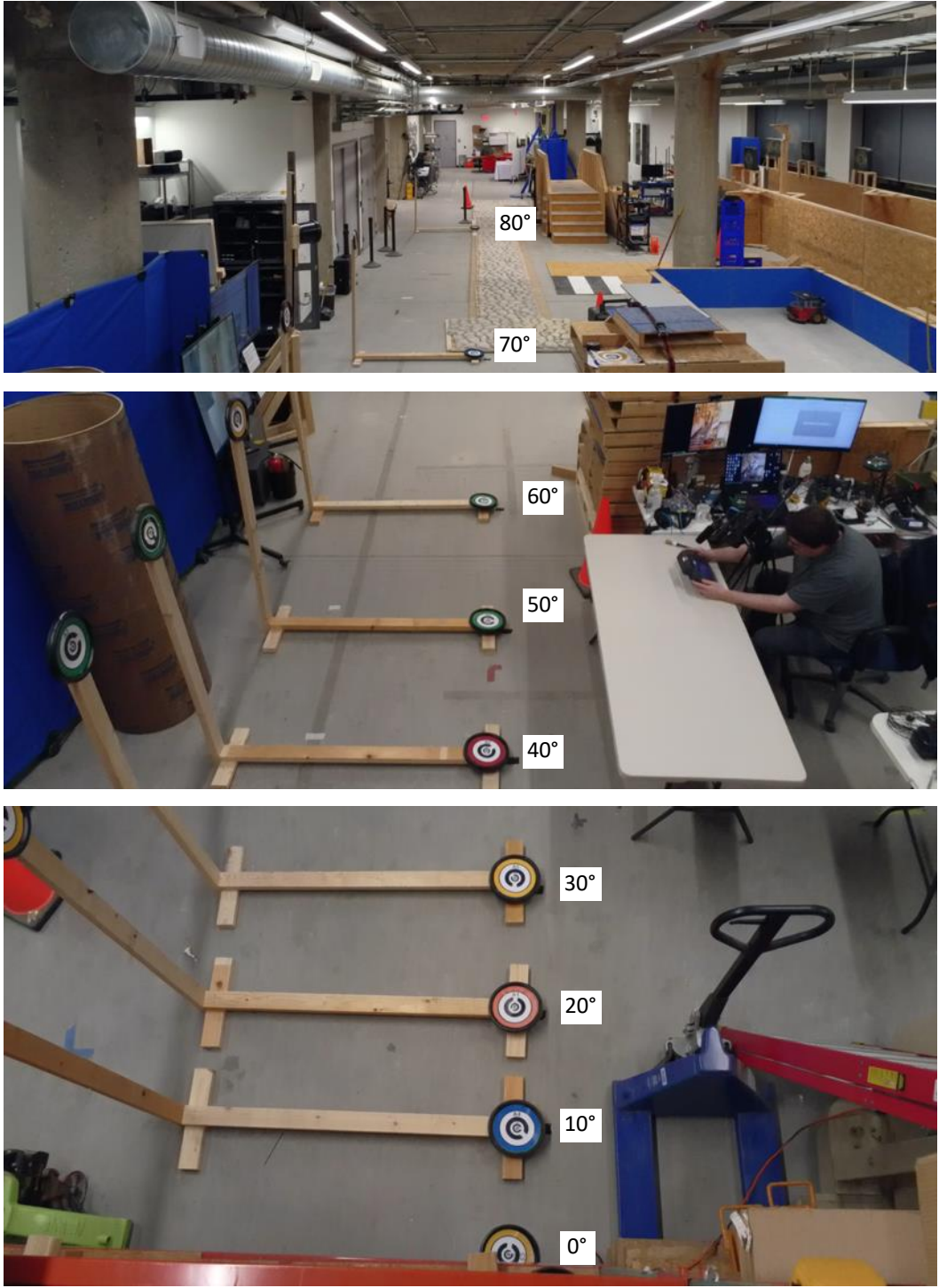

*Figure 26. Example views of the Perched Field of Regard test as seen by the Parrot ANAFI USA GOV camera.*

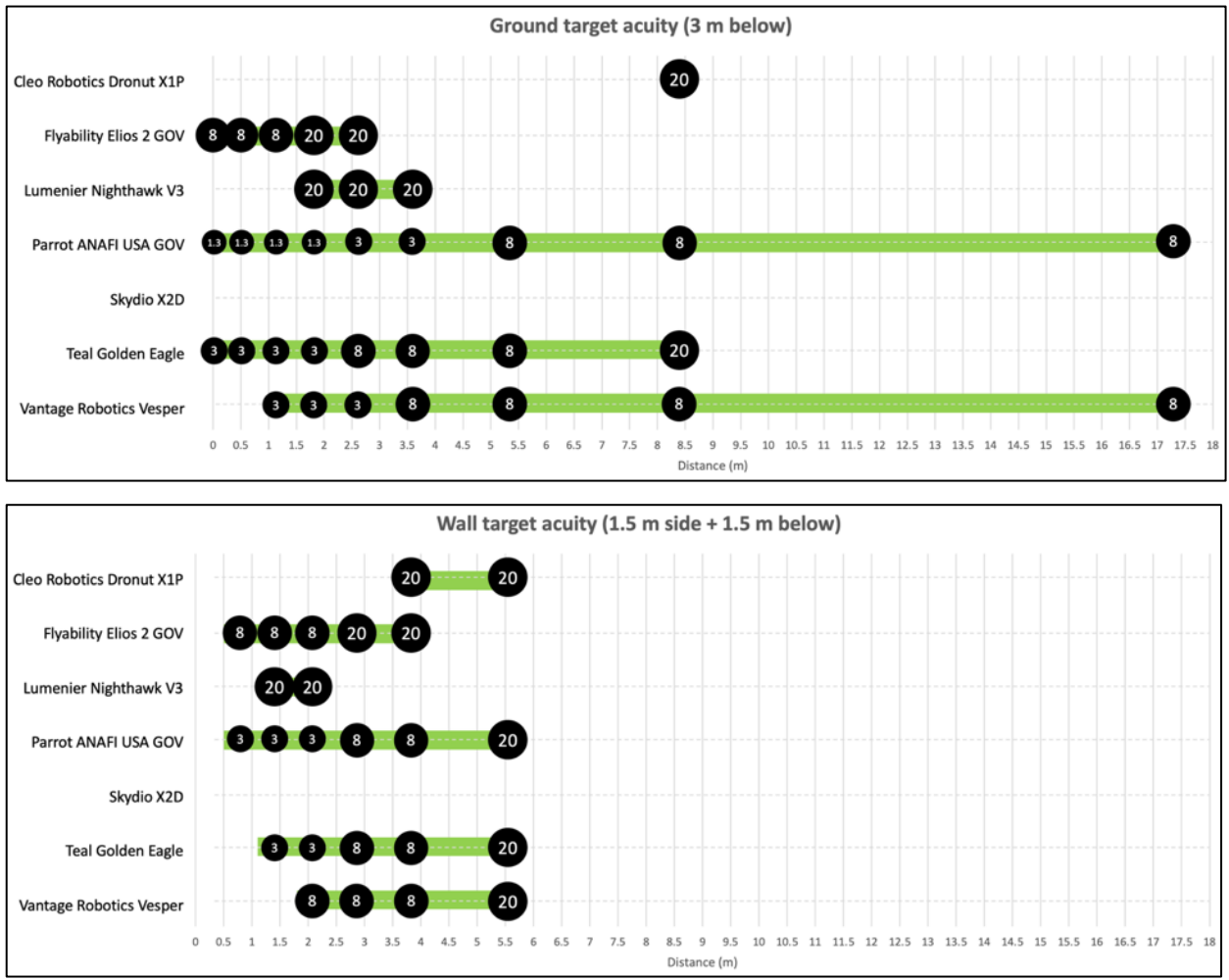

*Figure 27. Summarized Perched Field of Regard test results. The level of detail able to be resolved in the visual acuity target at each distance is noted inside each dot in millimeters.*

| sUAS | Metrics | Target | | | | | | | | | Avg |
|---|---|---|---|---|---|---|---|---|---|---|---|
| | | 0°<br>0 m | 10°<br>1.5 m | 20°<br>1.1 m | 30°<br>1.8 m | 40°<br>2.6 m | 50°<br>3.6 m | 60°<br>5.3 m | 70°<br>8.4 m | 80°<br>17.3 m | |
| Cleo Robotics Dronut X1P | Ground target acuity (mm) | N/A | N/A | N/A | N/A | N/A | N/A | N/A | 20 | - | 20 |
| | Ground target coverage | 11%: 70°, 8.4 m | | | | | | | | | |
| | Wall target acuity (mm) | N/A | N/A | N/A | N/A | N/A | 20 | 20 | - | - | 20 |
| | Wall target coverage | 22%: 50 – 60°, 3.6 – 5.3 m | | | | | | | | | |
| Flyability Elios 2 GOV | Ground target acuity (mm) | 8 | 8 | 8 | 20 | 20 | - | - | - | - | 12.8 (+/- 6.6) |
| | Ground target coverage | 56%: 0 – 40°, 0 – 2.6 m | | | | | | | | | |
| | Wall target acuity (mm) | N/A | 8 | 8 | 8 | 20 | 20 | - | - | - | 12.8 (+/- 6.6) |
| | Wall target coverage | 56%: 10 – 50°, 1.5 – 3.6 m | | | | | | | | | |
| Lumenier Nighthawk V3 | Ground target acuity (mm) | N/A | N/A | N/A | 20 | 20 | 20 | - | - | - | 20 |
| | Ground target coverage | 33%: 30 – 50°, 1.8 – 3.6 m | | | | | | | | | |
| | Wall target acuity (mm) | N/A | N/A | 20 | 20 | - | - | - | - | - | 20 |
| | Wall target coverage | 22%: 20 – 30°, 1.1 – 1.8 m | | | | | | | | | |
| Parrot ANAFI USA GOV | Ground target acuity (mm) | 1.3 | 1.3 | 1.3 | 1.3 | 3 | 3 | 8 | 8 | 8 | 3 (+/- 3.1) |
| | Ground target coverage | 100%: 0 - 80°, 0 – 17.3 m | | | | | | | | | |
| | Wall target acuity (mm) | N/A | 3 | 3 | 3 | 8 | 8 | 20 | - | - | 7.5 (+/- 6.6) |
| | Wall target coverage | 67%: 10 – 60°, 1.5 – 5.3 m | | | | | | | | | |
| Skydio X2D* | Ground target acuity (mm) | N/A | N/A | N/A | N/A | N/A | N/A | N/A | - | - | - |
| | Ground target coverage | 0% | | | | | | | | | |
| | Wall target acuity (mm) | N/A | N/A | N/A | N/A | N/A | N/A | N/A | - | - | - |
| | Wall target coverage | 0% | | | | | | | | | |
| Teal Golden Eagle | Ground target acuity (mm) | 3 | 3 | 3 | 3 | 8 | 8 | 8 | 20 | - | 7 (+/- 5.8) |
| | Ground target coverage | 89%: 0 – 70°, 0 – 8.4 m | | | | | | | | | |
| | Wall target acuity (mm) | N/A | N/A | 3 | 3 | 8 | 8 | 20 | - | - | 8.4 (+/- 6.9) |
| | Wall target coverage | 56%: 20 – 60°, 1.1 – 5.3 m | | | | | | | | | |
| Vantage Robotics Vesper | Ground target acuity (mm) | N/A | N/A | 3 | 3 | 3 | 8 | 8 | 8 | 8 | 5.9 (+/- 2.7) |
| | Ground target coverage | 78%: 20 – 80°, 1.1 – 17.3 m | | | | | | | | | |
| | Wall target acuity (mm) | N/A | N/A | N/A | 8 | 8 | 8 | 20 | - | - | 11 (+/- 6) |
| | Wall target coverage | 44%: 30 – 60°, 1.8 – 5.3 m | | | | | | | | | |

*Table 5. Perched Field of Regard test results. *Note: The Skydio X2D camera gimbal is not able to be manipulated while landed.*

# Exterior Building Clearing

## Test Method

**Purpose**

This test method evaluates the sUAS' capability at systematically visually inspecting the exterior of a building (i.e., "clearing" the exterior of the building).

**Summary of Test Method**

To run this test, a series of visual acuity targets in buckets are systematically positioned on and around a building to cover all sUAS inspection positions and angles that would be required to comprehensively inspect (or "clear") the exterior of the building. This is achieved by mounting visual acuity targets on the exterior (and interior, such that they can be seen from the exterior) of the building. The sUAS inspects all targets and metrics of coverage and visual acuity are evaluated. See Figure 28 for examples of target layouts on two types of buildings.

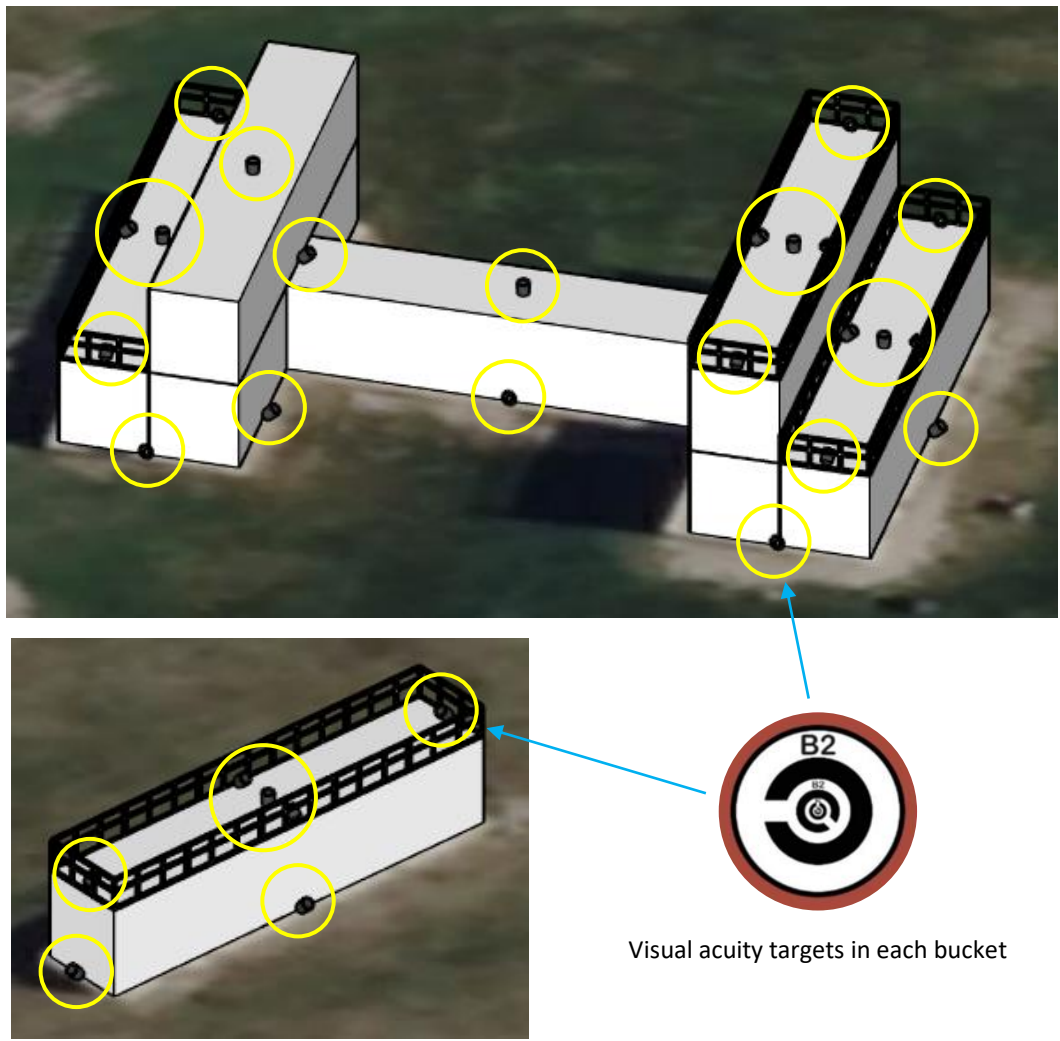

Visual acuity targets in each bucket

*Figure 28. Example layout of targets positioned on and around buildings to conduct the Exterior Building Clearing test.*

## Apparatus and Artifacts

Visual acuity targets consist of five nested Landolt C optotypes of varying sizes and orientations; the opening of each C corresponds to a different level of detail that, if the orientation of the C opening is able to be correctly identified, correspond to different levels of detail able to be visually resolved. The dimensions of a Landolt C are described using a standard unit (t); see ASTM E2566[9] for more information. For the visual acuity targets used in this test method, from the largest to the smallest C, the size of t = 20, 8, 3, 1.3, and 0.5 mm [0.8, 0.3, 0.125, 0.05, and 0.02 in]. See Figure 29.

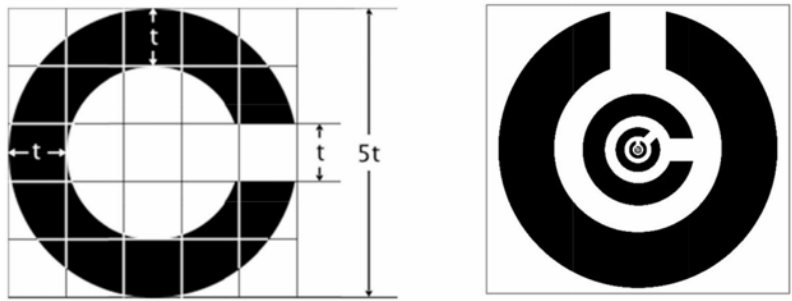

*Figure 29. <u>Left</u>: Dimensional configuration of a Landolt C using a standard unit (t). <u>Right</u>: Example nested Landolt C configuration from a visual acuity target. In this example, from the largest to the smallest C, the directions of the C openings are top, right, top-right, top, and top-left.*

The targets are mounted inside of 1-gallon buckets that measure approximately 20 cm$^3$ [8 in$^3$] which are then mounted to a wooden construction to orient the target appropriately and weigh it down to prevent it from being knocked over during testing. See Figure 30 for examples.

There are three types of targets that can be implemented:

1. <u>Upward-facing targets</u> (i.e., those that are viewable from above) are mounted on top of exposed horizontal surfaces/planes such as floors, roofs, catwalks, etc.
2. <u>Angled-facing targets</u> (i.e., those that are viewable from 45°) are mounted in the center of 90° intersections of horizontal and vertical surfaces/planes such as walls and floors, railings and floors, etc.
3. <u>Forward-facing targets</u> (i.e., those that are viewable from the front) are mounted on vertical surfaces/planes such as walls.

All types of targets can be mounted on the exterior of the building or on the interior such that they are viewable through an open door, window, or ceiling hatch. Only one target of each type is mounted against each distinct surface/intersection segment, with each segment delineated as it meets other surfaces/intersections. See Figure 31 for an abstract representation of target placement on exterior and interior surfaces.

A real-world environment is used with a building/structure with geometry that is representative of relevant mission parameters.

---

[9] ASTM International. ASTM E2566/E2566M – 24: Standard Test Method for Evaluating Response Robot Sensing: Visual Acuity. ASTM International Book of Standards Volume 15.08, DOI: 10.1520/E2566_E2566M-24.

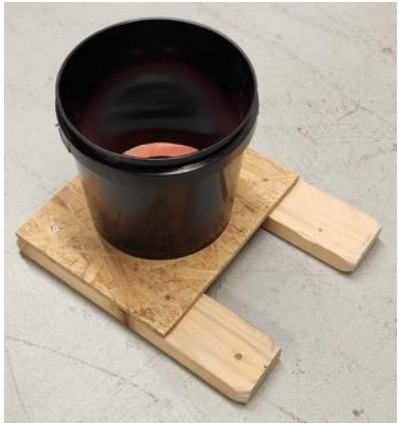
Upward-facing target (viewable from above)

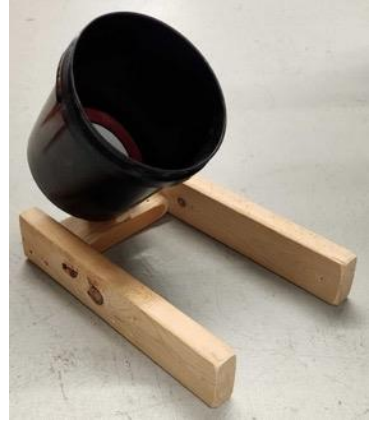
Angled target (viewable from 45°)

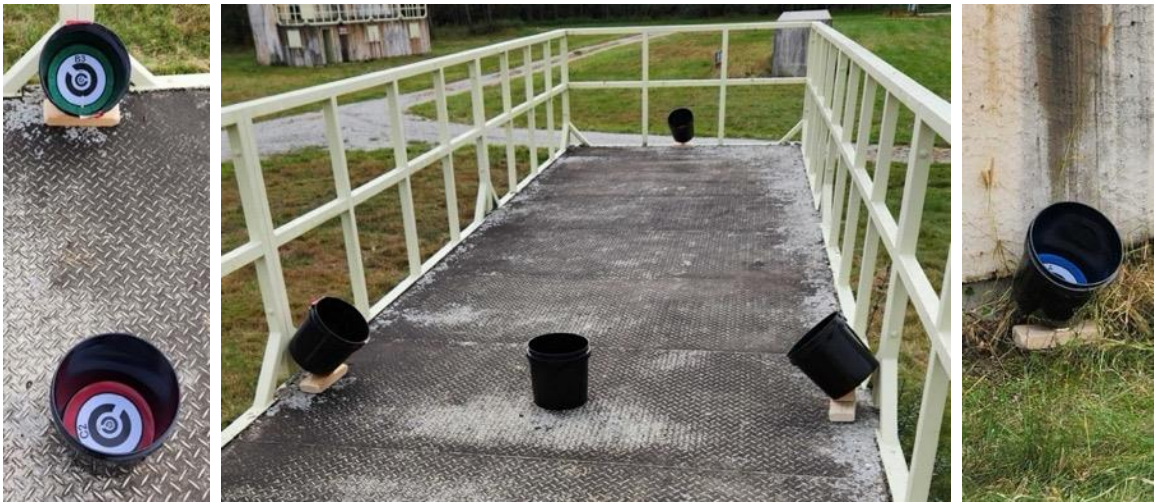
Targets mounted in a real-world environment

*Figure 30. Examples of targets used for the Exterior Building Clearing test method.*

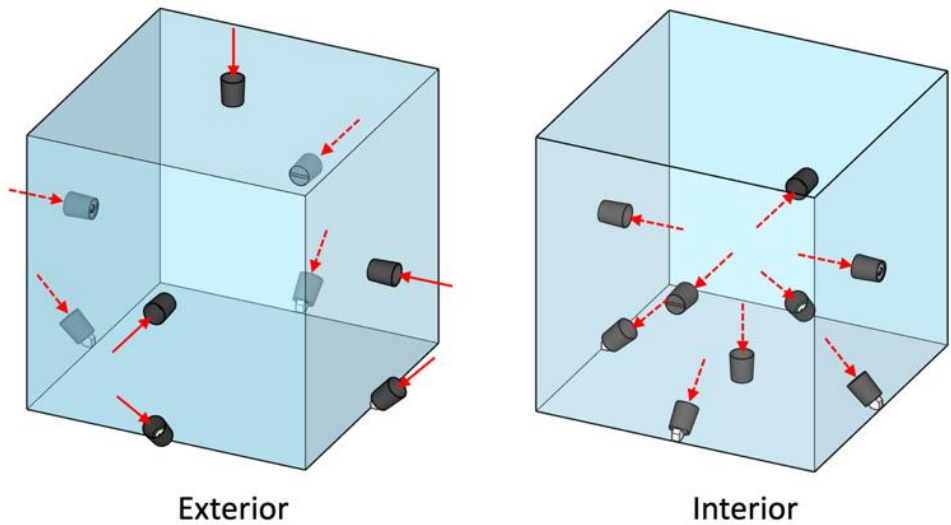

*Figure 31. All possible target placements on exterior and interior surfaces/intersections.*

**Equipment**

A timer is used to measure the duration of the test.

**Metrics**

- Duration: The amount of time between when the sUAS takes off to when it lands after inspecting all possible targets. Reported in minutes (min).
- Coverage: The number of targets able to be inspected compared to the number available. Reported as a percentage (%).
- Average visual acuity: Level of detail that can be resolved in the available Landolt C artifacts during the test, reported per visual acuity target able to be seen. Reported in millimeters (mm) as an average with standard deviation, smaller is better.

**Procedure**

1. Outfit the building that will be used for testing with upward-facing, angled-facing, and forward-facing targets as needed.
2. Instruct the operator to launch the sUAS and start the timer.
3. Instruct the operator to inspect as many of the targets as possible, reporting the directions of the Landolt C openings to indicate what level of detail is able to be resolved in each target, until they believe they have inspected all possible targets.
4. Once the operator believes they have inspected all possible targets, instruct them to fly the sUAS back to the start position and land.
5. Once the sUAS has landed, stop the timer.

## Test Results

Benchmarking was conducted at Fort Devens: Facility 15, Smithville, which consists of multiple structures made of connex containers in varying configurations (see Figure 32). Test results are derived from benchmarking conducted in October 2023, with tests conducted on a *simple* building structure and on a *complex* building structure. Only sUAS that are designed for outdoor operations were evaluated; however, the Vantage Robotics Vesper was not operational at testing time.

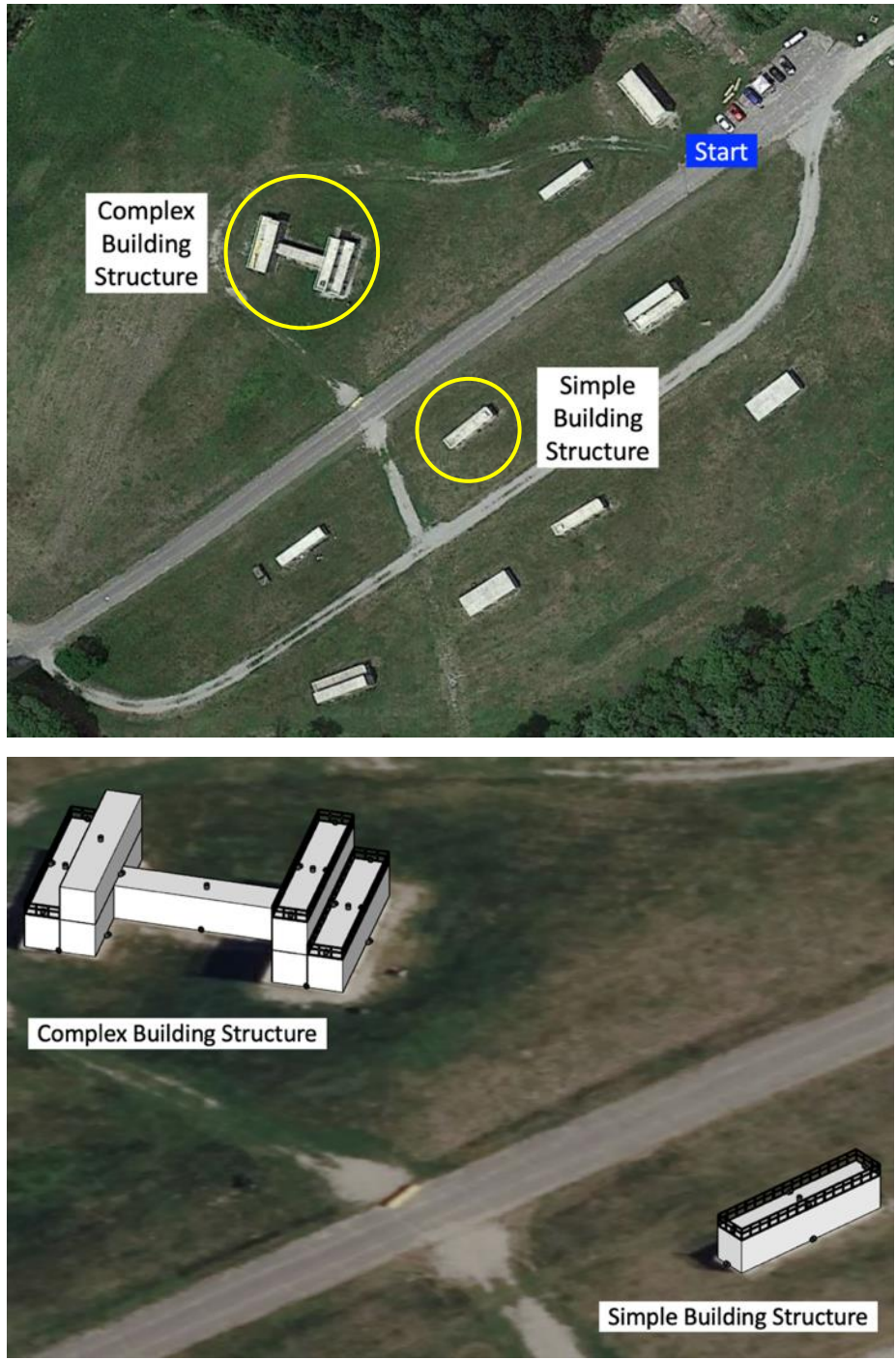

*Figure 32. Test site of the Exterior Building Clearing test method conducted at Fort Devens: Facility 15, Smithville.*

A summary of the test results is provided in Figure 33 followed by detailed results. Performing the test faster (i.e., shorter duration) and with higher acuity (i.e., smaller details able to be resolved) is desirable.

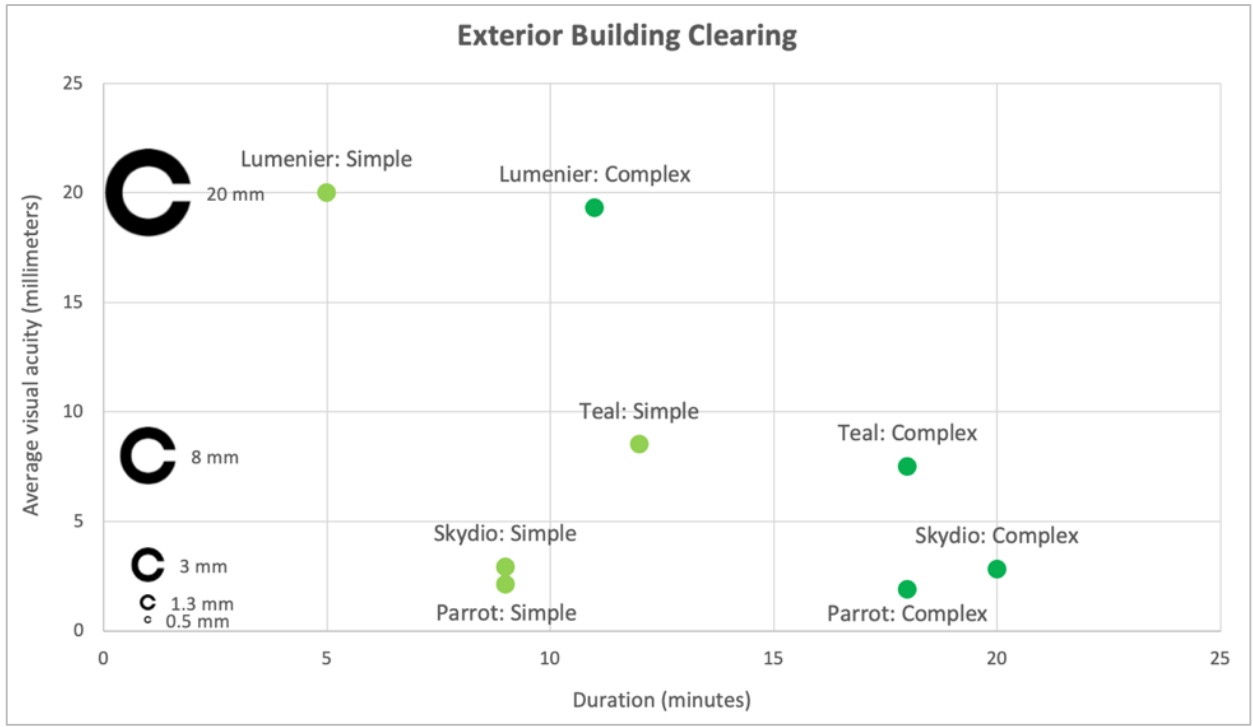

*Figure 33. Exterior Building Clearing test results. Note: Landolt Cs are plotted for reference but are not drawn to scale.*

## Simple Building Structure

This structure consists of a single story, two doors, a window, and a roof with a railing accessible via a hatch. A total of 13 targets were used: 8 angled-facing (placed at the edges where the wall meets the ground and where railing meets the roof/floor), 2 upward-facing (on top of the roof and inside the hatch), and 4 forward-facing (on the interior walls as viewable through doors and windows). See Figure 34.

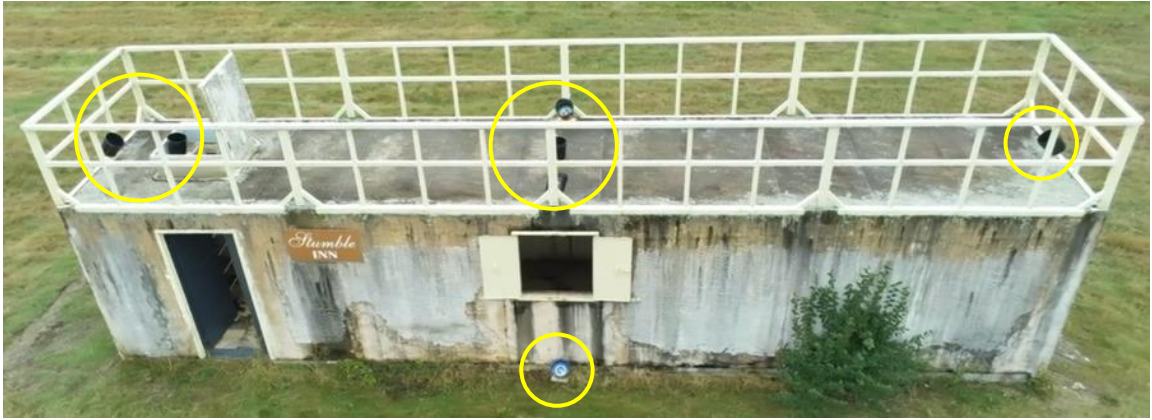

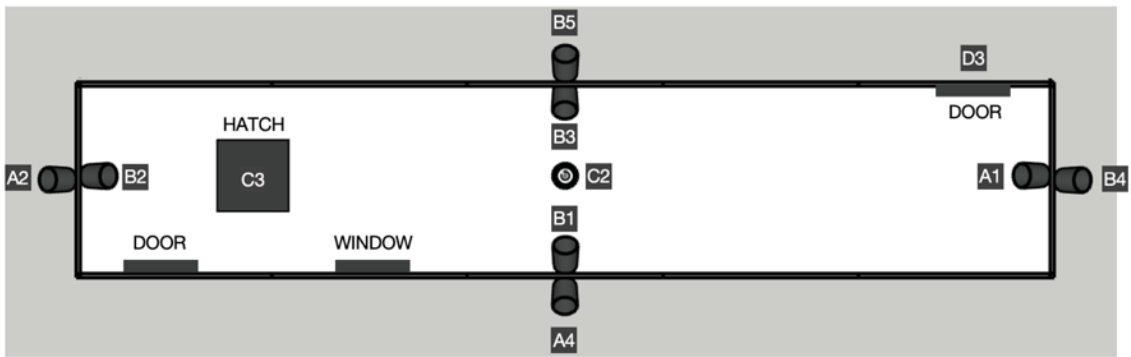

*Figure 34. Layout of the simple building structure used for the Exterior Building Clearing test.*

| sUAS | Metrics | Performance | Example Acuity Target |
|---|---|---|---|
| Lumenier Nighthawk V3 | Duration | 5 min | 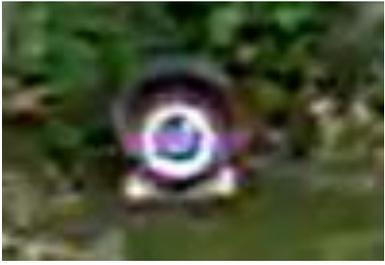 |
| | Coverage | 100% | |
| | Average visual acuity | 20 (+/- 0) mm | |
| Parrot ANAFI USA GOV | Duration | 9 min | 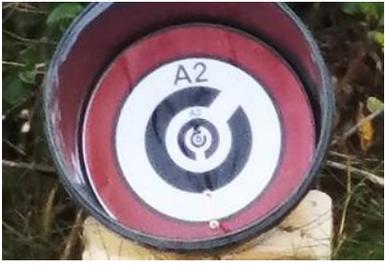 |
| | Coverage | 100% | |
| | Average visual acuity | 2.1 (+/- 0.9) mm | |
| Skydio X2D | Duration | 9 min | 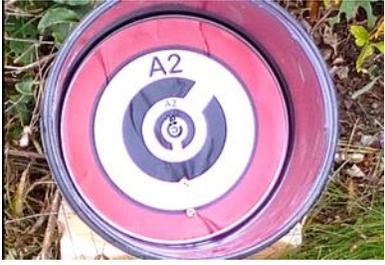 |
| | Coverage | 100% | |
| | Average visual acuity | 2.9 (+/- 0.5) mm | |
| Teal Golden Eagle* | Duration | 12 min | 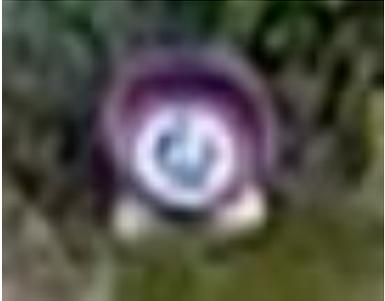 |
| | Coverage | 100% | |
| | Average visual acuity | 8.5 (+/- 3.7) mm | |

*Table 6. Exterior Building Clearing test results on the simple building structure. *Note: Screenshot from the Teal Golden Eagle recorded video is only from the zoomed out FPV camera as the system does not record any zooming performed by the operator during flight; i.e., the operator's view of the targets was much clearer through the OCU during flight.*

## Complex Building Structure

This structure consists of several interconnected connex containers up to two stories high, multiple doors and windows, with roof surfaces at one and two stories high accessible via a hatch or door, some with railings and some without. A total of 35 targets were used: 26 angled-facing (placed at the edges where the wall meets the ground and where railing/wall meets the roof/floor), 6 upward-facing (on top of the roof and inside the hatch), and 3 forward-facing (on the interior walls as viewable through doors and windows). See Figure 35.

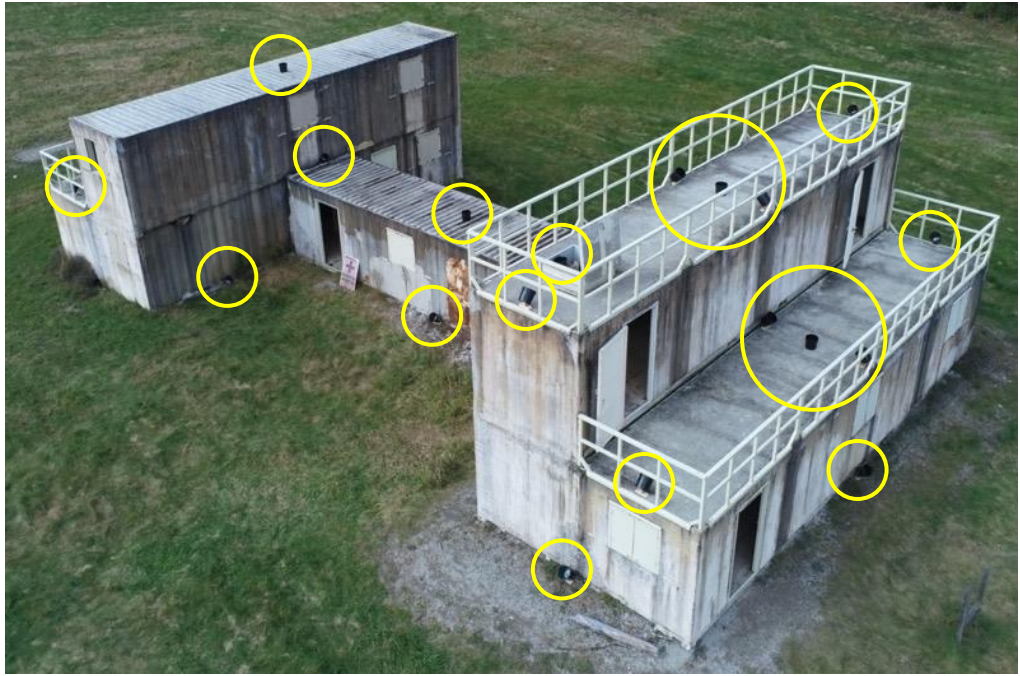

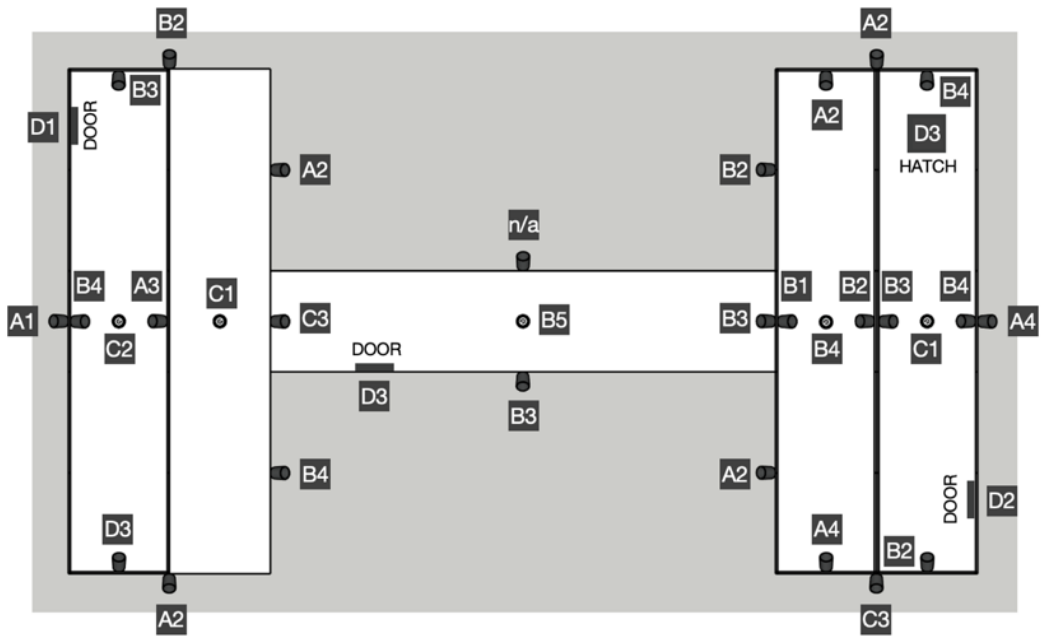

*Figure 35. Layout of the complex building structure used for the Exterior Building Clearing test.*

| sUAS | Metrics | Performance | Example Acuity Target |
|---|---|---|---|
| Lumenier Nighthawk V3 | Duration | 11 min | 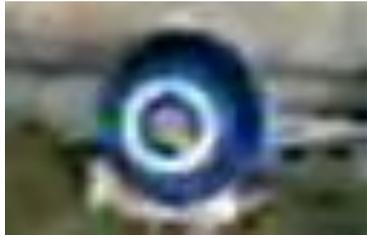 |
| | Coverage | 100% | |
| | Average visual acuity | 19.3 (+/- 2.9) mm | |
| Parrot ANAFI USA GOV | Duration | 18 min | 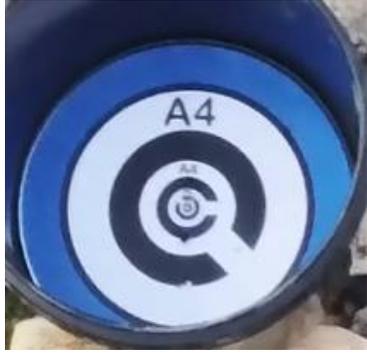 |
| | Coverage | 100% | |
| | Average visual acuity | 1.9 (+/- 0.8) mm | |
| Skydio X2D | Duration | 20 min | 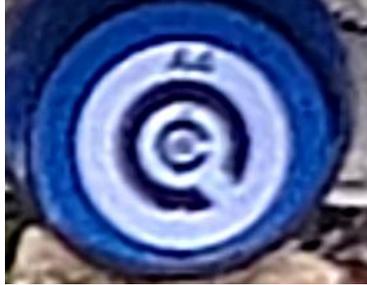 |
| | Coverage | 100% | |
| | Average visual acuity | 2.8 (+/- 0.6) mm | |
| Teal Golden Eagle* | Duration | 18 min | N/A |
| | Coverage | 91% | |
| | Average visual acuity | 7.5 (+/- 1.5) mm | |

Table 7. Exterior Building Clearing test results on the complex building structure. *Note: Recorded video of the Teal Golden Eagle test was corrupted so no screenshot is provided.

# Fly Through Confined Outdoor Spaces

## Test Method

**Purpose**

This test method evaluates sUAS capability to fly through confined spaces, such as alleyways and stairwells that are either entirely outdoors (GPS / GPS+VIO) or transition between outdoors and indoors (VIO).

**Summary of Test Method**

When navigating through a dense urban environment, a sUAS will need to fly between buildings and other structures. Doing so requires the ability to safely pass through horizontally confined spaces such as alleyways and corridors defined by the boundaries of the buildings and structures.

To run this test, the operator commands the sUAS to navigate through a confined outdoor space either by flying forward through the space (forward flight) or by ascending and descending through the space (ascension/descension). For each trial, the sUAS begins from a starting location that requires it to traverse in a direction not parallel to the flight path through the confined space, requiring the system to turn before flying through the confined space. Similarly, the end location for each trial also requires the sUAS to traverse in a direction not parallel to the navigation route. More simply, a single trial constitutes the sUAS traversing from the A side of the apparatus to the B side, navigating through the confined space, then traversing back over to the A side. Figure 36. Ideally, the sUAS will not collide with the boundaries of the space (i.e., walls, ground) during flight, but contact is allowed so long as it does not cause the system to crash in a way that requires human intervention for it to resume flight.

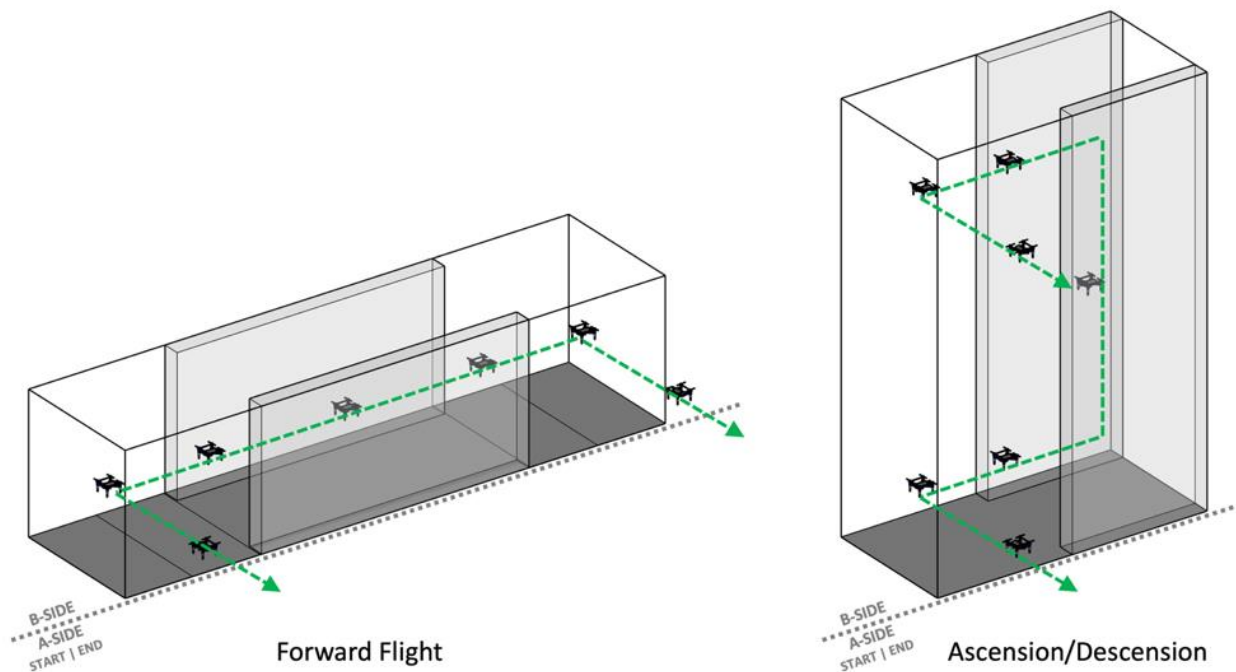

*Figure 36. Two configurations of the Fly Through Confined Outdoor Spaces test method.*

## Apparatus and Artifacts

While a fabricated apparatus can be used for this test, it is recommended that a real-world environment be used with a building/structure with geometry that is representative of relevant mission parameters. For example, a standard squad and platoon task and technique trainer (Station 2) as specified in TC 90-1[10] includes three widths of confined spaces for evaluating forward flight: corridor (3 m [10 ft] wide), passageway (2 m [6.6 ft] wide), and alleyway (1 m [3.3 ft] wide). See Figure 37.

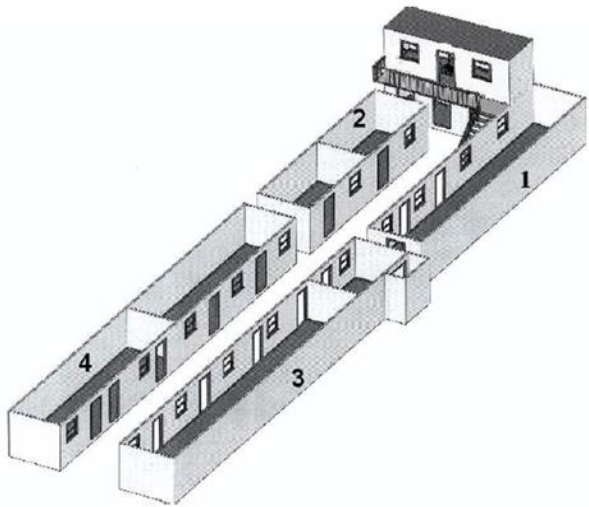
Standard squad and platoon task and technique trainer, Station 2 [TC 90-1[7]]

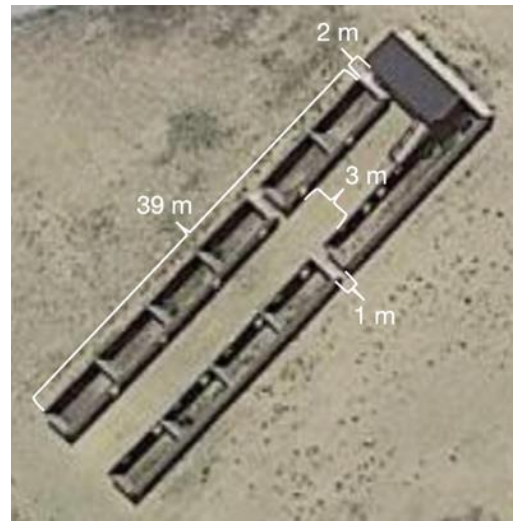
Station 2 at Fort Devens: Facility 12 Urban Assault Course

*Figure 37. Example of a real-world environment for running the Fly Through Confined Outdoor Spaces test: a standard squad and platoon task and technique trainer, Station 2.*

## Equipment

No additional equipment is required to conduct this test.

## Metrics

- Success: Whether or not the sUAS is able to successfully fly through the confined space, signified by crossing from the A side to the B side, through the confined space, and then back over to the A side on the opposite end of the apparatus. Each trial is reported as successful (✓) or failed (X).
- Collisions: Whether or not the sUAS collided with the apparatus boundaries. Reported as an integer.
- Completion: Reported as a percentage (%) of the number of successful trials compared to the number attempted.

---

[10] Headquarters, Department of the Army. TC 90-1: Training for Urban Operations. May 2008.

**Procedure**

1. Select the type of confined space flight that will be evaluated (forward flight or ascension/descension), measure the width of the confined space, and photograph the wall and ground/floor surface textures
2. Launch sUAS on the A side of the apparatus.
3. Command the sUAS to navigate to the B side of the apparatus. Once the full body of the sUAS crosses into the B side of the apparatus, start the timer.
4. Command the sUAS to fly through the confined space to the opposite end of the apparatus. If the sUAS crashes in a way that requires human intervention for it to resume flight, then that trial is considered a failure.
5. Command the sUAS navigate to side A of the apparatus, signifying the successful completion of a trial.
6. Repeat steps 3-5 until the desired number of successful trials has been achieved. Once the full body of the sUAS crosses into the A side of the apparatus on the final successful trial, stop the timer.

## Test Results

Benchmarking was conducted at Fort Devens: Facility 12, Urban Assault Course, which includes a standard squad and platoon task and technique trainer (Station 2) as specified in TC 90-1[11]. Three confined spaces were designated for testing: alleyway (1 m [3.3 ft] wide), passageway (2 m [6.6 ft] wide), and corridor (3 m [10 ft] wide). See Figure 38 and Figure 39. Each sUAS attempted to fly through these spaces in a circuit (alleyway, open area, passageway, corridor, repeat) for three trials each. If a system crashed and was not able to recover without manual intervention (e.g., the operator picking up the system, repairing it, and then resuming flight), then the test was stopped.

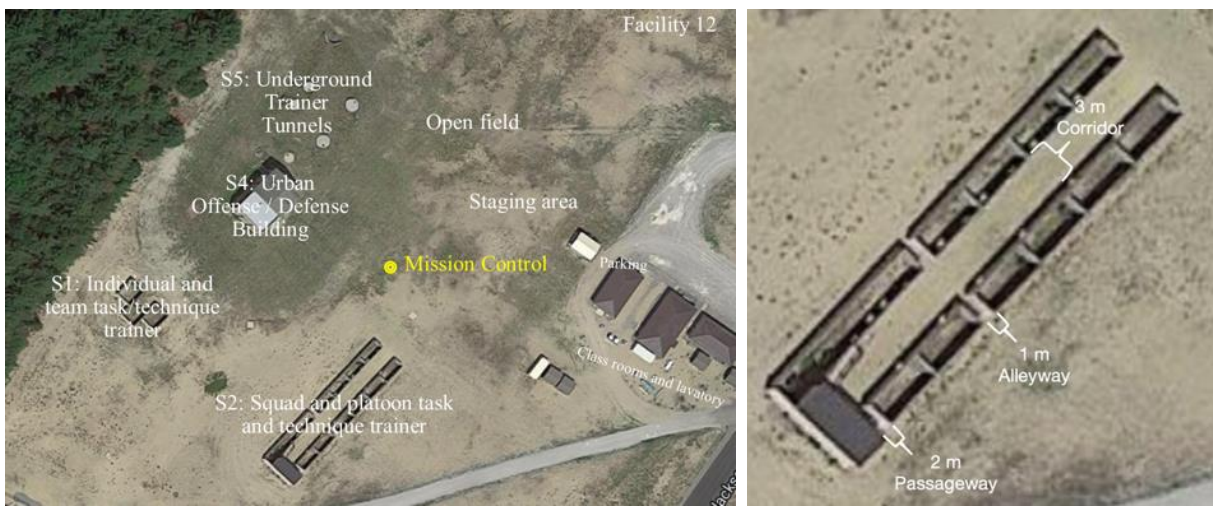

*Figure 38. Fort Devens: Facility 12, Urban Assault Course*

---

[11] Headquarters, Department of the Army. TC 90-1: Training for Urban Operations. May 2008.

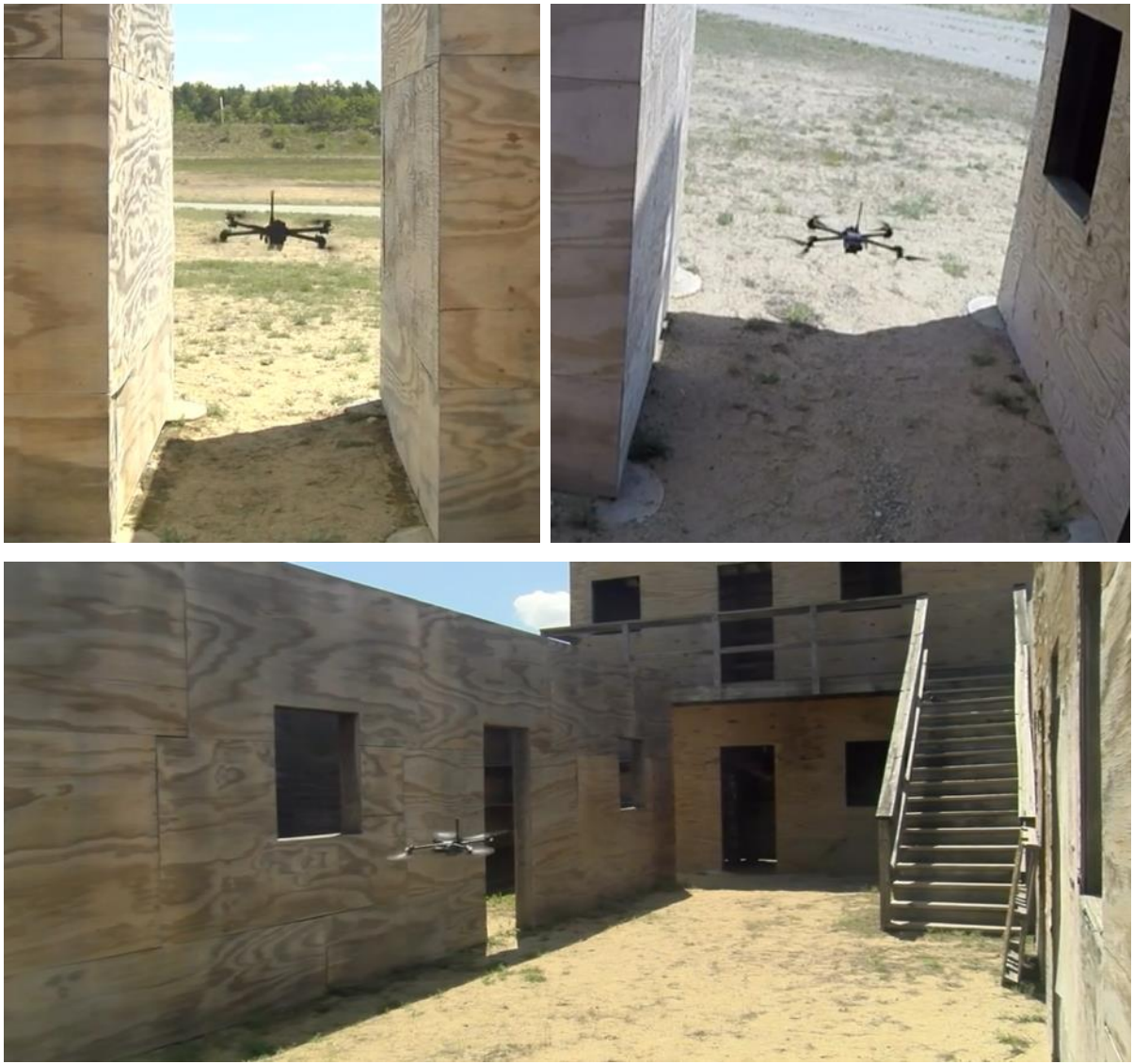

*Figure 39. The Skydio X2D flying through the alleyway (top left), passageway (top right), and corridor (bottom).*

| sUAS | Metrics | Alleyway (1 m wide) | | | Passageway (2 m wide) | | | Corridor (3 m wide) | | |
|---|---|---|---|---|---|---|---|---|---|---|
| | | 1 | 2 | 3 | 1 | 2 | 3 | 1 | 2 | 3 |
| Cleo Robotics Dronut X1P* | Success | ✓ | ✗ | | ✗ | | | - | | |
| | Collisions | | 1 | | 1 | | | | | |
| | Completion | 33% | | | 0% | | | 0% | | |
| FLIR Black Hornet 3 PRS | Success | ✓ | ✓ | - | ✓ | ✗ | | ✓ | - | |
| | Collisions | | | | | 1 | | | | |
| | Completion | 67% | | | 33% | | | 33% | | |
| Flyability Elios 2 GOV | Success | ✓ | ✓ | ✓ | ✓ | ✓ | ✓ | ✓ | ✓ | ✓ |
| | Collisions | | | 1 | 1 | 1 | | | | 1 |
| | Completion | 100% | | | 100% | | | 100% | | |
| Lumenier Nighthawk V3 | Success | ✓ | ✓ | ✓ | ✓ | ✓ | ✓ | ✓ | ✓ | ✓ |
| | Collisions | | 1 | | | | | | | |
| | Completion | 100% | | | 100% | | | 100% | | |
| Parrot ANAFI USA GOV | Success | ✓ | ✓ | ✓ | ✓ | ✓ | ✓ | ✓ | ✓ | ✓ |
| | Collisions | | | | | | | | | |
| | Completion | 100% | | | 100% | | | 100% | | |
| Skydio X2D | Success | ✓ | ✓ | ✓ | ✓ | ✓ | ✓ | ✓ | ✓ | ✓ |
| | Collisions | | | | | | | | | |
| | Completion | 100% | | | 100% | | | 100% | | |
| Vantage Robotics Vesper (CP)† | Success | ✓ | ✓ | ✗ | ✓ | ✓ | - | ✓ | ✓ | - |
| | Collisions | | 1 | 1 | | | | | | |
| | Completion | 67% | | | 67% | | | 67% | | |

*Table 8. Fly Through Confined Outdoor Spaces test results. *Note: Cleo Robotics Dronut X1P had to be flown from 3 meters away due to communications issues, while all other systems were flown from the start position which was 110 meters away. †Note: The Vesper was flown using its caged props (CP) configuration with protective cages to encase the propellers, which are recommended when flying in confined spaces.*

# Outdoor 2D Mapping Accuracy

## Test Method

**Purpose**

This test method evaluates the sUAS' capability at generating 2D maps of outdoor spaces using map generation software, which is typically done by collecting photos tagged with GPS coordinates or video during flight. The resulting map may be used for future mission planning, so the accuracy of its resulting dimensions is critical.

**Summary of Test Method**

To run this test, numbered markers are positioned on the ground throughout the environment and their GPS coordinates are logged to use as a ground truth for comparison. A high-accuracy GPS unit (e.g., within centimeter accuracy) should be used to log their positions. The sUAS is flown to generate a 2D map of the environment and the positions of the markers in the generated map and the ground truth are compared. This is done by measuring the distance between all markers in each map and comparing them to generate an average error. A minimum of 10 markers should be placed in the environment and they should be spread out approximately evenly. The sUAS can perform any type of flight pattern that will result in a 2D map, mostly likely following a grid pattern with its camera facing straight down (-90°), which can be either be programmed for the sUAS to follow automatically using GPS coordinates or manually flown by an operator. See Figure 40 for an example test method layout.

After conducting the mapping flight, the photos or video(s) collected by the sUAS are used by mapping software (e.g., Pix4D, Reveal Farsight) to stitch together the photos or frames from the video(s) to generate a map.

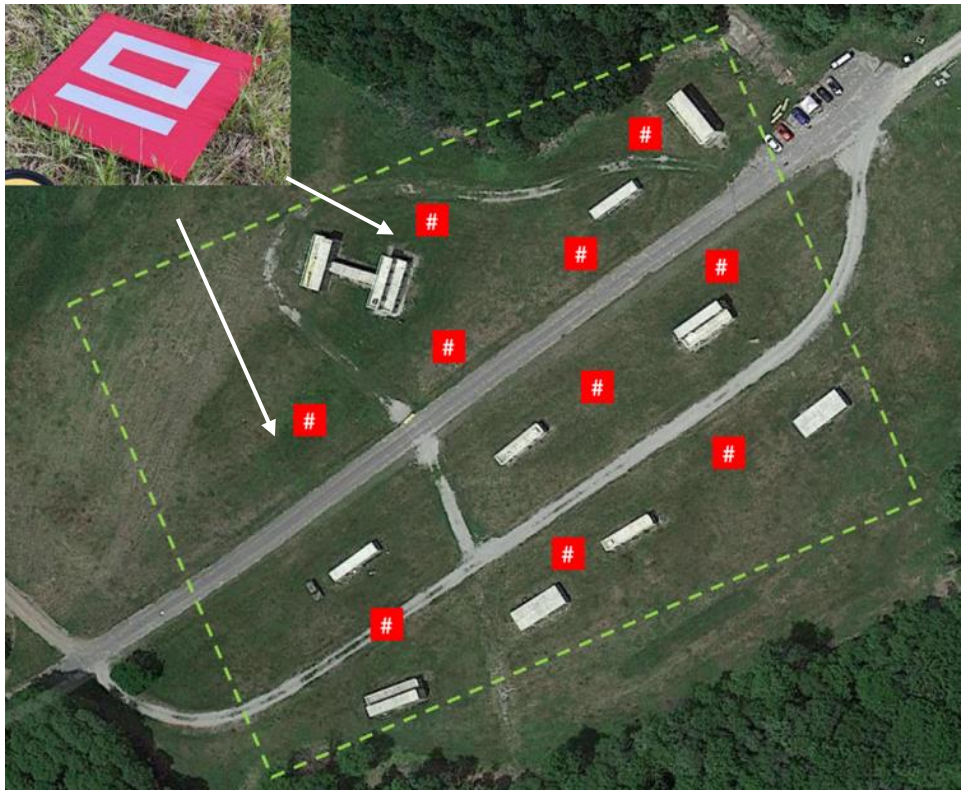

*Figure 40. Example layout for the Outdoor 2D Mapping Accuracy test with 10 numbered markers (example shown in the top left) placed throughout the environment.*

## Apparatus and Artifacts

Numbered markers of sufficient size and color contrast are placed at each ground truth position. The shape of the markers should have at least one distinct corner that is used as the position of its corresponding GPS coordinate for the ground truth. This corner and its sufficient size and color contrast will then enable it to be easily discerned when its coordinates in the evaluation map are being measured. See Figure 41 for examples of 60 cm [24 in] square distance markers consisting of a wooden panel covered in red duct tape with white numbers for increased visibility, where a specific corner is designated to use as the marker's position (e.g., the top-right corner).

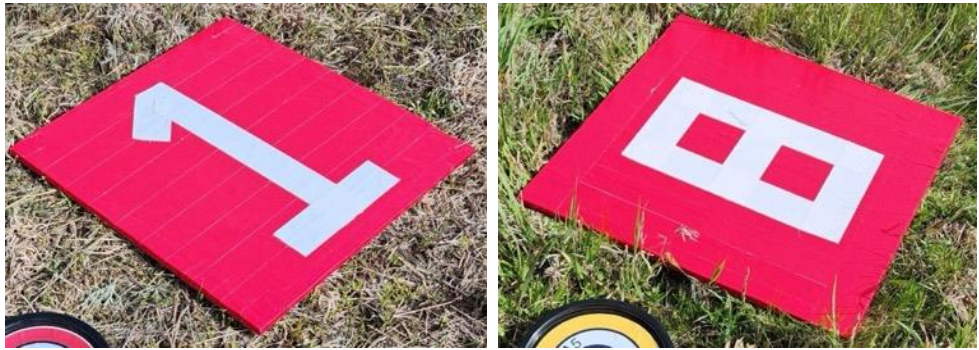

*Figure 41. Example numbered ground truth markers.*

A real-world environment is used of sufficient size that is representative of relevant mission parameters.

## Equipment

The GPS coordinates of the numbered ground truth markers must be logged using a high-accuracy GPS unit, ideally within centimeter accuracy. This is achievable using a system that subscribes to RTK GNSS corrections provided by satellite (e.g., Leica Zeno 20).

A timer is used to measure time taken to map the environment and the time taken to process the map.

## Metrics

- Coverage: The number of ground truth markers mapped compared to the number available. Reported as a percentage (%).
- Average error: Relative distance measurements of marker positions to one another across the entire map compared to the ground truth, averaged across all markers. Reported in meters (m), scaled from pixels. A custom software tool is available that evaluates this metric after manual input of marker locations in the ground truth and evaluation map.
- Map source size: The number of photos or length of the video used to generate the map.
- Mapping time: The amount of time between when the sUAS takes off to when it lands after mapping the environment. Reported in minutes (min).
- Processing time: The amount of time it takes the map generation software to process the photos or video(s) and generate a map for evaluation. Reported in minutes (min).

**Procedure**

1. Position ground truth markers throughout the space to be mapped.
2. Capture high-accuracy GPS coordinates of the markers to use as a ground truth.
3. Instruct the operator to launch the sUAS and begin mapping the area using either photos or video(s).
4. Once the sUAS has launched, start the timer.
5. Once the sUAS has finished mapping the area, instruct the operator to return home and land.
6. Once the sUAS has landed, stop the timer. This duration is used for the mapping time metric.
7. Download the data collected by the sUAS.
8. Launch the mapping software and load the photos or video(s) in order to generate a map.
9. Once the data starts processing, start the timer.
10. Once the mapping software has finished processing and the map is available to view, stop the timer. This duration is used for the processing time metric.
11. Record the pixel coordinates of each of the ground truth markers able to be identified in the map.
12. Compare the distances between markers in the ground truth and the evaluation map.

## Test Results

Benchmarking was conducted at Fort Devens: Facility 15, Smithville, which consists of multiple structures made of connex containers in varying configurations measuring approximately 20,000 m$^2$ (~5 acres) ; see Figure 42. Test results are derived from benchmarking conducted in October 2023. Only sUAS that have outdoor mapping capabilities were evaluated; however, the Vantage Robotics Vesper was not operational at testing time. All drones performed grid pattern flights at an elevation of approximately 38 m [125 ft]. Three types of mapping software were evaluated: Pix4Dreact, Pix4Dmapper, and Reveal Farsight.

A set of 10 markers were placed throughout the area and their GPS locations were recorded using a handheld Leica GPS unit with RTK corrections for centimeter accuracy (between 2.4 and 3.5 cm accuracy was reported across all points; see Table 9).

A summary of the test results is provided in Figure 43 followed by detailed results. Achieving a lower average error (i.e., positions of the markers in the evaluation map are closer to their positions in the ground truth) from a map that is generated faster (i.e., lower processing time) is desirable.

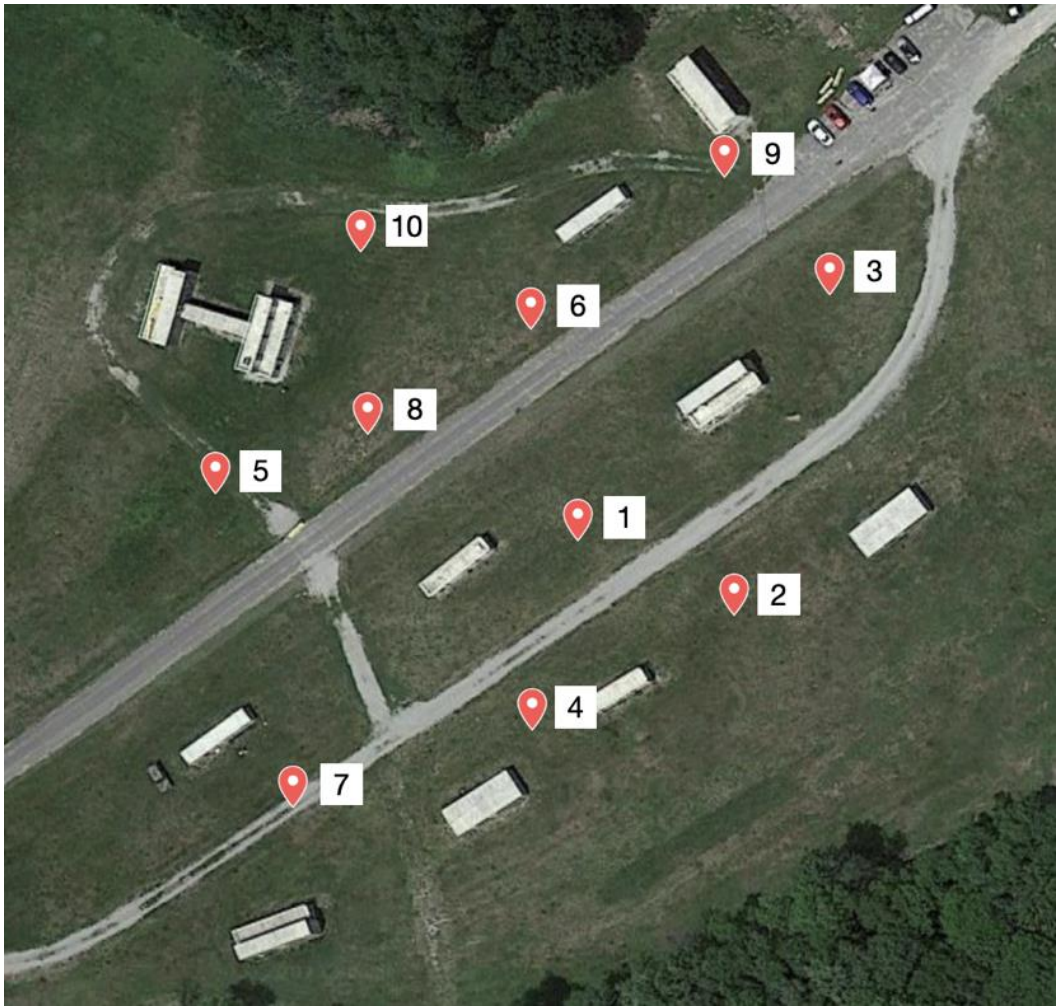

*Figure 42. Fort Devens: Facility 15, Smithville with ground truth markers plotted according to their GPS coordinates.*

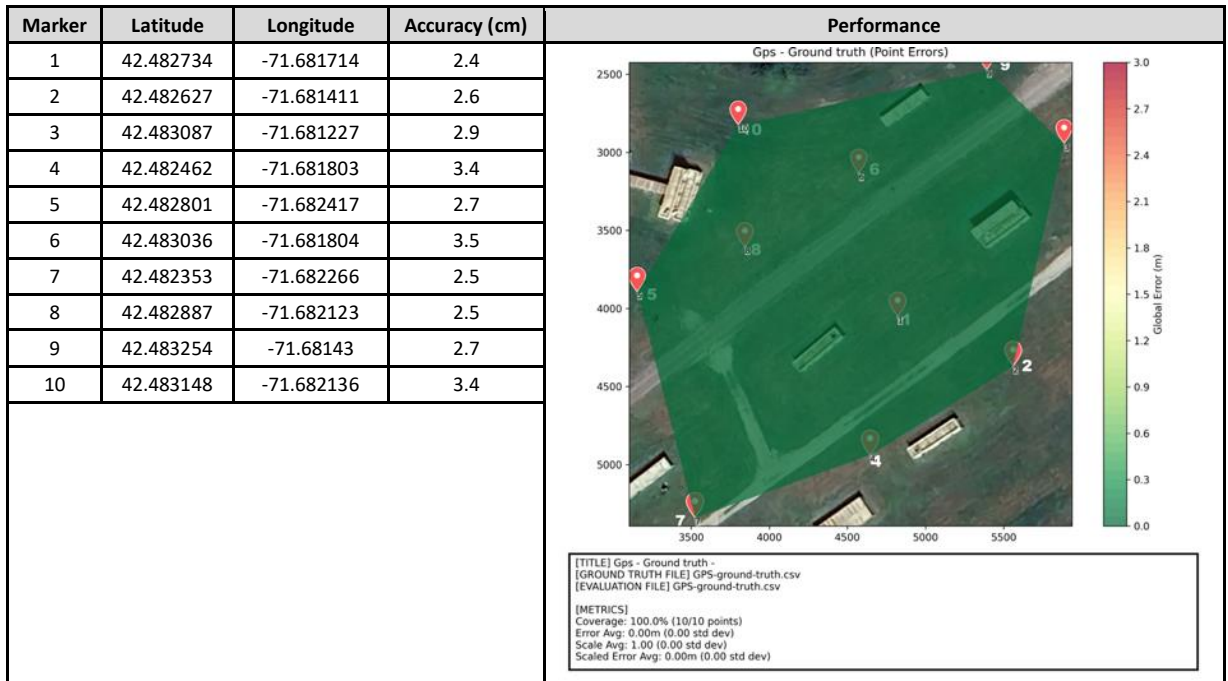

| Marker | Latitude | Longitude | Accuracy (cm) |
|---|---|---|---|
| 1 | 42.482734 | -71.681714 | 2.4 |
| 2 | 42.482627 | -71.681411 | 2.6 |
| 3 | 42.483087 | -71.681227 | 2.9 |
| 4 | 42.482462 | -71.681803 | 3.4 |
| 5 | 42.482801 | -71.682417 | 2.7 |
| 6 | 42.483036 | -71.681804 | 3.5 |
| 7 | 42.482353 | -71.682266 | 2.5 |
| 8 | 42.482887 | -71.682123 | 2.5 |
| 9 | 42.483254 | -71.68143 | 2.7 |
| 10 | 42.483148 | -71.682136 | 3.4 |

*Table 9. Ground truth map and marker locations with accuracy as reported by the Leica GPS unit used to mark them.*

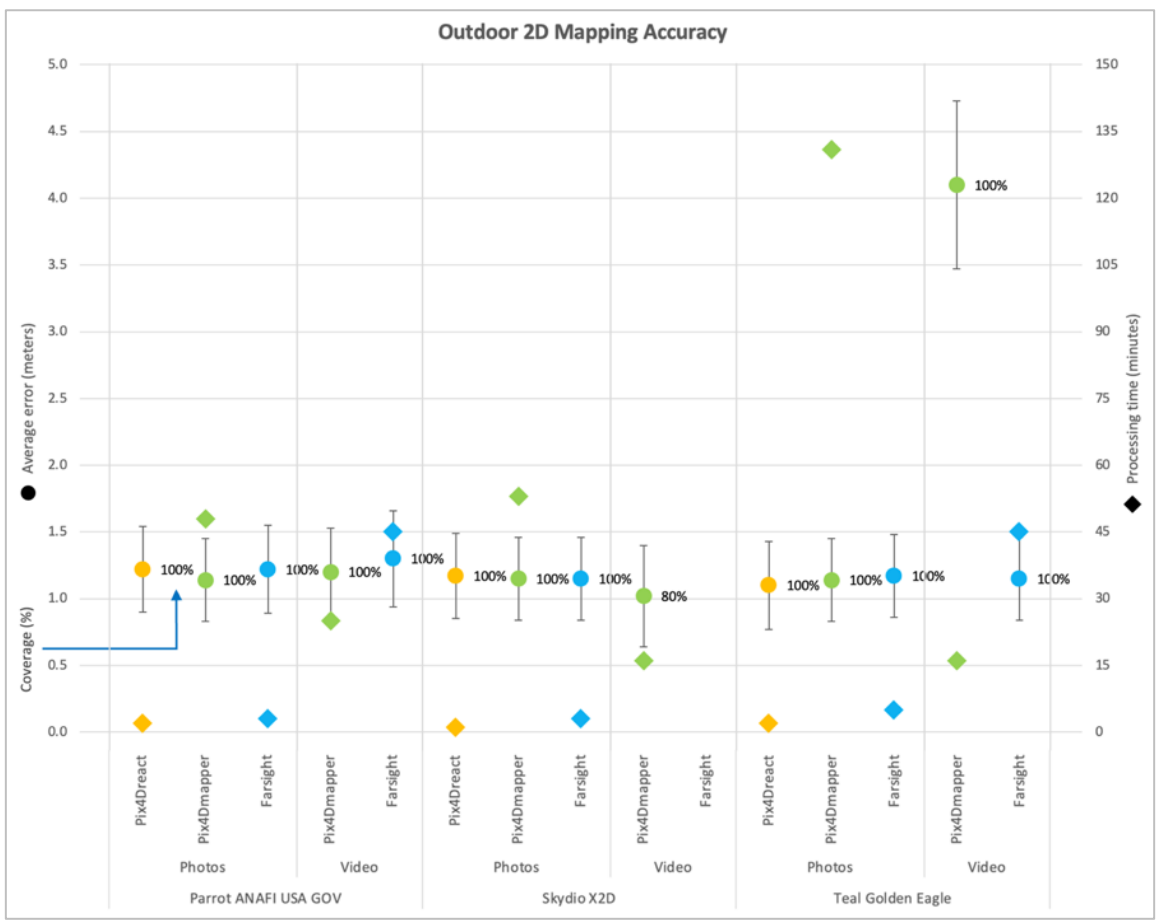

*Figure 43. Summarized Outdoor 2D Mapping Accuracy test results.*

## Parrot ANAFI USA GOV

| sUAS and Mapping Software | Metrics | Performance |
|---|---|---|
| Parrot ANAFI USA GOV<br><br>Pix4Dreact | Flight pattern | Grid |
| | Map source format | Photos |
| | Map source size | 164 photos |
| | Mapping time | 6 min |
| | Processing time | 2 min |
| | Coverage | 100% |
| | Average error | 1.22 m (+/- 0.32 m) |

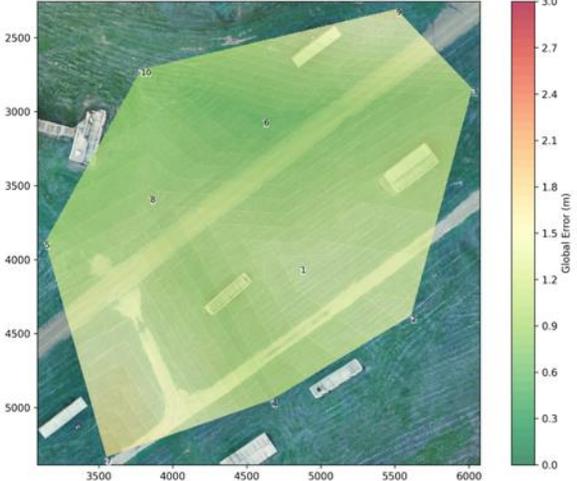

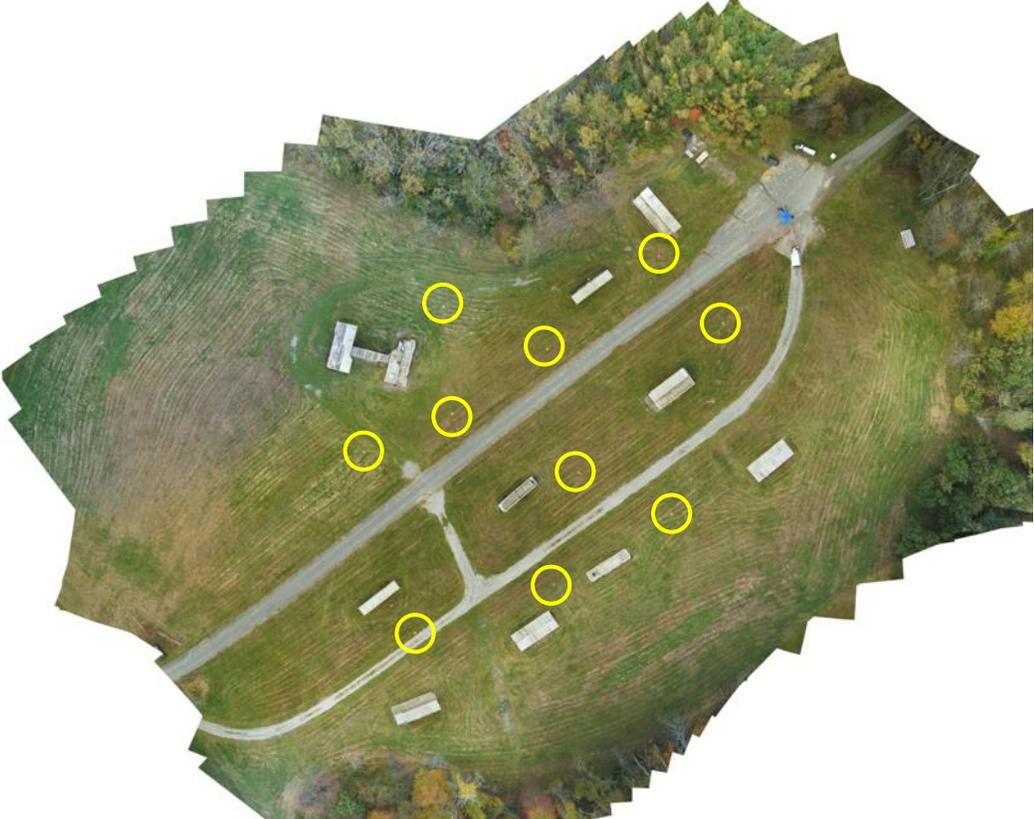

*Table 10. Outdoor 2D Mapping test results for the Parrot ANAFI USA GOV.*

| sUAS and Mapping Software | Metrics | Performance |
|---|---|---|
| Parrot ANAFI USA GOV<br><br>Pix4Dmapper | Flight pattern | Grid |
| | Map source format | Photos |
| | Map source size | 164 photos |
| | Mapping time | 6 min |
| | Processing time | 48 min |
| | Coverage | 100% |
| | Average error | 0.95 m (+/- 0.02 m) |

[TITLE] Parrot - Mapper photo -
[GROUND TRUTH FILE] gt.csv
[EVALUATION FILE] parrot-mapper-photo.csv

[METRICS]
Coverage: 100.0% (10/10 points)
Error Avg: 3.32m (0.93 std dev)
Scale Avg: 0.95 (0.02 std dev)
Scaled Error Avg: 1.14m (0.31 std dev)

*Table 9. (continued) Outdoor 2D Mapping test results for the Parrot ANAFI USA GOV.*

| sUAS and Mapping Software | Metrics | | Performance |
|---|---|---|---|
| Parrot ANAFI USA GOV<br><br>Farsight | Flight pattern | Grid | |
| | Map source format | Photos | |
| | Map source size | 164 photos | |
| | Mapping time | 6 min | |
| | Processing time | 3 min | |
| | Coverage | 100% | |
| | Average error | 1.22 m (+/- 0.33 m) | |

[TITLE] Parrot - Farsight photo -
[GROUND TRUTH FILE] gt.csv
[EVALUATION FILE] parrot-farsight-photo.csv

[METRICS]
Coverage: 100.0% (10/10 points)
Error Avg: 4.31m (1.16 std dev)
Scale Avg: 1.08 (0.03 std dev)
Scaled Error Avg: 1.22m (0.33 std dev)

*Table 9. (continued) Outdoor 2D Mapping test results for the Parrot ANAFI USA GOV.*

| sUAS and Mapping Software | Metrics | Performance | |
|---|---|---|---|
| Parrot ANAFI USA GOV<br><br>Pix4Dmapper | Flight pattern | Grid | 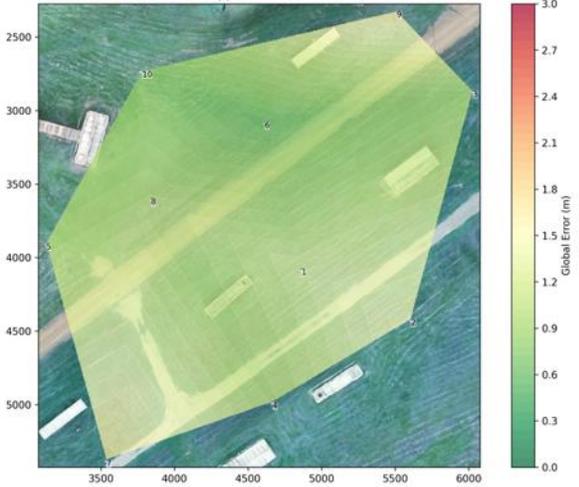 |
| | Map source format | Video | |
| | Map source size | 6:46 | |
| | Mapping time | 7 min | |
| | Processing time | 25 min | |
| | Coverage | 100% | |
| | Average error | 1.2 m (+/- 0.33 m) | |
| | | 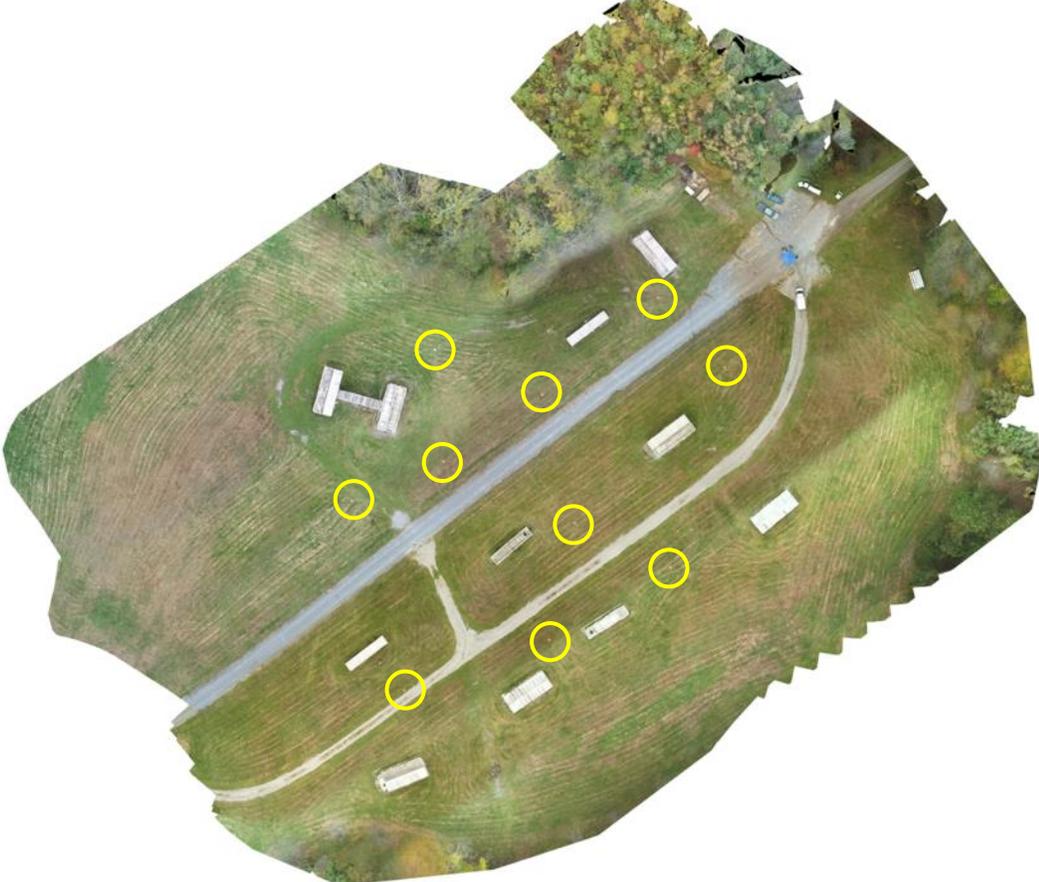 | |

Table 9. (continued) Outdoor 2D Mapping test results for the Parrot ANAFI USA GOV.

| sUAS and Mapping Software | Metrics | Performance | |
|---|---|---|---|
| Parrot ANAFI USA GOV<br><br>Farsight | Flight pattern | Grid | |
| | Map source format | Video | |
| | Map source size | 6:46 | |
| | Mapping time | 7 min | |
| | Processing time | 45 min | |
| | Coverage | 100% | |
| | Average error | 1.3 m (+/- 0.36 m) | |

[TITLE] Parrot - Farsight video -
[GROUND TRUTH FILE] gt.csv
[EVALUATION FILE] parrot-farsight-video.csv

[METRICS]
Coverage: 100.0% (10/10 points)
Error Avg: 1.91m (0.83 std dev)
Scale Avg: 1.03 (0.03 std dev)
Scaled Error Avg: 1.30m (0.36 std dev)

*Table 9. (continued) Outdoor 2D Mapping test results for the Parrot ANAFI USA GOV.*

## Skydio X2D

| sUAS and Mapping Software | Metrics | Performance | |
|---|---|---|---|
| Skydio X2D<br><br>Pix4Dreact | Flight pattern | Grid | 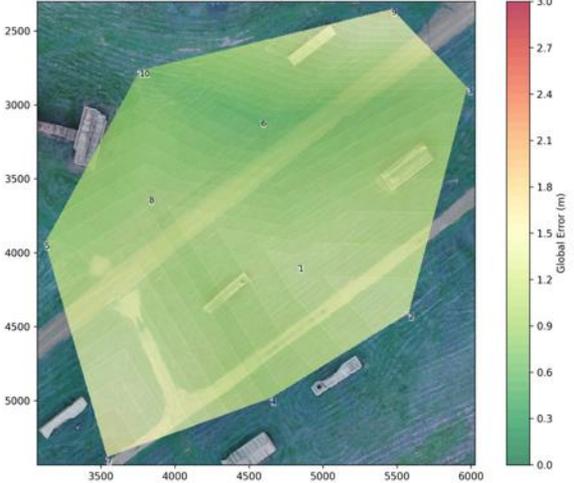 |
| | Map source format | Photos | |
| | Map source size | 178 photos | |
| | Mapping time | 7 min | |
| | Processing time | 1 min | |
| | Coverage | 100% | |
| | Average error | 1.17 m (+/- 0.32 m) | |
| | 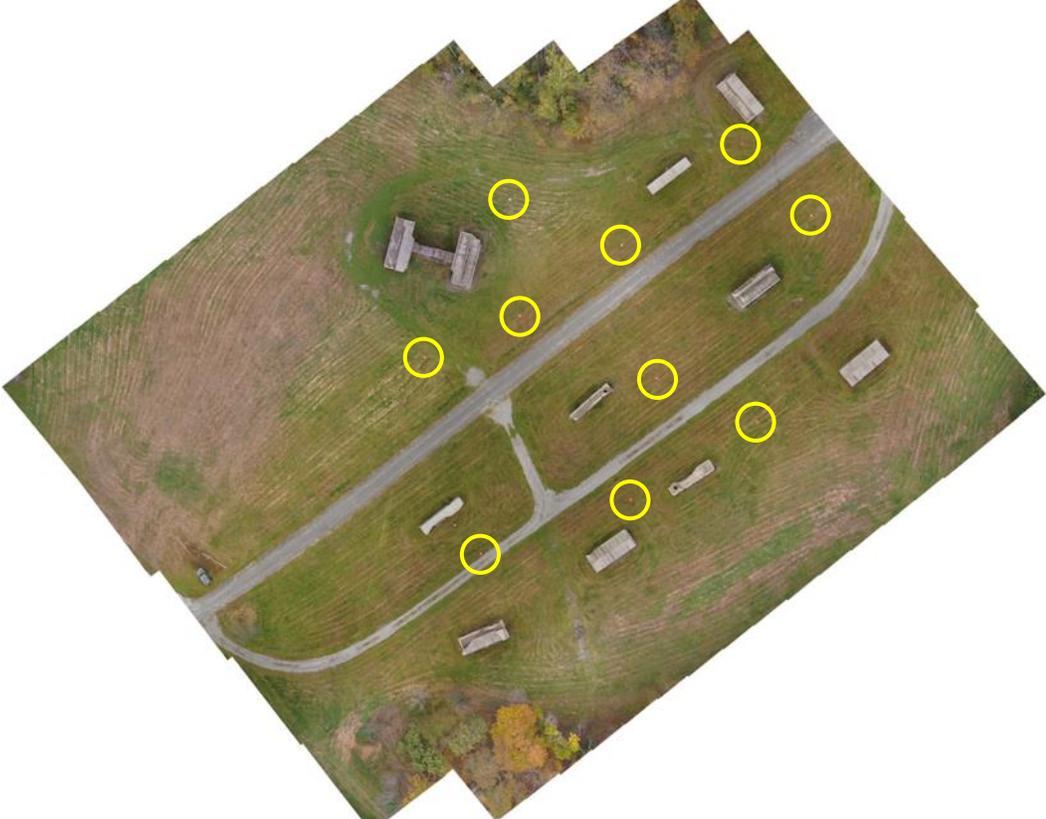 | | |

*Table 11. Outdoor 2D Mapping test results for the Skydio X2D.*

| sUAS and Mapping Software | Metrics | Performance | |
|---|---|---|---|
| Skydio X2D<br><br>Pix4Dmapper | Flight pattern | Grid | 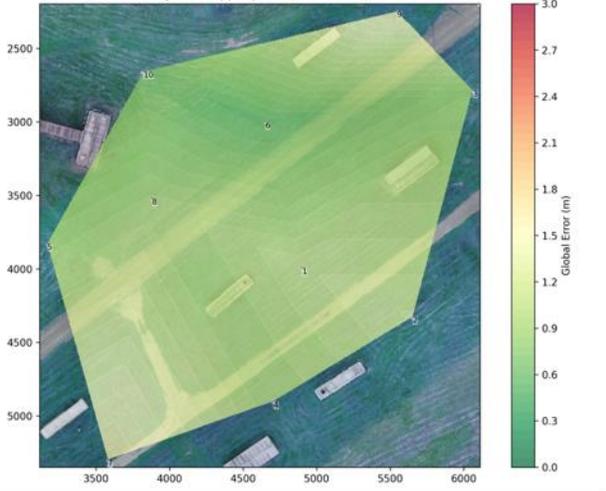 |
| | Map source format | Photos | |
| | Map source size | 178 photos | |
| | Mapping time | 7 min | |
| | Processing time | 53 min | |
| | Coverage | 100% | |
| | Average error | 1.15 m (+/- 0.31 m) | |
| | | 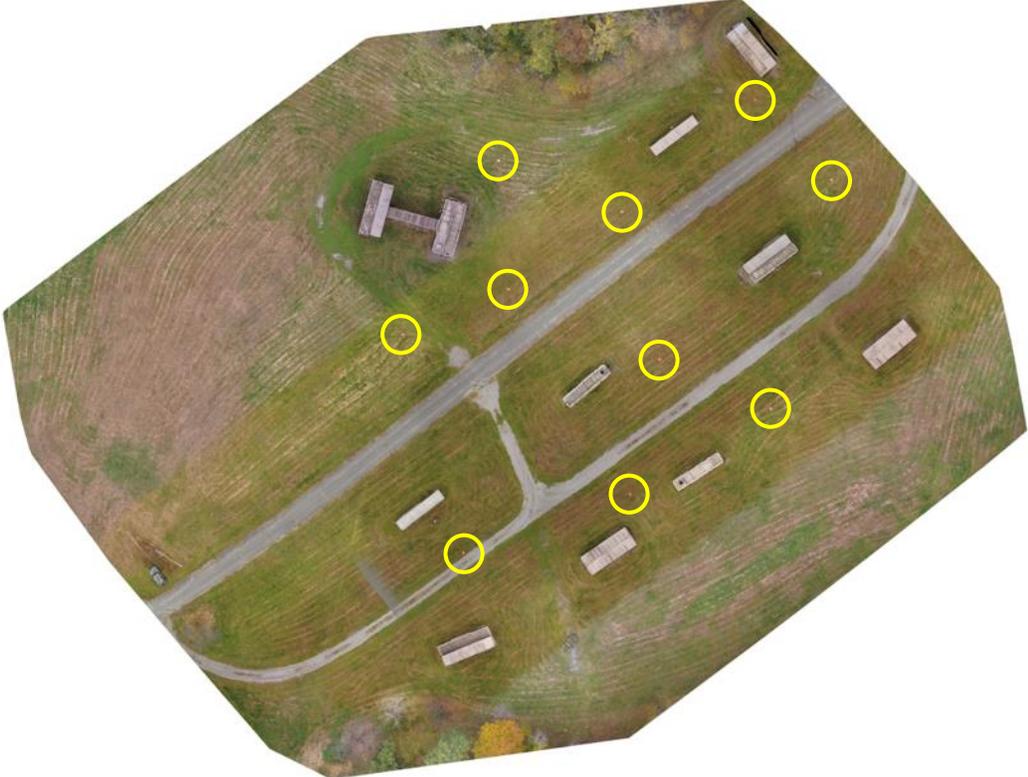 | |

*Table 10. (continued) Outdoor 2D Mapping test results for the Skydio X2D.*

| sUAS and Mapping Software | Metrics | Performance | |
|---|---|---|---|
| Skydio X2D Farsight | Flight pattern | Grid | 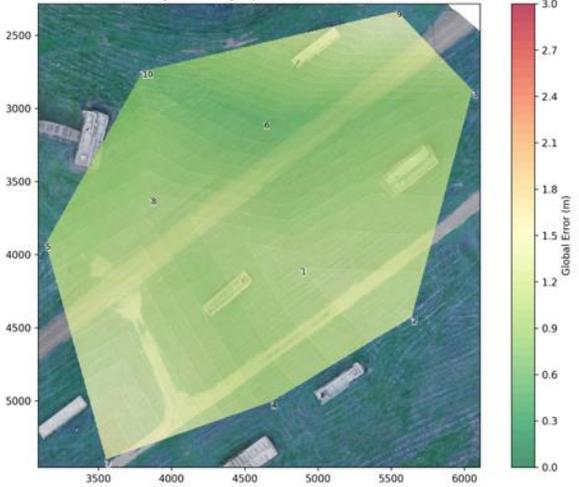 |
| | Map source format | Photos | |
| | Map source size | 178 photos | |
| | Mapping time | 7 min | |
| | Processing time | 3 min | |
| | Coverage | 100% | |
| | Average error | 1.15 m (+/- 0.31 m) | |
| | | 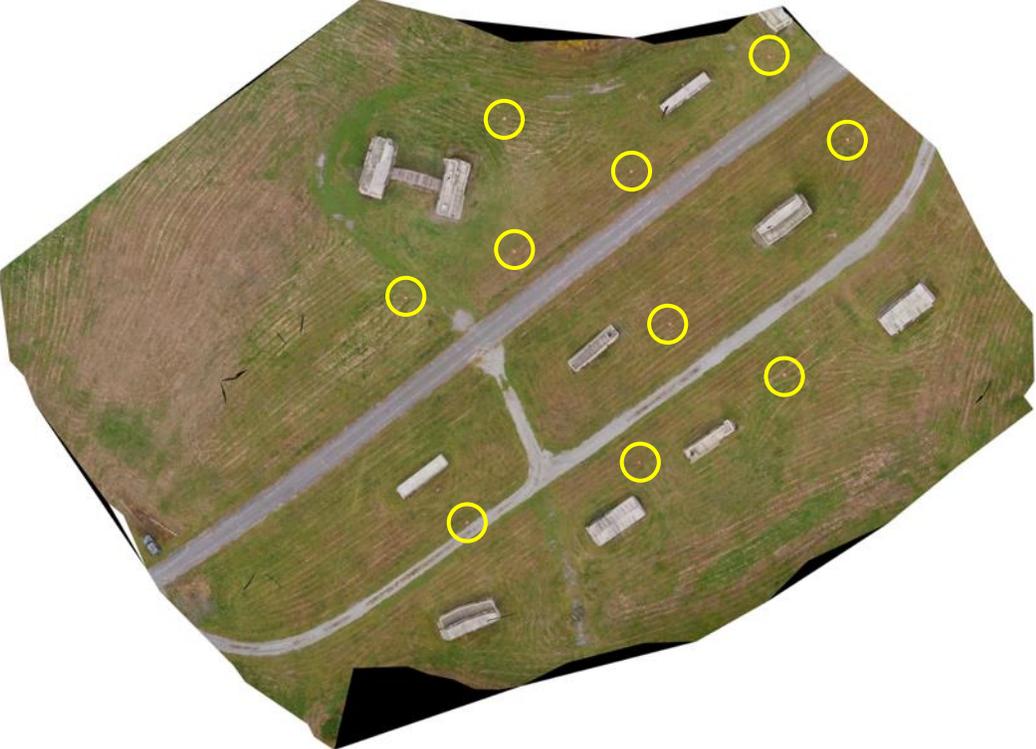 | |

*Table 10. (continued) Outdoor 2D Mapping test results for the Skydio X2D.*

| sUAS and Mapping Software | Metrics | | Performance |
|---|---|---|---|
| Skydio X2D<br><br>Pix4Dmapper | Flight pattern | Grid | 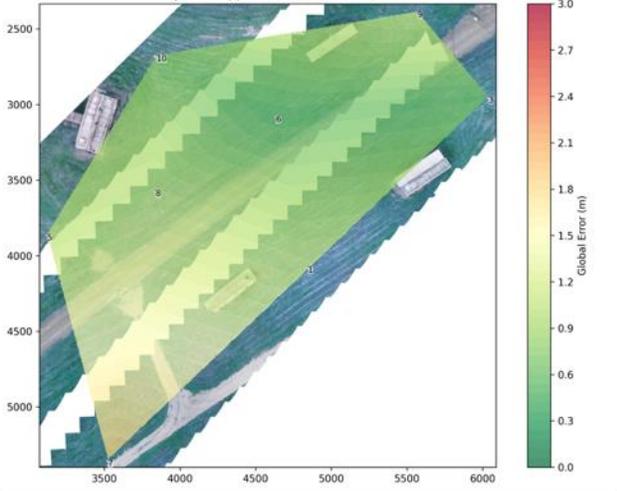 |
| | Map source format | Video | |
| | Map source size | 6:25 | |
| | Mapping time | 6 min | |
| | Processing time | 16 min | |
| | Coverage | 80% | |
| | Average error | 1.02 m (+/- 0.38 m) | |
| | 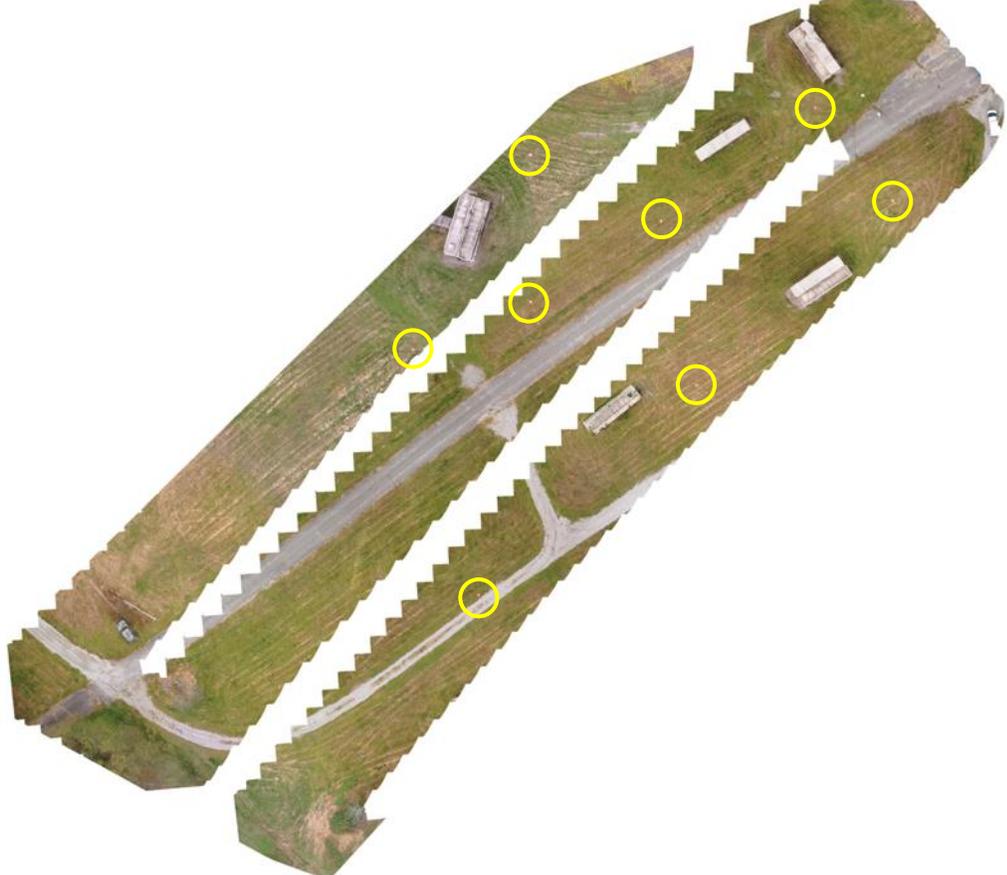 | | |

*Table 10. (continued) Outdoor 2D Mapping test results for the Skydio X2D.*

| sUAS and Mapping Software | Metrics | | Performance | |
|---|---|---|---|---|
| Skydio X2D  Farsight | Flight pattern | Grid | N/A | |
| | Map source format | Video | | |
| | Map source size | 6:25 | | |
| | Mapping time | 6 min | | |
| | Processing time | N/A | | |
| | Coverage | N/A | | |
| | Average error | N/A | | |
| | | | | |
| | This map was not able to be processed | | | |

Table 10. (continued) Outdoor 2D Mapping test results for the Skydio X2D.

## Teal Golden Eagle

| sUAS and Mapping Software | Metrics | Performance | |
|---|---|---|---|
| Teal Golden Eagle<br><br>Pix4Dreact | Flight pattern | Grid | |
| | Map source format | Photos | |
| | Map source size | 95 photos | |
| | Mapping time | 7 min | |
| | Processing time | 2 min | |
| | Coverage | 100% | |
| | Average error | 1.1 m (+/- 0.33 m) | |

*Table 12. Outdoor 2D Mapping test results for the Teal Golden Eagle.*

| sUAS and Mapping Software | Metrics | | Performance |
|---|---|---|---|
| Teal Golden Eagle<br><br>Pix4Dmapper | Flight pattern | Grid | |
| | Map source format | Photos | |
| | Map source size | 95 photos | |
| | Mapping time | 7 min | |
| | Processing time | 131 min | |
| | Coverage | 100% | |
| | Average error | 1.14 m (+/- 0.31 m) | |

*Table 11. (continued) Outdoor 2D Mapping test results for the Teal Golden Eagle.*

| sUAS and Mapping Software | Metrics | | Performance |
|---|---|---|---|
| Teal Golden Eagle Farsight | Flight pattern | Grid | 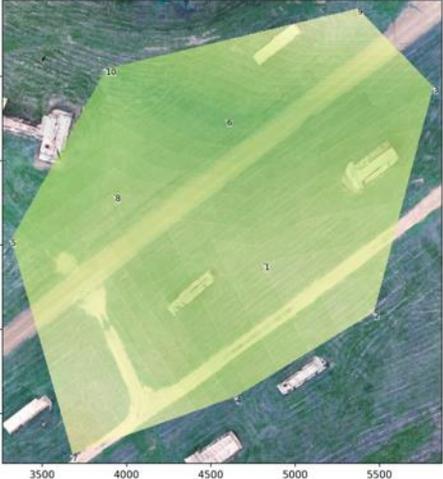 |
| | Map source format | Photos | |
| | Map source size | 95 photos | |
| | Mapping time | 7 min | |
| | Processing time | 5 min | |
| | Coverage | 100% | |
| | Average error | 1.17 m (+/- 0.31 m) | |
| | 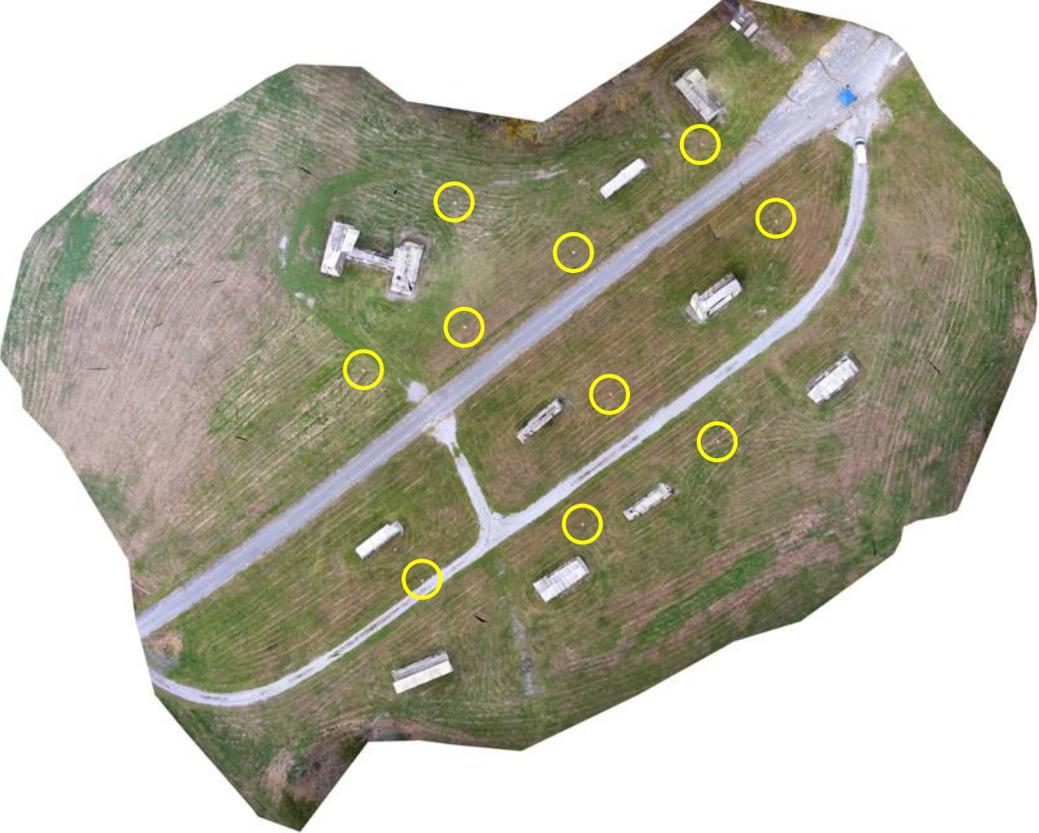 | | |

Table 11. (continued) Outdoor 2D Mapping test results for the Teal Golden Eagle.

| sUAS and Mapping Software | Metrics | Performance | |
|---|---|---|---|
| Teal Golden Eagle<br><br>Pix4Dmapper | Flight pattern | Grid | |
| | Map source format | Video | |
| | Map source size | 6:40 | |
| | Mapping time | 7 min | |
| | Processing time | 16 min | |
| | Coverage | 100% | |
| | Average error | 4.1 m (+/- 0.63 m) | |

[TITLE] Teal - Mapper video -
[GROUND TRUTH FILE] gt.csv
[EVALUATION FILE] teal-mapper-video.csv
[METRICS]
Coverage: 100.0% (10/10 points)
Error Avg: 4.59m (1.37 std dev)
Scale Avg: 0.95 (0.08 std dev)
Scaled Error Avg: 4.10m (0.63 std dev)

*Table 11. (continued) Outdoor 2D Mapping test results for the Teal Golden Eagle.*

| sUAS and Mapping Software | Metrics | Performance | |
|---|---|---|---|
| Teal Golden Eagle<br><br>Farsight | Flight pattern | Grid | |
| | Map source format | Video | |
| | Map source size | 6:40 | |
| | Mapping time | 7 min | |
| | Processing time | 45 min | |
| | Coverage | 100% | |
| | Average error | 1.15 m (+/- 0.31 m) | |

*Table 11. (continued) Outdoor 2D Mapping test results for the Teal Golden Eagle.*

# Outdoor 3D Mapping Accuracy

## Test Method

**Purpose**

This test method evaluates the sUAS' capability at generating 3D maps of outdoor volumes using map generation software, which is typically done by collecting video during flight. The resulting map may be used for future mission planning, so the accuracy of its resulting dimensions and visual features is critical.

**Summary of Test Method**

To run this test, a building/structure is selected that contains features of interest (e.g., signage, doors, windows, railings, etc.). To establish a ground truth of the building, a set of its dimensions are recorded (e.g., length, width, height of building; length and height of doors, etc.) and photos are taken of its visual features of interest (e.g., open/closed doors or windows, hazmat signage, etc.). The accuracy of the resulting 3D map to portray these dimensions and visual features will be evaluated against the ground truth. Typically, sUAS will perform an orbit flight around one or more target buildings/structures while recording video and any other onboard sensors (e.g., proximity distance) with its camera angled downwards (~30-60°), circling the target volume one or more times, which can be either be programmed for the sUAS to follow automatically or manually flown by an operator.

After conducting the mapping flight, the data collected by the sUAS is used by mapping software (e.g., Pix4D, Reveal Farsight) to generate a map typically using some form of photogrammetry. Using the software used to generate the map or another 3D model viewer program (e.g., CloudCompare), the two sets of ground truth data are evaluated in the resulting map:

- <u>Dimensional Accuracy</u>: Using a distance measurement tool provided by the mapping software, measure the dimensions of the building/structure and compare them to the ground truth.
- <u>Feature Recognition</u>: Observing the photo texture on the map, determine if the visual features can be identified properly, including what type of feature it is (e.g., door vs. window) and the state of that feature (e.g., open vs. closed). Ideally, this evaluation would be performed by someone other than the sUAS operator that collected the data.

See Figure 44 for an example of a building/structure used for Outdoor 3D Mapping Accuracy.

**Apparatus and Artifacts**

A real-world environment is used with a building/structure with geometry that is representative of relevant mission parameters.

**Equipment**

A timer is used to measure time taken to map the environment and the time taken to process the map. A measure tape or similar is used to measure the dimensions of the building/structure for ground truth. A camera is used to take photos of the visual features for ground truth.

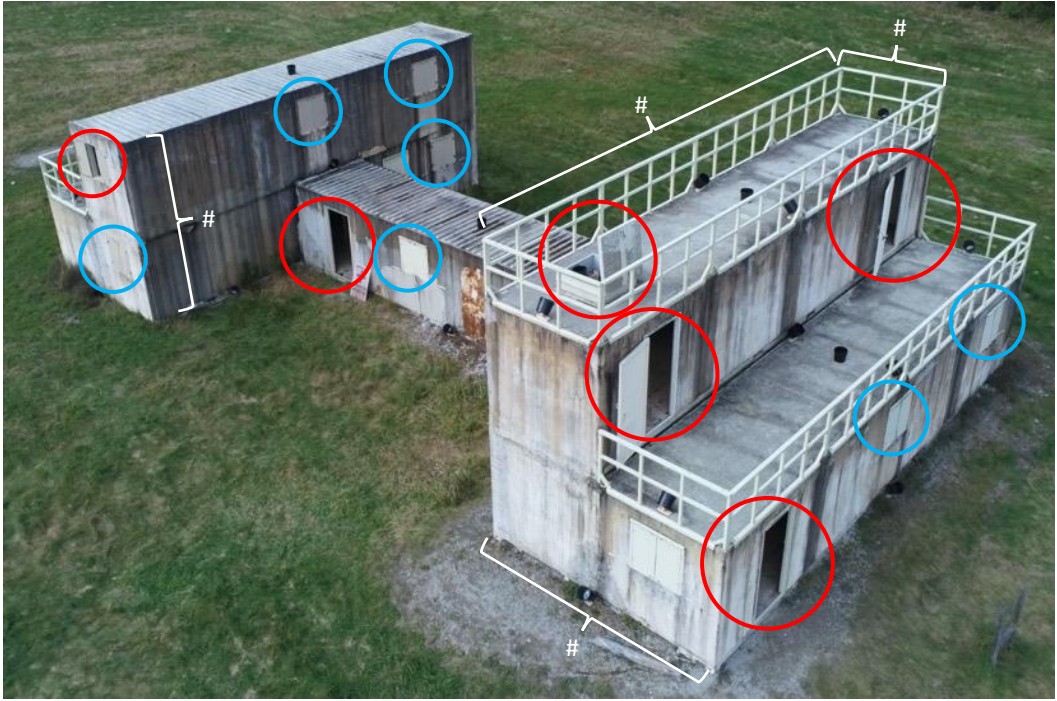

*Figure 44. Example layout for the Outdoor 3D Mapping Accuracy test on a building with dimensions of the building marked for Dimensional Accuracy and doors, windows, and hatches circled (open = red, closed = blue) for Feature Identification.*

## Metrics

- Map source size: The number of photos or length of the video used to generate the map.
- Mapping time: The amount of time between when the sUAS takes off to when it lands after mapping the environment. Reported in minutes (min).
- Processing time: The amount of time it takes the map generation software to process the video(s) and generate a map for evaluation. Reported in minutes (min).
- Dimensional Accuracy:
    - Completeness: The number of dimensions that are able to be measured compared to the number in the ground truth. Reported as a percentage (%).
    - Average error: The difference in dimensions in the evaluation map compared to the ground truth, averaged across all dimensions. Reported in meters (m).
- Feature Recognition:
    - Completeness: The number of visual features that can be accurately compared to the ground truth, as either clearly recognizable, partially recognizable, or not recognizable / missing. Reported as a percentage (%).
    - Correct state identification: The number of visual features that can be accurately identified as being open or closed compared to the ground truth. Reported as a percentage (%).

**Procedure**

1. Measure all dimensions to be used for evaluation for the ground truth.
2. Take photos of all visual features to be used for evaluation for the ground truth.
3. Instruct the operator to launch the sUAS and begin mapping the area.
4. Once the sUAS has launched, start the timer.
5. Once the sUAS has finished mapping the area, instruct the operator to return home and land.
6. Once the sUAS has landed, stop the timer. This duration is used for the <u>mapping time</u> metric.
7. Download the data collected by the sUAS.
8. Launch the mapping software and load the data in order to generate a map.
9. Once the data starts processing, start the timer.
10. Once the mapping software has finished processing and the map is available to view, stop the timer. This duration is used for the <u>processing time</u> metric.
11. Dimensional Accuracy:
    a. Using the mapping software or another 3D model viewer, measure the dimensions of the evaluation map using the digital tools available in the software.
    b. Compare the dimensions in the evaluation map to the ground truth.
12. Feature Recognition:
    a. Using the mapping software or another 3D model viewer, take screenshots of the visual features in the map that are able to be recognized.
    b. Label each screenshot with what the visual feature is (e.g., door or window) and its state (e.g., open or closed).
    c. Compare the visual features in the evaluation map to the ground truth.

## Test Results

Benchmarking was conducted at Fort Devens: Facility 15, Smithville, using a complex building structure that consists of several interconnected connex containers up to two stories high, 10 doors, 20 windows, and 1 hatch, with roof surfaces at one and two stories high, some with railings and some without (the same *complex* building structure used in the **Exterior Building Clearing** test method). The building measures 21.9 m [72 ft] long by 12.2 m [40 ft] wide by 6.1 m (20 ft 2 in) tall to the railing on the roof, or approximately 1,630 m$^3$ [58,080 ft$^3$]; see Figure 45. Test results are derived from benchmarking conducted in July 2023. Only sUAS that have outdoor mapping capabilities were evaluated: Parrot ANAFI USA GOV, Skydio X2D, Teal Golden Eagle, and Vantage Robotics Vesper. All drones performed orbit pattern flights with 3 rotations around the building at an elevation of approximately 35 m [115 ft]. Two types of mapping software were evaluated: Pix4Dmapper and Reveal Farsight.

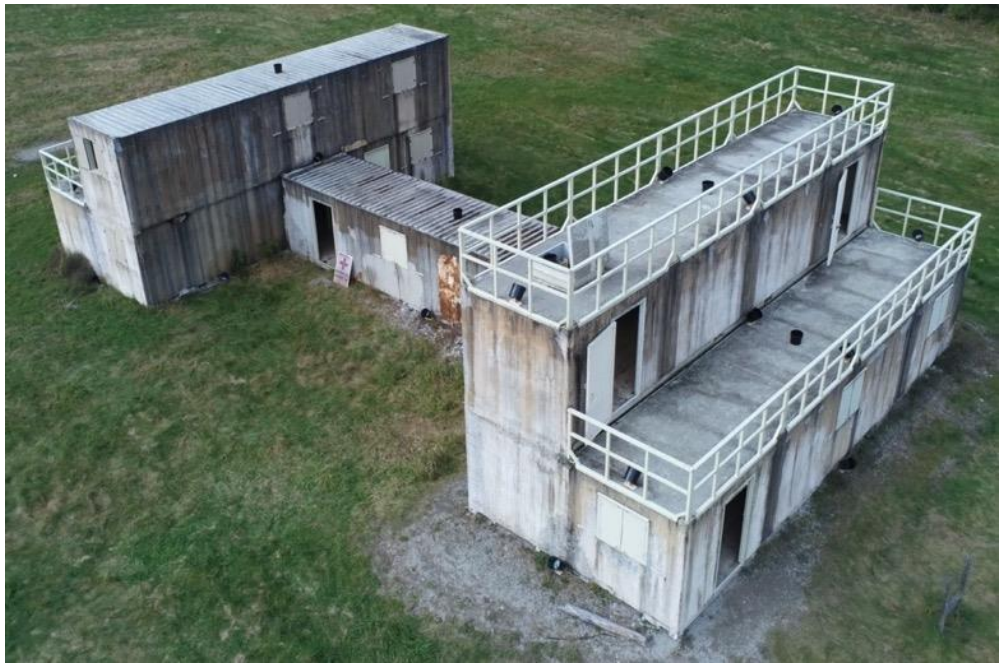

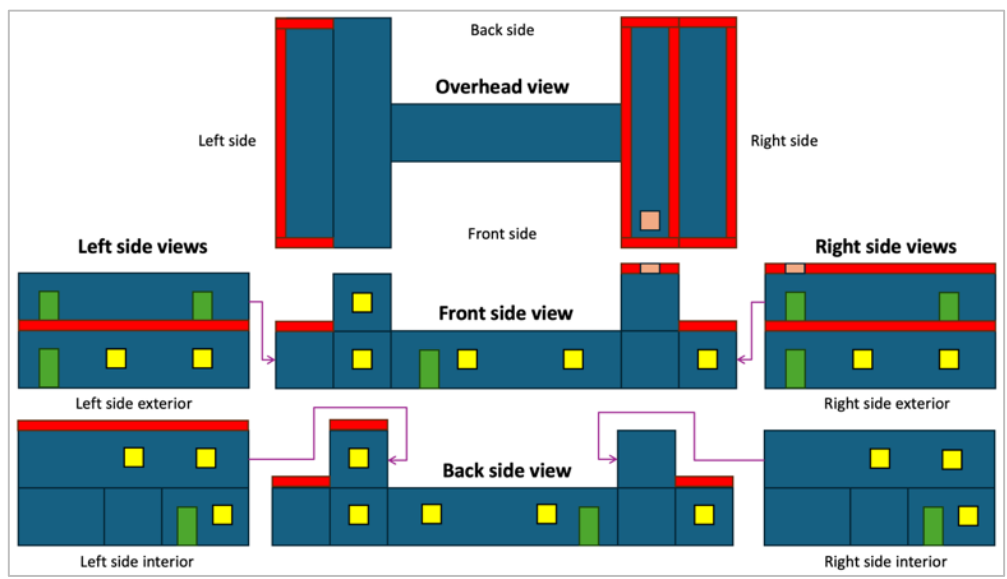

*Figure 45. Fort Devens: Facility 15, Smithville, complex building structure*

A summary of the test results is provided in Figure 46 and Table 13, comparing five metrics: Processing time (compared to the average), Dimensional Accuracy: Completeness, Dimensional Accuracy: Average error (compared to the average), Feature Recognition: Completeness, and Feature Recognition: Correct state identification. Detailed results for both types of evaluation are presented after the summary. Note: Reveal Farsight provides dimensional measurements rounded to whole feet, which have been converted to meters for this report.

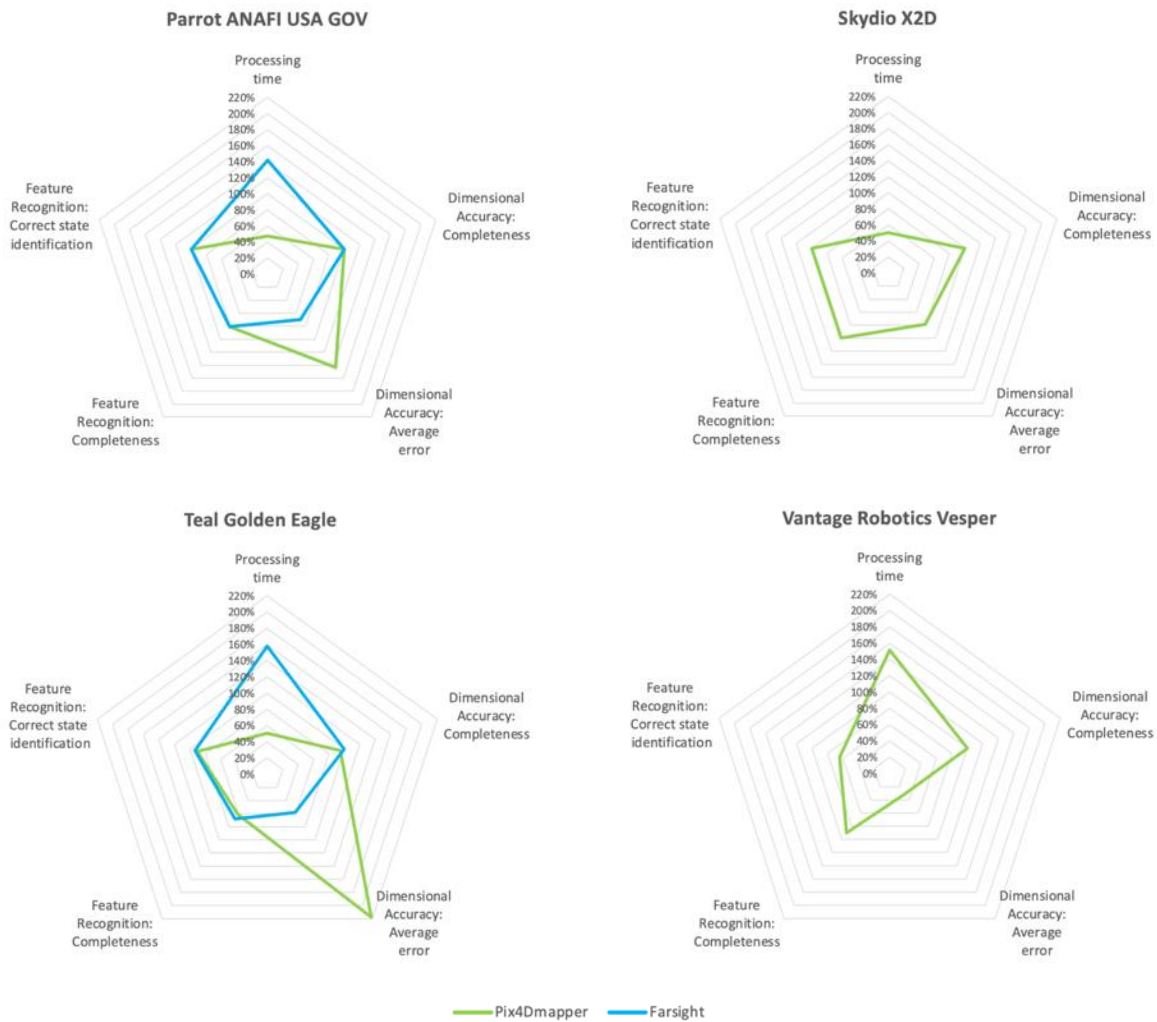

*Figure 46. Summarized Outdoor 3D Mapping Accuracy test results.*

| sUAS | Mapping Software | Processing time | | Dimensional Accuracy | | | Feature Recognition | |
|---|---|---|---|---|---|---|---|---|
| | | (min) | Compared to the average (32 min) | Completeness | Average error | | Completeness | Correct state identification |
| | | | | | (m) | Compared to the average (0.45 m) | | |
| Parrot ANAFI USA GOV | Pix4Dmapper | 15 min | 47% | 100% | 0.64 m (+/- 0.88 m) | 144% | 81% | 100% |
| | Farsight | 45 min | 142% | 100% | 0.31 m (+/- 0.27 m) | 70% | 81% | 100% |
| Skydio X2D | Pix4Dmapper | 16 min | 51% | 100% | 0.35 m (+/- 0.39 m) | 79% | 100% | 100% |
| | Farsight | This map was not able to be processed | | | | | | |
| Teal Golden Eagle | Pix4Dmapper | 16 min | 51% | 95% | 0.97 m (+/- 1.38 m) | 218% | 61% | 90% |
| | Farsight | 50 min | 158% | 100% | 0.26 m (+/- 0.23 m) | 58% | 68% | 94% |
| Vantage Robotics Vesper | Pix4Dmapper | 48 min | 152% | 100% | 0.14 m (+/- 0.2 m) | 31% | 90% | 65% |
| | Farsight | This map was not able to be processed | | | | | | |

*Table 13. Summarized Outdoor 3D Mapping Accuracy test results.*

## Dimensional Accuracy

A total of 37 dimensions were evaluated: 10 of the building (various widths, lengths, and heights), 12 of the doors (width and height of 6 visible doors), 12 of the windows (width and height of 6 visible windows), and 3 of the hatch (width and length of the hatch, height of the hatch door). See Figure 47 for a diagram of all dimensions.

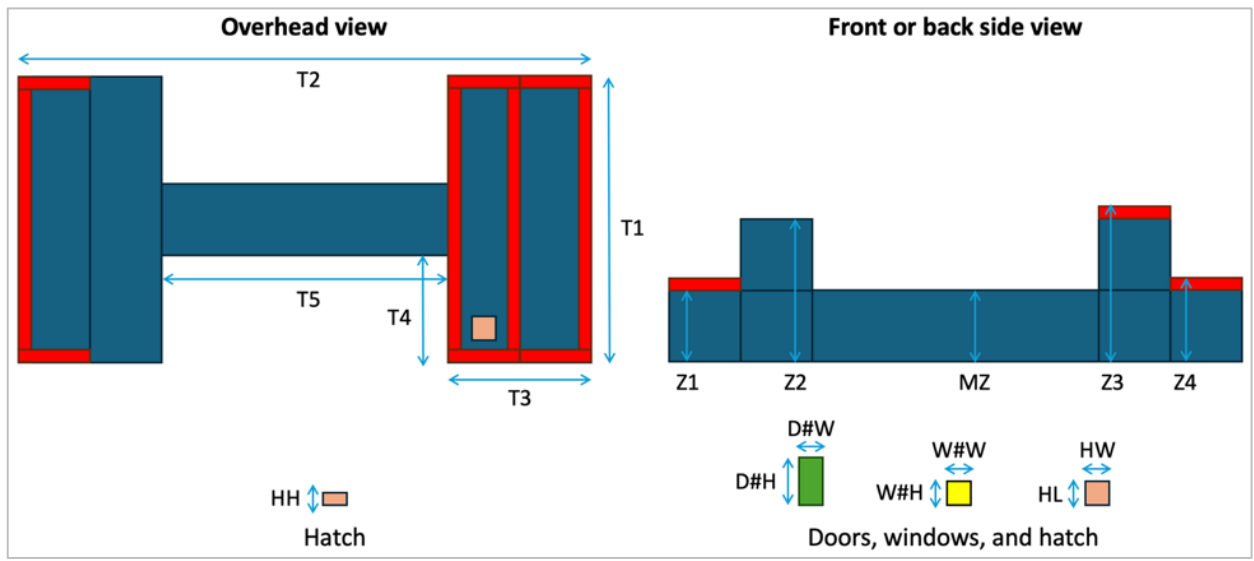

| Building Width/Length | | Building Height | | Doors | | | | Windows | | | | Hatch | |
|---|---|---|---|---|---|---|---|---|---|---|---|---|---|
| T1 | 12.19 | Z1 | 2.59 | D1W | 0.97 | D4W | 0.97 | W1W | 0.97 | W4W | 0.97 | HL | 0.99 |
| T2 | 21.95 | Z2 | 5.18 | D1H | 2.03 | D4H | 2.03 | W1H | 0.97 | W4H | 0.97 | HW | 0.99 |
| T3 | 4.88 | MZ | 2.59 | D2W | 0.97 | D5W | 0.97 | W2W | 0.97 | W5W | 0.97 | HH | 1.24 |
| T4 | 4.88 | Z3 | 6.15 | D2H | 2.03 | D5H | 2.03 | W2H | 0.97 | W5H | 0.97 | | |
| T5 | 12.19 | Z4 | 3.63 | D3W | 0.97 | D6W | 0.97 | W3W | 0.97 | W6W | 0.97 | | |
| | | | | D3H | 2.03 | D6H | 2.03 | W3H | 0.97 | W6H | 0.97 | | |

*Figure 47. Ground truth measurements of the complex building structure in meters.*

A summary of the Dimensional Accuracy test results is provided in Figure 48 followed by detailed results.

Achieving a lower average error (i.e., dimensions of the map are closer to that of the ground truth) at a faster processing time (i.e., less wait time to inspect the map) as desirable.

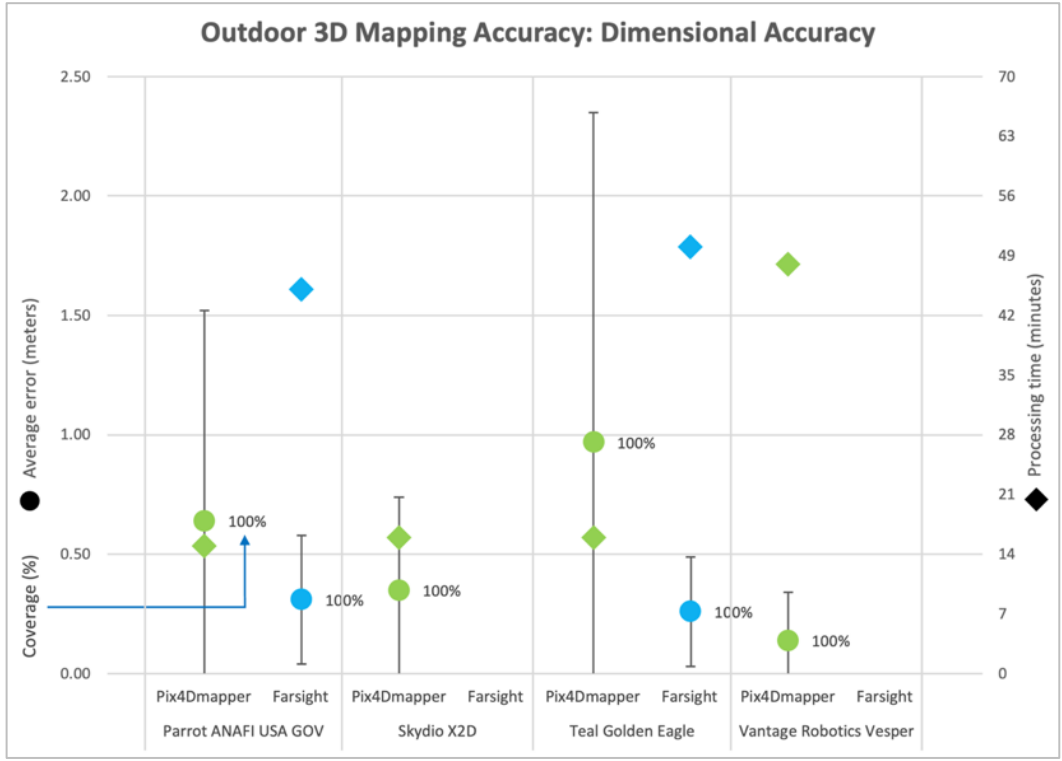

Figure 48. Summarized Outdoor Mapping Accuracy: Dimensional Accuracy results.

**Parrot ANAFI USA GOV**

| sUAS | Mapping Software | Metrics | Performance | Evaluation Map |
|---|---|---|---|---|
| Parrot ANAFI USA GOV | Pix4Dmapper | Flight pattern | Orbit | |
| | | Map source format | Video | |
| | | Map source size | 2:48 | |
| | | Mapping time | 3 min | |
| | | Processing time | 15 min | 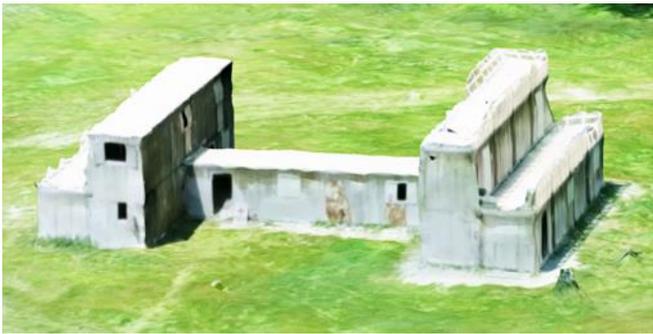 |
| | | Completeness | 100% | |
| | | Average error | 0.64 m (+/- 0.88 m) | |
| | Farsight | Processing time | 45 min | 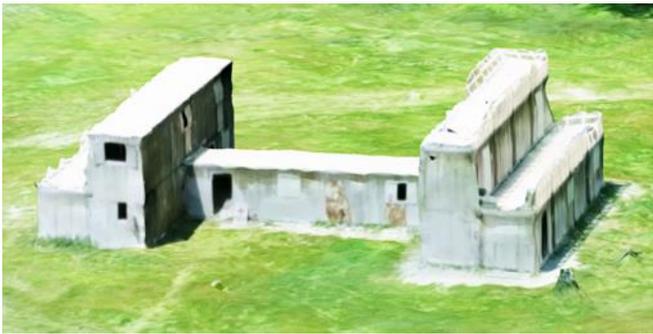 |
| | | Completeness | 100% | |
| | | Average error | 0.31 m (+/- 0.27 m) | |

*Table 14. Outdoor 3D Mapping: Dimensional Accuracy test results for the Parrot ANAFI USA GOV.*

## Skydio X2D

| sUAS | Mapping Software | Metrics | Performance | Evaluation Map |
|---|---|---|---|---|
| Skydio X2D | Pix4Dmapper | Flight pattern | Orbit | 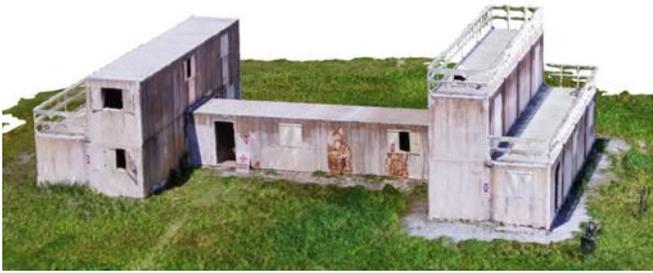 |
| | | Map source format | Video | |
| | | Map source size | 1:12 | |
| | | Mapping time | 1 min | |
| | | Processing time | 16 min | |
| | | Completeness | 100% | |
| | | Average error | 0.35 m (+/- 0.39 m) | |
| | Farsight | Processing time | N/A | This map was not able to be processed |
| | | Completeness | | |
| | | Average error | | |

*Table 15. Outdoor 3D Mapping: Dimensional Accuracy test results for the Skydio X2D.*

## Teal Golden Eagle

| sUAS | Mapping Software | Metrics | Performance | Evaluation Map |
|---|---|---|---|---|
| Teal Golden Eagle | Pix4Dmapper | Flight pattern | Orbit | |
| | | Map source format | Video | |
| | | Map source size | 1:13 | |
| | | Mapping time | 1 min | |
| | | Processing time | 16 min | 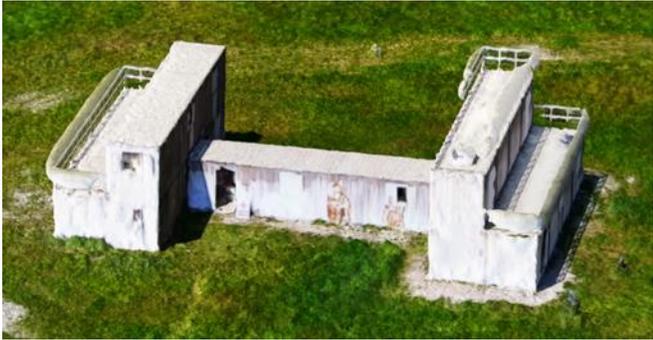 |
| | | Completeness | 95% | |
| | | Average error | 0.97 m (+/- 1.38 m) | |
| | Farsight | Processing time | 50 min | 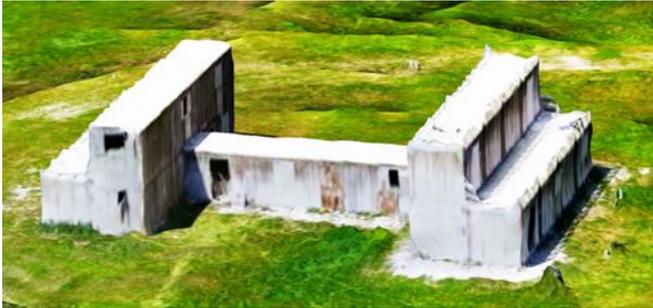 |
| | | Completeness | 100% | |
| | | Average error | 0.26 m (+/- 0.23 m) | |

*Table 16. Outdoor 3D Mapping: Dimensional Accuracy test results for the Teal Golden Eagle.*

**Vantage Robotics Vesper**

| sUAS | Mapping Software | Metrics | Performance | Evaluation Map |
|---|---|---|---|---|
| Vantage Robotics Vesper | Pix4Dmapper | Flight pattern | Orbit | 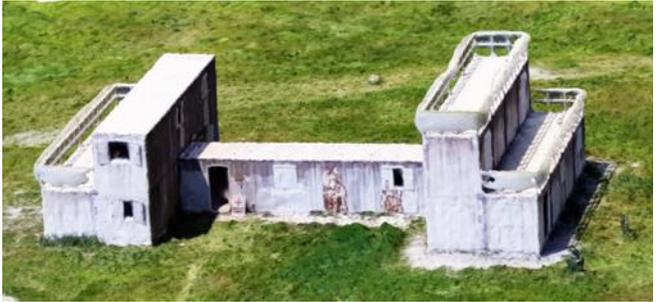 |
| | | Map source format | Video | |
| | | Map source size | 2:35 | |
| | | Mapping time | 3 min | |
| | | Processing time | 48 min | |
| | | Completeness | 100% | |
| | | Average error | 0.14 m (+/- 0.2 m) | |
| | Farsight | Processing time | N/A | This map was not able to be processed |
| | | Completeness | | |
| | | Average error | | |

Table 17. Outdoor 3D Mapping: Dimensional Accuracy test results for the Vantage Robotics Vesper.

# Feature Recognition

A total of 31 features were evaluated: 10 doors, 20 windows, and 1 hatch. See Table 18 for all ground truth photos (taken on a different day) and screenshots from the generated maps of each sUAS + mapping software pair.

A summary of the Feature Recognition test results as provided in Figure 49.

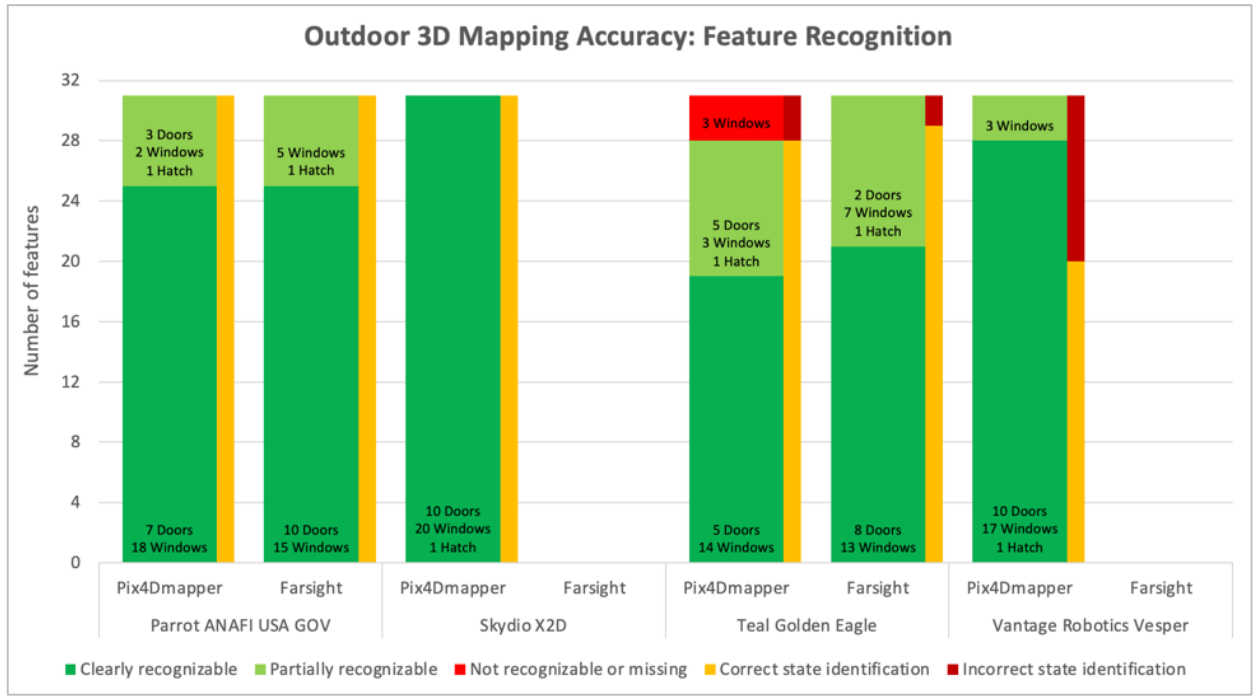

Table 18. Outdoor 3D Mapping: Feature Recognition ground truth and screenshots from each system's evaluation map for comparison.

Figure 49. Outdoor 3D Mapping Accuracy: Feature Recognition results.